\documentclass[runningheads]{llncs}
\usepackage{graphicx}
\usepackage{comment}
\usepackage{amsmath,amssymb} %
\usepackage{color}
\usepackage{subfig}
\usepackage{floatrow}
\usepackage[dvipsnames]{xcolor}
\usepackage[ruled]{algorithm2e}
\usepackage[utf8]{inputenc}
\usepackage[english]{babel}
\usepackage{siunitx}
\usepackage{pifont}
\usepackage{booktabs}
\newcolumntype{P}[1]{>{\centering\arraybackslash}p{#1}}
\newcolumntype{M}[1]{>{\centering\arraybackslash}m{#1}}
\usepackage{bbm}
\usepackage{multirow}

\newcommand{\etal}{\textit{et. al.}~}
\usepackage[utf8]{inputenc} %
\usepackage[T1]{fontenc}    %
\usepackage{url}            %
\usepackage{booktabs}       %
\usepackage{amsfonts}       %
\usepackage{nicefrac}       %
\usepackage{microtype}      %
\usepackage{tabularx}
\usepackage{paralist}
\usepackage{lmodern}

\usepackage{caption}

\usepackage[colorlinks=true,linkcolor=red]{hyperref}

\DeclareMathOperator*{\argmax}{arg\,max}
\definecolor{green}{RGB}{0,150,10}

\newcommand{\cmark}{\ding{51}}%
\newcommand{\xmark}{\ding{54}}%

\newcommand{\figLabel}{Fig.~}
\newcommand{\eg}{\textit{e.g.~}}
\newcommand{\ie}{\textit{i.e.~}}
\newcommand{\eqLabel}[1]{{Eq (#1)}}
\newcommand{\secLabel}{Section~}

\newcommand{\mysection}[1]{\noindent\textbf{#1.}}
\newcommand{\supp}{\textbf{supplement~}}
\newcommand{\specialcell}[2][c]{%
  \begin{tabular}[#1]{@{}c@{}}#2\end{tabular}}

\begin{document}
\pagestyle{headings}
\mainmatter
\def\ECCVSubNumber{1541}  %

\title{AdvPC: Transferable Adversarial Perturbations on 3D Point Clouds} %

\titlerunning{AdvPC: Transferable Adversarial Perturbations on 3D Point Clouds}
\author{Abdullah Hamdi \and Sara Rojas  \and Ali Thabet \and Bernard Ghanem }
\authorrunning{A. Hamdi, S. Rojas, A. Thabet, B. Ghanem}
\institute{King Abdullah University of Science and Technology (KAUST), Thuwal, Saudi Arabia\\
\email{\{abdullah.hamdi, sara.rojasmartinez, ali.thabet, bernard.ghanem\}@kaust.edu.sa}}
\maketitle

\begin{abstract}
Deep neural networks are vulnerable to adversarial attacks, in which imperceptible perturbations to their input lead to erroneous network predictions.
This phenomenon has been extensively studied in the image domain, and has only recently been extended to 3D point clouds. In this work, we present novel data-driven adversarial attacks against 3D point cloud networks. 
We aim to address the following problems in current 3D point cloud adversarial attacks: they do not transfer well between different networks, and they are easy to defend against via simple statistical methods. 
To this extent, we develop a new point cloud attack (dubbed AdvPC) that exploits the input data distribution by adding an adversarial loss, after Auto-Encoder reconstruction, to the  objective it optimizes. AdvPC leads to perturbations that are resilient against current defenses, while remaining highly transferable compared to state-of-the-art attacks. We test AdvPC using four popular point cloud networks: PointNet, PointNet++ (MSG and SSG), and DGCNN. Our proposed attack increases the attack success rate by up to 40\% for those transferred to unseen networks (transferability), while maintaining a high success rate on the attacked network. AdvPC also increases the ability to break defenses by up to 38\% as compared to other baselines on the ModelNet40 dataset. The code is available at \url{https://github.com/ajhamdi/AdvPC}.
\end{abstract}
\section{Introduction} \label{sec:intro}
Deep learning has shown impressive results in many perception tasks. Despite its performance, several works show that deep learning algorithms can be susceptible to adversarial attacks. These attacks craft small perturbations to the inputs that push the network to produce incorrect outputs. There is significant progress made in 2D image adversarial attacks, where extensive work shows diverse ways to attack 2D neural networks \cite{first-attack,fast-sign,projected-gradient,deepfool,carlini,strike,physicalattack,sada,semantic-robustness}. In contrast, there is little focus on their 3D counterparts \cite{pcattack,Deflecting,pointdrop,robustshapeattack}. 3D point clouds captured by 3D sensors like LiDAR are now widely processed using deep networks for safety-critical applications, including but not limited to self-driving \cite{lidar-adv,3d-obj-attk}. However, as we show in this paper, 3D deep networks tend to be vulnerable to input perturbations, a fact that increases the risk of using them in such applications. In this paper, we present a novel approach to attack deep learning algorithms applied to 3D point clouds with a primary focus on attack transferability between networks.

\begin{figure}[t]
\begin{center}
   \includegraphics[width=1\linewidth,trim={0cm 1.1cm 0cm 0.3cm},clip,]{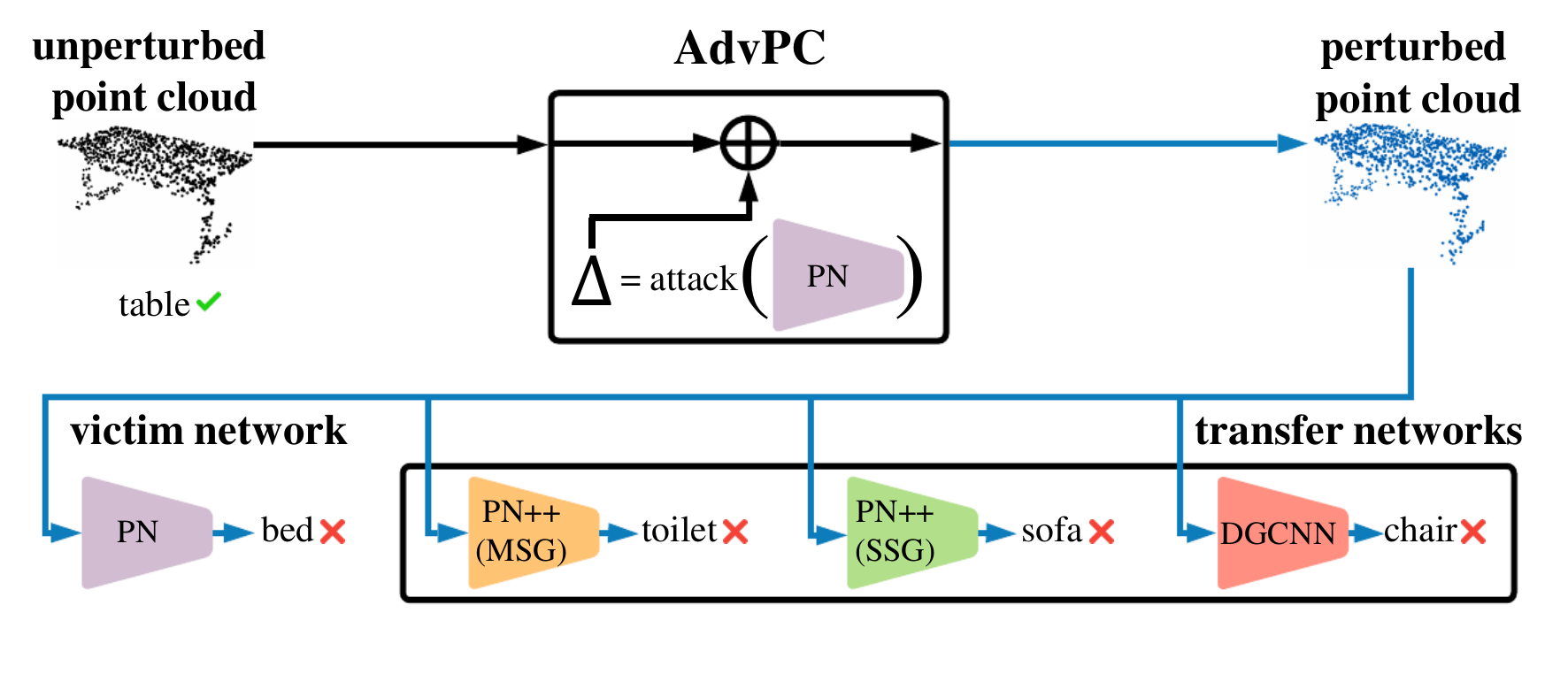} %
\end{center}
   \caption{\small \textbf{Transferable Adversarial Perturbations on 3D point clouds}: Generating adversarial attacks to fool PointNet \cite{pointnet}(PN) by perturbing a Table point cloud. The perturbed 3D object not only forces PointNet to predict an incorrect class, but also induces misclassification on other unseen 3D networks (PointNet++ \cite{pointnet++}, DGCNN \cite{dgcn}) that are not involved in generating the perturbation. Fooling unseen networks poses a threat to 3D deep vision models.}
\label{fig:intro}
\end{figure}

The concept of attack transferability has been extensively studied in the 2D image domain  \cite{universal-attack,learnattack1,learnattack2}. Transferability allows an adversary to fool any network, without access to the network's architecture. Clearly, transferable attacks pose a serious security concern, especially in the context of deep learning model deployment. In this work, the goal is to generate adversarial attacks with network-transferability, \ie the attack to a given point cloud is generated using a single and accessible \textit{victim} network, and the perturbed sample is directly applied to an unseen and inaccessible \textit{transfer} network.%
Accessibility here refers to whether the parameters and architecture of the network are known, while optimizing the attack (white-box). \figLabel{\ref{fig:intro}} illustrates the concept of transferability. The perturbation generated by our method for a 3D point cloud not only flips the class label of a victim network to a wrong class (\ie it is adversarial), but it also induces a misclassification for the transfer networks that are not involved in generating the perturbation (\ie it is transferable).  

Very few adversarial attacks have been developed for 3D point clouds. The first method was introduced by Xiang \etal \cite{pcattack} and it proposes point perturbation and adversarial point generation as two attack modes. More recently, Tsai \etal \cite{robustshapeattack} proposed to make point cloud attacks more smooth and natural by incorporating a K-Nearest Neighbor (KNN) loss on the points, thus making the attacks physically realizable. We identify two main shortcomings in current 3D adversarial perturbations methods \cite{pcattack,robustshapeattack}. First, their attacks are unsuccessful in the presence of simple defenses, such as Statistical Outlier Removal \cite{Deflecting}. Second, they are limited to the victim network and do not transfer well to other networks \cite{pcattack}. In contrast, our work not only focuses on adversarial perturbations that are significantly more resilient against currently available point cloud defenses, but also on those that transfer well between different point cloud networks.

To generate more transferable attacks, we use a point cloud Auto-Encoder (AE), which can effectively reconstruct the unperturbed input after it is perturbed, and then add a data adversarial loss. We optimize the perturbation added to the input to fool the classifier \textit{before} it passes through the AE (regular adversarial loss) and \textit{after} it passes through the AE (data adversarial loss). In doing so, the attack tends to be less dependent on the victim network, and generalizes better to different networks. Our attack is dubbed ``AdvPC'', and our full pipeline is optimized end-to-end from the classifier output to the perturbation. The AE learns the natural distribution of the data to generalize the attack to a broader range of unseen classifiers \cite{autozoom}, thus making the attack more dangerous. Our attacks surpass state-of-the-art attacks \cite{pcattack,robustshapeattack} by a large margin (up to 40\%) on point cloud networks operating on the standard  ModelNet40 dataset \cite{modelnet} and for the same maximum allowed perturbation norms (norm-budgets). 

\mysection{Contributions} Our contributions are two-fold. \textbf{(1)} We propose a new pipeline and loss function to perform transferable adversarial perturbations on 3D point clouds. By introducing a data adversarial loss targeting the victim network after reconstructing the perturbed input with a point cloud AE, our approach can be successful in both attacking the victim network and transferring to unseen networks. %
Since the AE is trained to leverage the point cloud data distribution, incorporating it into the attack strategy enables better transferability to unseen networks. To the best of our knowledge, we are the first to introduce network-transferable adversarial perturbations for 3D point clouds. %
\textbf{(2)} We perform extensive experiments under constrained norm-budgets to validate the transferability of our attacks. We transfer our attacks between four point cloud networks and show superiority against the state-of-the-art. Furthermore, we demonstrate how our attacks outperform others when targeted by currently available point cloud defenses.

\section{Related Work} \label{sec:related}
\subsection{Deep Learning for 3D Point Clouds}
PointNet \cite{pointnet} paved the way as the first deep learning algorithm to operate directly on 3D point clouds. PointNet computes point features independently, and aggregates them using an order invariant function like max-pooling. An update to this work was PointNet++ \cite{pointnet++}, where points are aggregated at different 3D scales. Subsequent works focused on how to aggregate more local context \cite{pc_engelmann2018} or on more complex aggregation strategies like RNNs \cite{pc_huang2018recurrent,pc_ye20183d}. More recent methods run convolutions across neighbors of points, instead of using point-wise operations \cite{dgcn,pc_li2018pointcnn,pc_tatarchenko2018tangent,pc_landrieu2018large,pc_landrieu2019point,pc_li2018pointcnn,pc_wang2018sgpn,pc_li2018so}. Contrary to PointNet and its variants, these works achieve superior recognition results by focusing on local feature representation. In this paper and to evaluate/validate our adversarial attacks, we use three point-wise networks, PointNet \cite{pointnet} and PointNet++ \cite{pointnet++} in single-scale (SSG) and multi-scale (MSG) form, and a Dynamic Graph convolutional Network, DGCNN \cite{dgcn}. We study the sensitivity of each network to adversarial perturbations and show the transferability of AdvPC attacks between the networks. %

\subsection{Adversarial Attacks} \label{sec:related_adver}
\mysection{Pixel-based Adversarial Attacks}
The initial image-based adversarial attack was introduced by Szegedy \etal \cite{first-attack}, who cast the attack problem as optimization with pixel perturbations being minimized so as to fool a trained classifier into predicting a wrong class label. Since then, the topic of adversarial attacks has attracted much attention \cite{fast-sign,projected-gradient,deepfool,carlini,pgd-madry}. More recent works take a learning-based approach to the attack \cite{learnattack1,learnattack2,gan-attack}. They train a neural network (adversary) to perform the attack and then use the trained adversary model to attack unseen samples. 
These learning approaches \cite{learnattack1,learnattack2,gan-attack} tend to have better transferability properties than the optimizations approaches \cite{fast-sign,projected-gradient,deepfool,carlini,pgd-madry}, while the latter tend to achieve higher success rates on the victim networks. %
As such, our proposed AdvPC attack is a \textit{hybrid} approach, in which we leverage an AE to capture properties of the data distribution but still define the attack as an optimization for each sample. In doing so, AdvPC captures the merits of both learning \textit{and} optimization methods to achieve high success rates on the victim networks as well as better transferability to unseen networks.

\mysection{Adversarial Attacks in 3D}
Several adversarial attacks have moved beyond pixel perturbations to the 3D domain. One line of work focuses on attacking image-based CNNs by changing the 3D parameters of the object in the image, instead of changing the pixels of the image  \cite{sada,physicalattack,strike,semantic-robustness,meshadv}.
Recently, Xiang \etal \cite{pcattack} developed adversarial perturbations on 3D point clouds, which were successful in attacking PointNet \cite{pointnet}; however, this approach has two main shortcomings. First, it can be easily defended against by simple statistical operations \cite{Deflecting}. Second, the attacks are non-transferable and only work on the attacked network \cite{pcattack,Deflecting}. In contrast, Zheng \etal \cite{pointdrop} proposed dropping points from the point cloud using a saliency map, to fool trained 3D deep networks. As compared to \cite{pointdrop}, our attacks are modeled as an optimization on the additive perturbation variable with a focus on point perturbations instead of point removal.  As compared to \cite{pcattack}, our AdvPC attacks are significantly more successful against available defenses and more transferable beyond the victim network, since AdvPC leverages the point cloud data distribution through the AE. Concurrent to our work is the work of Tsai \etal \cite{robustshapeattack}, in which the attack is crafted with KNN loss to make smooth and natural shapes. The motivation of their work is to craft natural attacks on 3D point clouds that can be 3D-printed into real objects. In comparison, our novel AdvPC attack utilizes the data distribution of point clouds by utilizing an AE to generalize the attack.

\mysection{Defending Against 3D Point Cloud Attacks}
Zhou \etal \cite{Deflecting} proposed a Statistical Outlier Removal (SOR) method as a defense against point cloud attacks. SOR uses KNN to identify and remove point outliers.
They also propose DUP-Net, which is a combination of their SOR and a point cloud up-sampling network PU-Net \cite{punet}. %
Zhou \etal also proposed removing unnatural points by Simple Random Sampling (SRS), where each point has the same probability of being randomly removed. %
Adversarial training on the attacked point cloud is also proposed as a mode of defense by \cite{pcattack}.
 Our attacks surpass state-of-the-art attacks \cite{pcattack,robustshapeattack} on point cloud networks by a large margin (up to 38\%) on the standard ModelNet40 dataset \cite{modelnet} against the aforementioned defenses \cite{Deflecting}.

\begin{figure*}[t]
\begin{center}
   \includegraphics[width=1\linewidth]{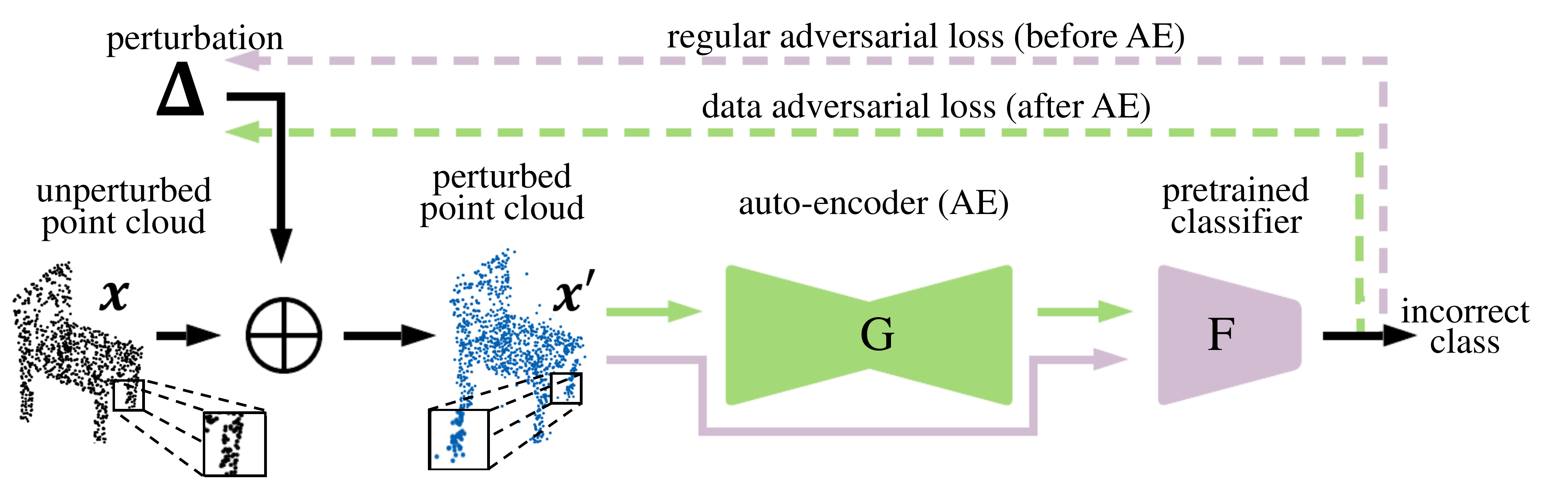} %
     \end{center}
   \caption{\small \textbf{AdvPC Attack Pipeline:} We optimize for the constrained perturbation variable $\boldsymbol{\Delta}$ to generate the perturbed sample $\mathcal{X}^{\prime} = \mathcal{X} + \boldsymbol{\Delta}$. The perturbed sample fools a trained classifier $\mathbf{F}$ (\ie $\mathbf{F}(\mathcal{X}^{\prime})$ is incorrect), and at the same time, if the perturbed sample is reconstructed by an Auto-Encoder (AE) $\mathbf{G}$, it too fools the classifier (\ie $\mathbf{F}(\mathbf{G}(\mathcal{X}^{\prime}))$ is incorrect). The AdvPC loss for network $\mathbf{F}$ is defined in \eqLabel{\ref{eq:final-objective}} and has two parts: network adversarial loss (\textit{\textit{purple}}) and data adversarial loss (\textit{green}). Dotted lines are gradients flowing to the perturbation variable $\boldsymbol{\Delta}$.}

\label{fig::pipeline}
\end{figure*}
\section{Methodology} \label{sec:methodology}
The pipeline of AdvPC is illustrated in \figLabel{\ref{fig::pipeline}}. It consists of an Auto-Encoder (AE) $\mathbf{G}$, which is trained to reconstruct 3D point clouds and a point cloud classifier $\mathbf{F}$.  
We seek to find a perturbation variable $\boldsymbol{\Delta}$ added to the input $\mathcal{X}$ to fool $\mathbf{F}$  before \emph{and} after it passes through the AE for reconstruction. The setup makes the attack less dependent on the victim network and more dependent on the data. As such, we expect this strategy to generalize to different networks.
Next, we describe the main components of our pipeline: 3D point cloud input, AE, and point cloud classifier. Then, we present our attack setup and loss. %
\subsection{AdvPC Attack Pipeline}
\mysection{3D Point Clouds ($\mathcal{X}$)}
We define a point cloud $\mathcal{X} \in \mathbb{R}^{N \times 3}$, as a set of $N$ 3D points, where each point  $\mathbf{x}_i \in \mathbb{R}^{3}$ is represented by its 3D coordinates $(x_i, y_i, z_i)$. 

\mysection{Point Cloud Networks ($\mathbf{F}$)}
We focus on 3D point cloud classifiers with a feature max pooling layer as detailed in \eqLabel{\ref{eq:pcnet}}, where $h_{\text{mlp}}$ and $h_{\text{conv}}$ are MLP and Convolutional ($1\times1$ or edge) layers, respectively. This produces a K-class classifier $\mathbf{F}$.
\begin{equation}
\label{eq:pcnet}
\mathbf{F}(\mathcal{X})=h_{\text{mlp}} ( \max _{\mathbf{x}_{i} \in \mathcal{X}}\left\{h_{\text{conv}}\left(\mathbf{x}_{i}\right)\right\})
\end{equation}
Here, $\mathbf{F} : ~ \mathbb{R}^{N \times 3} \rightarrow \mathbb{R}^K $ produces the logits layer of the classifier with size $K$. 
For our attacks, we take $\mathbf{F}$ to be one of the following widely used networks in the literature: PointNet \cite{pointnet}, PointNet++ \cite{pointnet++} in single-scale form (SSG)  and multi-scale form (MSG), and DGCNN \cite{dgcn}. %
\secLabel{\ref{sec:sens}} delves deep into the differences between them in terms of their sensitivities to adversarial perturbations.

\mysection{Point Cloud Auto-Encoder ($\mathbf{G}$)}
An AE learns a representation of the data and acts as an effective defense against adversarial attacks. It ideally projects a perturbed point cloud onto the natural manifold of inputs. Any AE architecture in point clouds can be used, but we select the one in \cite{pc-ae} because of its simple structure and effectiveness in recovering from adversarial perturbation. 
The AE $\mathbf{G}$ consists of an encoding part, $\mathbf{g}_{\text{encode}}:\mathbb{R}^{N \times 3} \xrightarrow{}\mathbb{R}^{q} $ (similar to \eqLabel{\ref{eq:pcnet}}),
and an MLP decoder, $\mathbf{g}_{\text{mlp}}:\mathbb{R}^{q} \xrightarrow{}\mathbb{R}^{N \times 3} $, to produce a point cloud. It can be described formally as: $\mathbf{G}(.) = \mathbf{g}_{\text{mlp}}\big(\mathbf{\mathbf{g}_{\text{encode}}(\mathcal{.})} \big)$.
We train the AE with the Chamfer loss as in \cite{pc-ae} on the same data used to train $\mathbf{F}$, such that it can reliably encode and decode 3D point clouds. We freeze the AE weights during the optimization of the adversarial perturbation on the input.  
 Since the AE learns how naturally occurring point clouds look like, the gradients updating the attack, which is also tasked to fool the reconstructed sample after the AE, actually become more dependent on the data and less on the victim network. The enhanced data dependency of our attack results in the success of our attacks on unseen transfer networks besides the success on the victim network.
As such, the proposed composition allows the crafted attack to successfully attack the victim classifier, as well as, fool transfer classifiers that operate on a similar input data manifold. 
\subsection{AdvPC Attack Loss} 
\begin{figure}[t]

\tabcolsep=0.2cm

\begin{tabular}{cccccc}

\includegraphics[trim={1.5cm 2.5cm 1.5cm 3.4cm},clip, width = 0.75in]{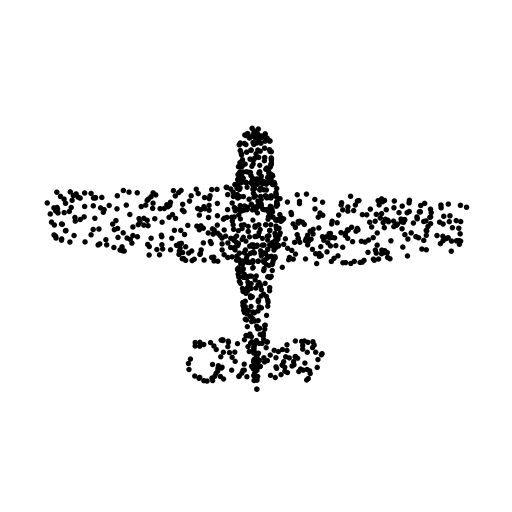} &
 \includegraphics[trim={0cm 1cm 0cm 1.5cm},clip, width = 0.75in]{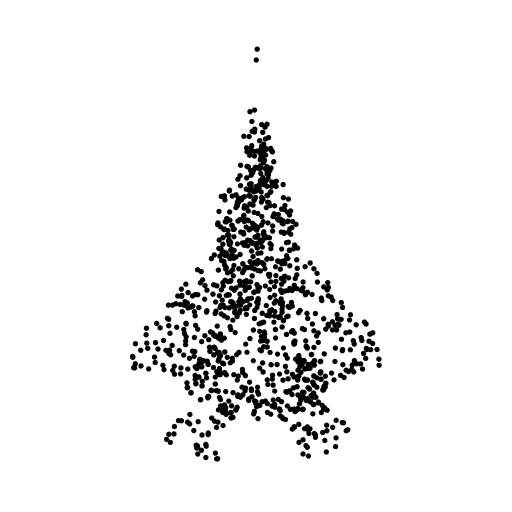} & 
\includegraphics[trim={5.5cm 7cm 6cm 1.5cm},clip, width = 0.25in]{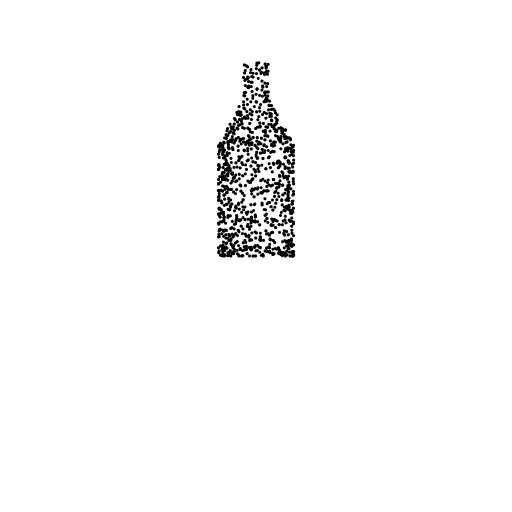} &
\includegraphics[trim={6cm 7cm 6cm 1.5cm},clip, width = 0.19in]{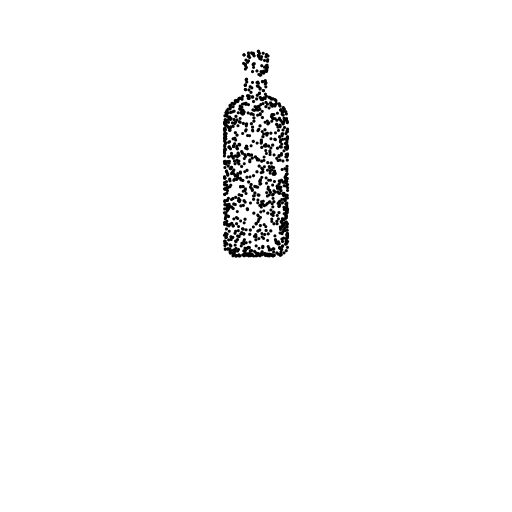} & 
\includegraphics[trim={4cm 6cm 4cm 1.5cm},clip, width = 0.55in]{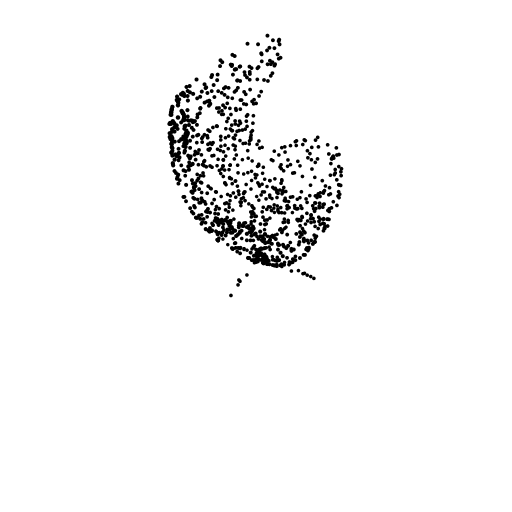} &
\includegraphics[trim={4cm 6cm 4cm 1.5cm},clip, width = 0.55in]{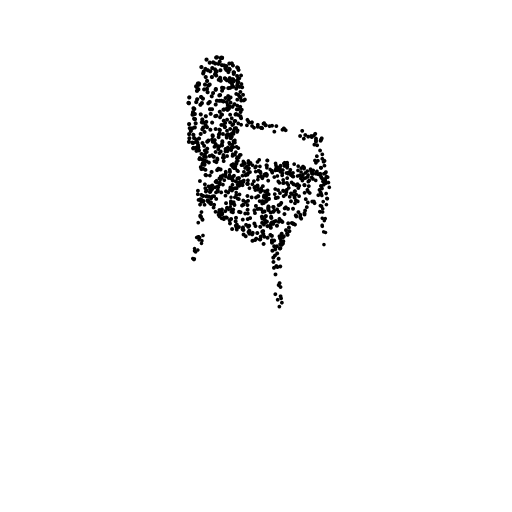} \\

airplane \color{ForestGreen}\cmark &
airplane \color{ForestGreen}\cmark &
bottle \color{ForestGreen}\cmark &
bottle \color{ForestGreen}\cmark &
chair \color{ForestGreen}\cmark &
chair \color{ForestGreen}\cmark \\

\includegraphics[trim={1.5cm 2.5cm 1.5cm 3.4cm},clip, width = 0.75in]{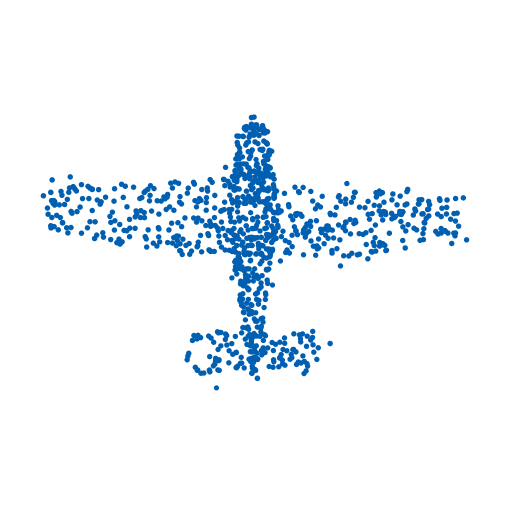} &
 \includegraphics[trim={0cm 1cm 0cm 1.5cm},clip, width = 0.75in]{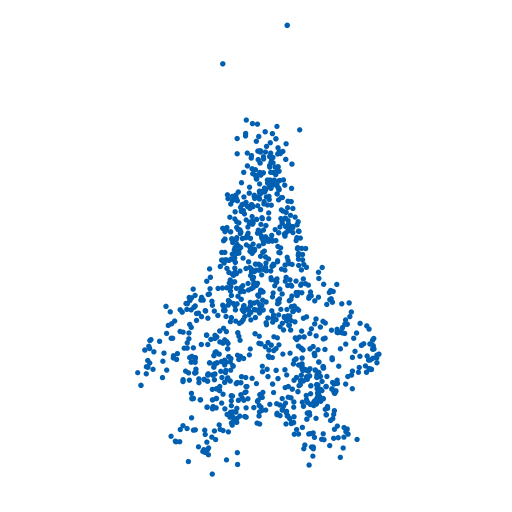} &
\includegraphics[trim={5.5cm 7cm 6cm 1.5cm},clip, width = 0.25in]{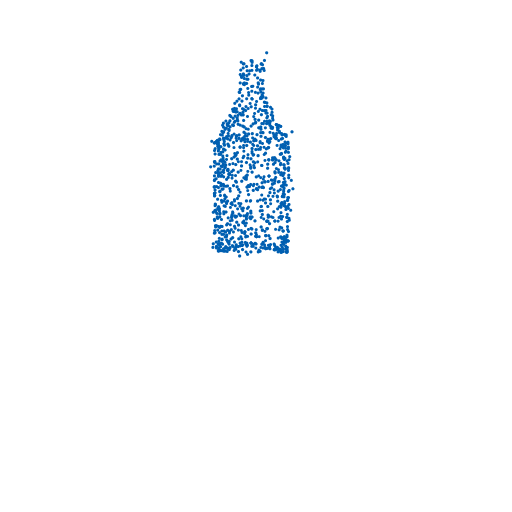} &
\includegraphics[trim={6cm 7cm 6cm 1.5cm},clip, width = 0.19in]{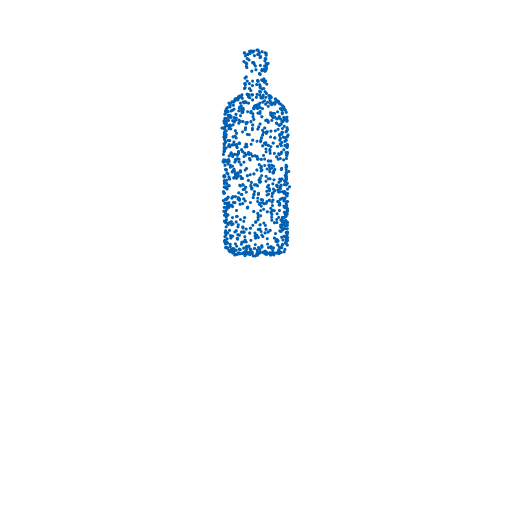} &
\includegraphics[trim={4cm 6cm 4cm 1.5cm},clip, width = 0.55in]{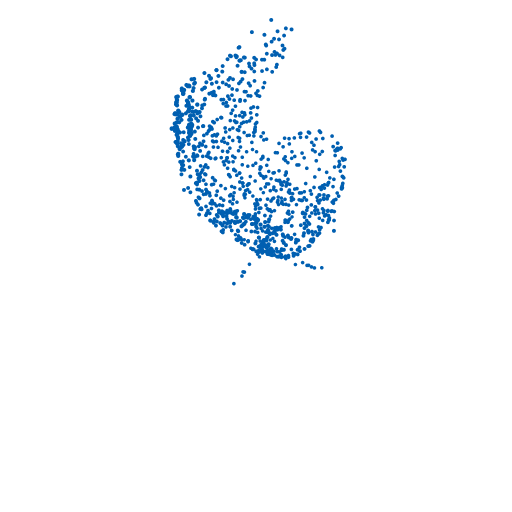} &
\includegraphics[trim={4cm 6cm 4cm 1.5cm},clip, width = 0.55in]{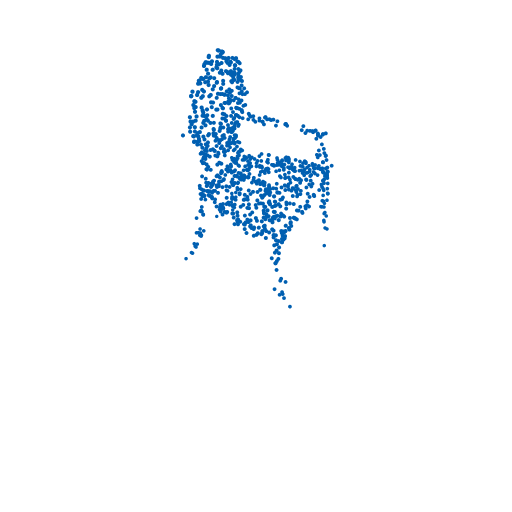} \\
PN++ (SSG):&
DGCNN: &
PN &
PN++ (MSS):&
PN:&
PN:\\
bookshelf  \color{red}\xmark &
bottle \color{red}\xmark &
table \color{red}\xmark &
monitor  \color{red}\xmark &
airplane \color{red}\xmark &
vase \color{red}\xmark \\
\end{tabular}
\caption{\small \textbf{Examples of AdvPC Attacks:} Adversarial attacks are generated for victim networks PointNet, PointNet ++ (MSG/SSG) and DGCNN using AdvPC. The unperturbed point clouds are in black (\textit{top}) while the perturbed examples are in blue (\textit{bottom}). The network predictions are shown under each point cloud. The wrong prediction of each perturbed point cloud matches the target of the AdvPC attack. %
}
\label{fig:qualitative}
\end{figure}

\mysection{Soft Constraint Loss}
In AdvPC attacks, like the ones in \figLabel{\ref{fig:qualitative}}, we focus solely on perturbations of the input. We modify each point $\mathbf{x}_i$ by a an addictive perturbation variable $\delta_i$. Formally, we define the perturbed point set $\mathcal{X}^{\prime} = \mathcal{X} + \boldsymbol{\Delta}$, where $\boldsymbol{\Delta} \in \mathbb{R}^{N \times 3}$ is the perturbation parameter we are optimizing for. Consequently, each pair ($\mathbf{x}_i, \mathbf{x}^{\prime}_i$) are in correspondence. Adversarial attacks are commonly formulated as in \eqLabel{\ref{eq:adv-attack}}, where the goal is to find an input perturbation $\boldsymbol{\Delta}$ that successfully fools $\mathbf{F}$ into predicting an incorrect label $t^{\prime}$, while 
keeping $\mathcal{X^{\prime}}$ and $\mathcal{X}$ close under distance metric  $\mathcal{D}\colon \mathbb{R}^{N\times3} \times \mathbb{R}^{N\times3} \rightarrow \mathbb{R}$. 
\begin{equation} 
\label{eq:adv-attack}
\min_{\boldsymbol{\Delta}} ~~\mathcal{D}\left(\mathcal{X}, \mathcal{X}^{\prime}\right) \quad \text { s.t. } \left[\argmax_{i}~\mathbf{F}\left(\mathcal{X}^{\prime}\right)_{i}\right]=t^{\prime}
\end{equation}
The formulation in \eqLabel{\ref{eq:adv-attack}} can describe targeted attacks (if $t^\prime$ is specified before the attack) or untargeted attacks (if $t^\prime$ is any label other than the true label of $\mathcal{X}$). We adopt the following choice of $t^\prime$ for untargeted attacks: $t^\prime = \left[\argmax_{i\neq \text{true}}~\mathbf{F}\left(\mathcal{X}^{\prime}\right)_{i}\right]$. Unless stated otherwise, we primarily use untargeted attacks  in this paper.
As pointed out in \cite{carlini}, it is difficult to directly solve \eqLabel{\ref{eq:adv-attack}}. Instead, previous works like \cite{pcattack,robustshapeattack} have used the well-known C\&W formulation, giving rise to the commonly known soft constraint attack: $\min_{\boldsymbol{\Delta}} ~~f_{t^{\prime}}\left(\mathbf{F}(\mathcal{X}^{\prime})\right)+\lambda \mathcal{D}\left(\mathcal{X}, \mathcal{X}^{\prime}\right)$ 
where $f_{t^{\prime}}\left(\mathbf{F}(\mathcal{X}^{\prime})\right)$ is the adversarial loss function defined on the network $\mathbf{F}$ to move it to label $t^{\prime}$ as in \eqLabel{\ref{eq:adv-loss}}.  %
\begin{equation} 
\label{eq:adv-loss}
f_{t^{\prime}}\left(\mathbf{F}(\mathcal{X}^{\prime})\right)=\max\left(\max _{i \neq t^{\prime}}\left(\mathbf{F}\left(\mathcal{X}^{\prime}\right)_{i}\right)-\mathbf{F}\left(\mathcal{X}^{\prime}\right)_{t^{\prime}} + \kappa,0\right),
\end{equation}
where $\kappa$ is a loss margin. The 3D-Adv attack \cite{pcattack} uses $\ell_2$ for $\mathcal{D}\left(\mathcal{X}, \mathcal{X}^{\prime}\right)$, while the KNN Attack \cite{robustshapeattack} uses Chamfer Distance.

\mysection{Hard Constraint Loss}
An alternative to \eqLabel{\ref{eq:adv-attack}} is to put $\mathcal{D}\left(\mathcal{X}, \mathcal{X}^{\prime}\right)$ as a hard constraint, where the objective can be minimized using Projected Gradient Descent (PGD) \cite{projected-gradient,pgd-madry} as follows. 
\begin{equation} 
\label{eq:attack-hard}
\min_{\boldsymbol{\Delta}} ~~ f_{t^{\prime}}\left(\mathbf{F}(\mathcal{X}^{\prime})\right) ~~~~ s.t.~~ \mathcal{D}\left(\mathcal{X}, \mathcal{X}^{\prime}\right) \leq \epsilon 
\end{equation}
Using a hard constraint sets a limit to the amount of added perturbation in the attack. This limit is defined by $\epsilon$ in \eqLabel{\ref{eq:attack-hard}}, which we call norm-budget in this work. Having this bound ensures a fair comparison between different attack schemes. We compare these schemes by measuring their attack success rate at different levels of norm-budget.
Using PGD, the above optimization in \eqLabel{\ref{eq:attack-hard}} with $\ell_p$ distance $\mathcal{D}_{\ell_p}\left(\mathcal{X}, \mathcal{X}^{\prime}\right)$ can be solved by iteratively projecting the perturbation $\boldsymbol{\Delta}$ onto the $\ell_p$ sphere of size $\epsilon_p$ after each gradient step such that:  $\boldsymbol{\Delta}_{t+1} =  \Pi_{p}\left( \boldsymbol{\Delta}_{t} - \eta \nabla _{\boldsymbol{\Delta}_{t}}f_{t^{\prime}}\left(\mathbf{F}(\mathcal{X}^{\prime})\right),\epsilon_p \right)$.
Here, $\Pi_{p}\left(\boldsymbol{\Delta},\epsilon_p\right)$  projects the perturbation $\boldsymbol{\Delta}$ onto the $\ell_p$ sphere of size $\epsilon_p$, and $\eta$ is a step size. The two most commonly used $\ell_p$ distance metrics in the literature are $\ell_2$, which measures the energy of the perturbation, and $\ell_\infty$, which measures the maximum point perturbation of each $\boldsymbol{\delta}_i \in \boldsymbol{\Delta}$. In our experiments, we choose to use the $\ell_\infty$ distance defined as 
$\mathcal{D}_{\ell_\infty}\left(\mathcal{X}, \mathcal{X}^{\prime}\right)=  \max _{i} 
\left\|\boldsymbol{\delta}_i\right\|_{\infty}$,
The projection of $\boldsymbol{\Delta}$ onto the $\ell_\infty$ sphere of size $\epsilon_\infty$ is: $\Pi_{\infty}\left(\boldsymbol{\Delta},\epsilon_{\infty}\right) = \text{SAT}_{\epsilon_{\infty}}(\boldsymbol{\delta}_{i}),~ \forall \boldsymbol{\delta}_{i} \in \boldsymbol{\Delta}$,
where $\text{SAT}_{\epsilon_\infty}\left(\boldsymbol{\delta}_{i}\right)$ is the element-wise saturation function that takes every element of vector $\boldsymbol{\delta}_{i}$ and limits its range to $[-\epsilon_\infty,\epsilon_\infty]$. Norm-budget $\epsilon_\infty$ is used throughout the experiments in this work.

In \supp\!\!, we detail our formulation when $\ell_2$ is used as the distance metric and report similar superiority over the baselines just as the $\ell_\infty$ results. For completeness, we also show in the supplement the effect of using different distance metrics ($\ell_2$, Chamfer, and Earth Mover Distance) as soft constraints on transferability and attack effectiveness.

\mysection{Data Adversarial Loss}
The objectives in \eqLabel{\ref{eq:adv-attack}, \ref{eq:attack-hard}} focus solely on the network $\mathbf{F}$. We also want to add more focus on the data in crafting our attacks. We do so by fooling $\mathbf{F}$ using both the perturbed input $\mathcal{X}^{\prime}$ and the AE reconstruction $\mathbf{G}(\mathcal{X}^{\prime})$ (see \figLabel{\ref{fig::pipeline}}). Our new objective becomes:
\begin{equation} 
\label{eq:pre-final-objective}
\begin{aligned}
\min_{\boldsymbol{\Delta}}  ~~\mathcal{D}\left(\mathcal{X}, \mathcal{X}^{\prime}\right) \quad
    \text { s.t. } [\argmax _{i}~\mathbf{F}\left(\mathcal{X}^{\prime}\right)_{i}]= t^{\prime};~~ [\argmax _{i}~\mathbf{F}\left(\mathbf{G}(\mathcal{X}^{\prime})\right)_{i}] = t^{\prime\prime}
\end{aligned}
\end{equation}
Here, $t^{\prime\prime}$ is any incorrect label $ t^{\prime\prime} \neq \argmax _{i}\mathbf{F}\left(\mathcal{X}\right)_{i}$ and $t^{\prime}$ is just like \eqLabel{\ref{eq:adv-attack}}. The second constraint ensures that the prediction of the perturbed sample after the AE %
differs from the true label of the unperturbed sample. %
Similar to \eqLabel{\ref{eq:adv-attack}}, this objective is hard to optimize, so we follow similar steps as in \eqLabel{\ref{eq:attack-hard}} and optimize the following objective for AdvPC using PGD (with $\ell_\infty$ as the distance metric): 
\begin{equation} 
\label{eq:final-objective}
\begin{aligned}
\min_{\boldsymbol{\Delta}} ~ (1 - \gamma) ~f_{t^{\prime}}\left(\mathbf{F}(\mathcal{X}^{\prime})\right) + \gamma ~ f_{t^{\prime\prime}}\left(\mathbf{F}\left(\mathbf{G}(\mathcal{X}^{\prime})\right)\right)~~~ s.t. ~~ \mathcal{D}_{\ell_\infty}\left(\mathcal{X}, \mathcal{X}^{\prime}\right) \leq \epsilon_\infty 
\end{aligned}
\end{equation}
Here, $f$ is as in \eqLabel{\ref{eq:adv-loss}}, while $\gamma$ is a hyper-parameter that trades off the attack's success before and after the AE . %
When $\gamma = 0$, the formulation in \eqLabel{\ref{eq:final-objective}} becomes \eqLabel{\ref{eq:attack-hard}}. We use PGD to solve \eqLabel{\ref{eq:final-objective}} just like \eqLabel{\ref{eq:attack-hard}}. We follow the same procedures as in \cite{pcattack} when solving  \eqLabel{\ref{eq:final-objective}} by keeping a record of any $\boldsymbol{\Delta}$ that satisfies the constraints in \eqLabel{\ref{eq:pre-final-objective}} and by trying different initializations for $\boldsymbol{\Delta}$.

\section{Experiments} \label{sec:experiments}
\subsection{Setup} \label{sec:setup}
\mysection{Dataset and Networks}
We use ModelNet40 \cite{modelnet} to train the classifier network ($\mathbf{F}$) and the AE network ($\mathbf{G}$), as well as test our attacks. ModelNet40 contains 12,311 CAD models from 40 different classes. These models are divided into 9,843 for training and 2,468 for testing. Similar to previous work \cite{Deflecting,pcattack,pointdrop}, we sample 1,024 points from each object. We train the $\mathbf{F}$ victim networks: PointNet\cite{pointnet}, PointNet++ in both Single-Scale (SSG) and Multi-scale  (MSG) \cite{pointnet++} settings, and DGCNN \cite{dgcn}. %
For a fair comparison, we adopt the subset of ModelNet40 detailed in \cite{pcattack} to perform and evaluate our attacks against their work (we call this the attack set). In the attack set, 250 examples are chosen from 10 ModelNet40 classes. We train the AE using the full ModelNet40 training set with the Chamfer Distance loss and then fix the AE when the attacks are being generated.  

\mysection{Adversarial Attack Methods}
We compare AdvPC against the state-of-the-art baselines 3D-Adv \cite{pcattack} and KNN Attack \cite{robustshapeattack}. For all attacks, we use Adam optimizer \cite{adam} with learning rate $\eta = 0.01$, and perform 2 different initializations for the optimization of $\boldsymbol{\Delta}$ (as done in \cite{pcattack}). The number of iterations for the attack optimization for all the networks is 200. We set the loss margin $\kappa=30$ in \eqLabel{\ref{eq:adv-loss}} for both 3D-Adv \cite{pcattack} and AdvPC and $\kappa=15$ for KNN Attack \cite{robustshapeattack} (as suggested in their paper). For other hyperparameters of \cite{pcattack,robustshapeattack}, we follow what is reported in their papers. We pick $\gamma =0.25$ in \eqLabel{\ref{eq:final-objective}} for AdvPC because it strikes a balance between the success of the attack and its transferability (refer to \secLabel{\ref{sec:ablation}} for details). In all of the attacks, we follow the same procedure as \cite{pcattack}, where the best attack that satisfies the objective during the optimization is reported. We add the hard $\ell_\infty$ projection $\Pi_{\infty}\left(\boldsymbol{\Delta},\epsilon_{\infty}\right)$ described in \secLabel{\ref{sec:methodology}} to all the methods to ensure fair comparison on the same norm-budget $\epsilon_\infty$. We report the best performance of the baselines obtained under this setup.

\begin{figure}[t]
\tabcolsep=0.03cm
\begin{tabular}{cccc}

\includegraphics[width=0.25\linewidth]{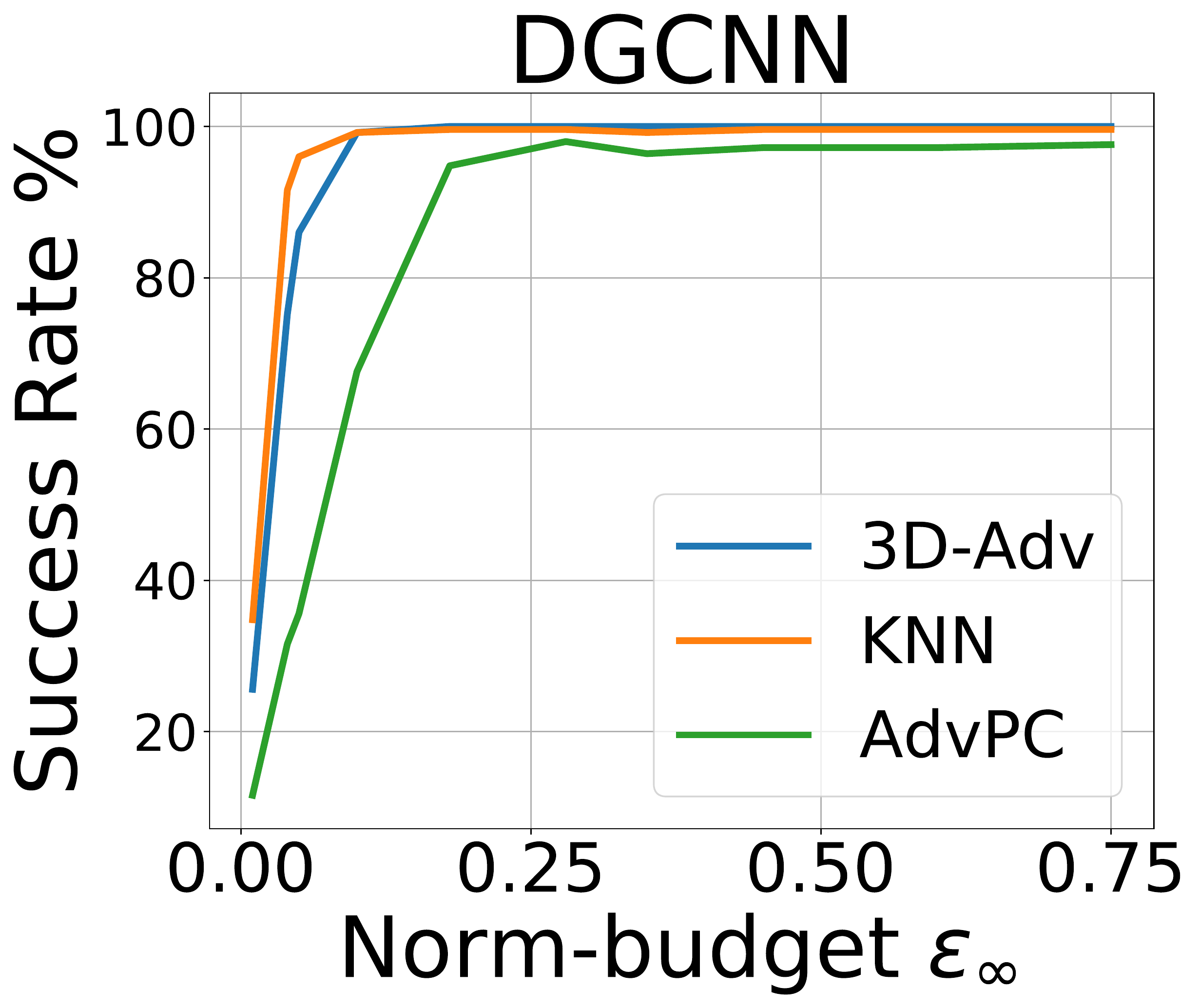} &
 \includegraphics[width=0.25\linewidth]{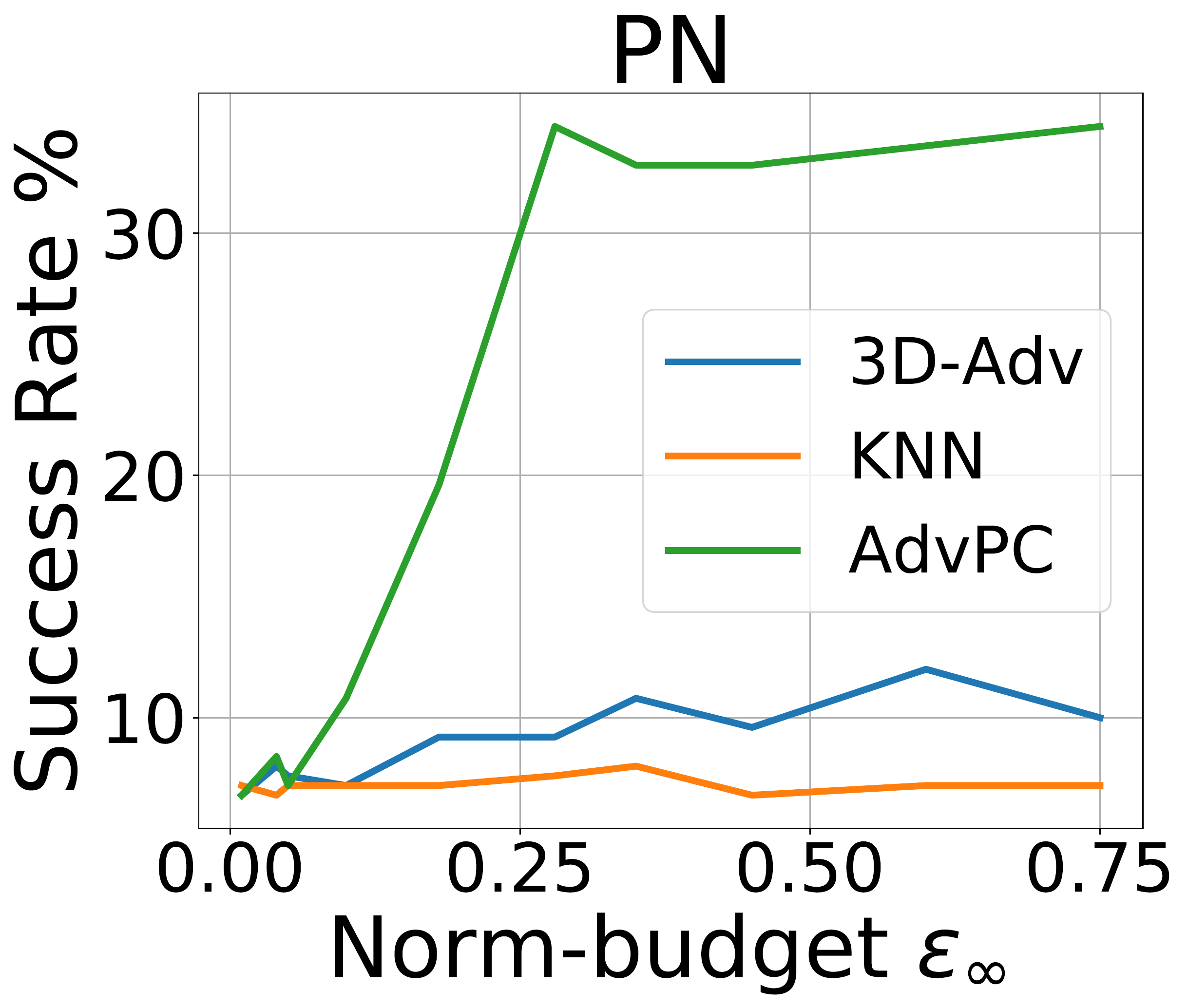} &
\includegraphics[width=0.25\linewidth]{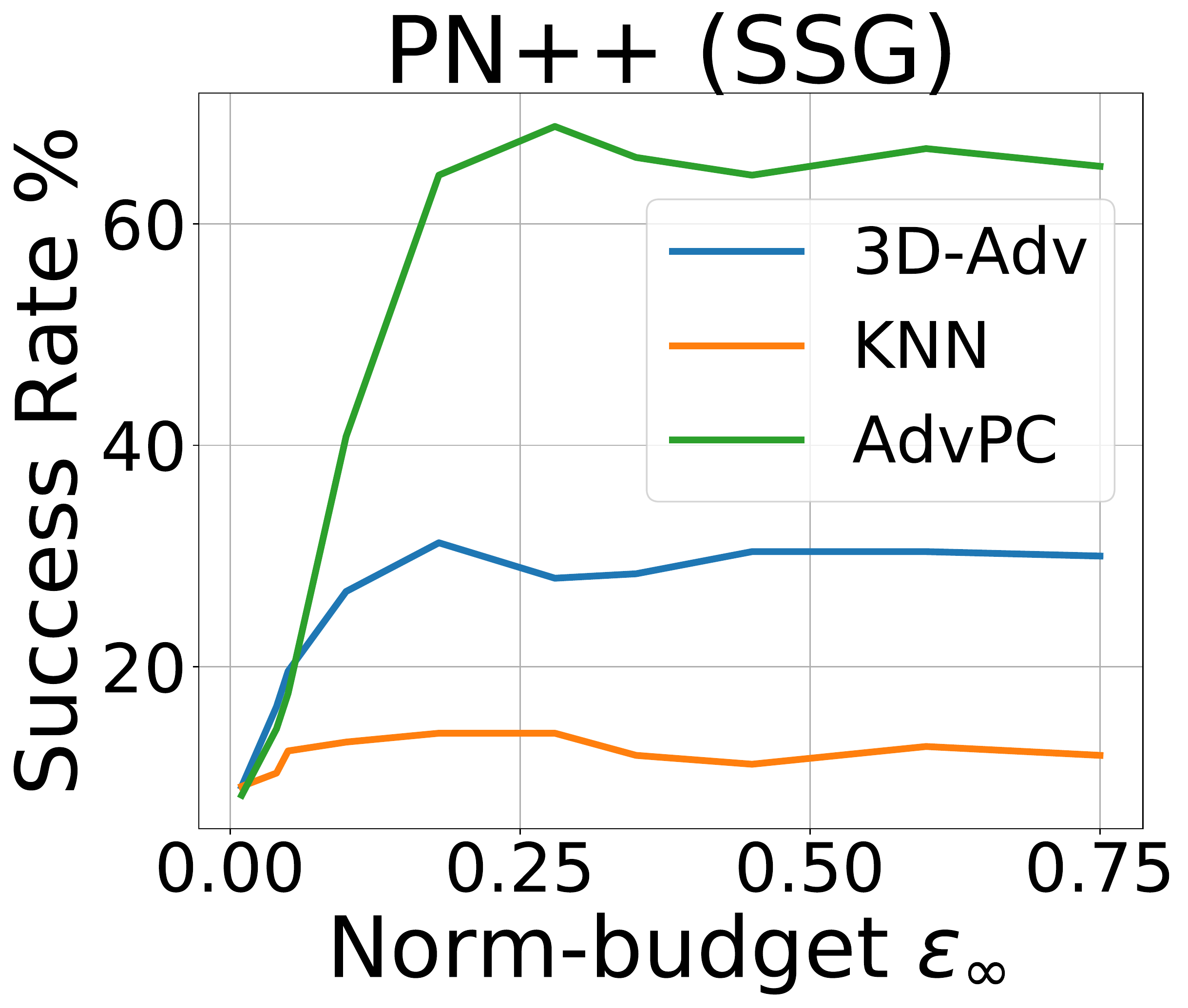} &
\includegraphics[width=0.25\linewidth]{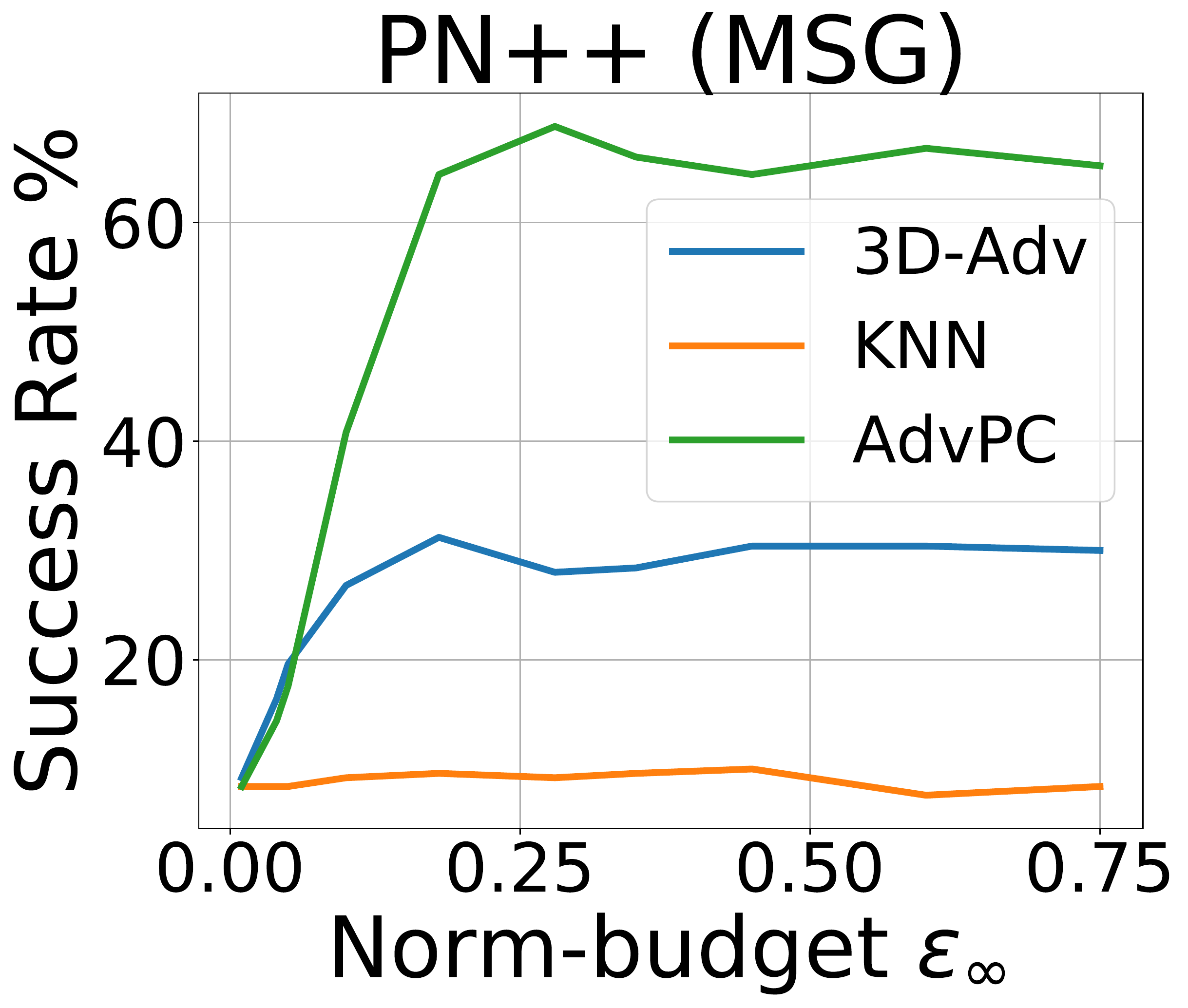} \\

\end{tabular}

\caption{\small \textbf{Transferability Across Different Norm-Budgets}: Here, the victim network is DGCNN \cite{dgcn} and the attacks are optimized using different \textbf{$\epsilon_\infty$} norm-budgets. We report the attack success on 
DGCNN and on the transfer networks (PointNet, PointNet ++ MSG, and PointNet++ SSG). We note that our AdvPC transfers better to the other networks across different  $\epsilon_\infty$ as compared to the baselines 3D-Adv\cite{pcattack} and KNN Attack \cite{robustshapeattack}. Similar plots for the other victim networks are provided in the \supp\!\!.}
\label{fig:transferbility}
\end{figure}

\mysection{Transferability}
We follow the same setup as \cite{learnattack1,learnattack2} by generating attacks using the constrained $\ell_\infty$ metric and measure their success rate at different norm-budgets $\epsilon_\infty$ taken to be in the range $[0,0.75]$. This range is chosen because it enables the attacks to reach 100\% success on the victim network, as well as offer an opportunity for transferability to other networks. We compare AdvPC against the state-of-the-art baselines \cite{pcattack,robustshapeattack} under these norm-budgets (\eg see \figLabel{\ref{fig:transferbility}} for attacking DGCNN). To measure the success of the attack, we compute the percentage of samples out of all attacked samples that the victim network misclassified. We also measure transferability from each victim network to the transfer networks. For each pair of networks, we optimize the attack on one network (victim) and measure the success rate of this optimized attack when applied as input to the other network (transfer). We report these success rates for all network pairs. No defenses are used in the transferability experiment. All the attacks performed in this section are untargeted attacks (following the convention for transferability experiments \cite{pcattack}).

\begin{table}[]
\footnotesize
\setlength{\tabcolsep}{0.5pt} %
\renewcommand{\arraystretch}{1.1} %
\centering
\resizebox{0.992\textwidth}{!}{%
\begin{tabular}{cc|cccc|cccc} 
\toprule
- & - &\multicolumn{4}{c|}{$\epsilon_\infty = 0.18$} &\multicolumn{4}{c}{$\epsilon_\infty = 0.45$}\\
\specialcell{\textbf{Victim}\\\textbf{Network}} & \textbf{Attack}  &\textbf{PN} & \textbf{\specialcell{ PN++ \\(MSG)}} & \textbf{\specialcell{ PN\scriptsize++\\ (SSG)}} & \textbf{DGCNN}  &\textbf{PN} & \textbf{\specialcell{ PN++ \\(MSG)}} & \textbf{\specialcell{ PN++\\ (SSG)}} & \textbf{DGCNN}   \\
\midrule
 \multirow{3}*{\textbf{PN}} & 3D-Adv \cite{pcattack}  &  100 &  8.4 & 10.4 &  6.8 & 100 &  8.8 &  9.6 &  8.0 \\
 & KNN \cite{robustshapeattack} & 100 &  9.6 & 10.8 &  6.0 & 100 &  9.6 &  8.4 &  6.4 \\
 & AdvPC (Ours)  &  98.8 & \textbf{20.4} & \textbf{27.6} & \textbf{22.4} &  98.8 & \textbf{18.0} & \textbf{26.8} & \textbf{20.4} \\ \hline
 \multirow{3}*{\textbf{\specialcell{ PN++ \\(MSG)}}} & 3D-Adv \cite{pcattack} & 6.8 & 100 & 28.4 & 11.2 &  7.2 & 100 & 29.2 & 11.2 \\ & KNN \cite{robustshapeattack} & 6.4 & 100 & 22.0 &  8.8 &  6.4 & 100 & 23.2 &  7.6 \\ & AdvPC (Ours)  & \textbf{13.2} &  97.2 & \textbf{54.8} & \textbf{39.6} & \textbf{18.4} &  98.0 & \textbf{58.0} & \textbf{39.2} \\ \hline
 \multirow{3}*{\textbf{\specialcell{ PN++\\ (SSG)}}} &  3D-Adv \cite{pcattack}  & 7.6 &  9.6 & 100 &  6.0 &  7.2 & 10.4 & 100 &  7.2 \\ & KNN \cite{robustshapeattack} &  6.4 &  9.2 & 100 &  6.4 &  6.8 &  7.6 & 100 &  6.0 \\ & AdvPC (Ours)  & \textbf{12.0} & \textbf{27.2} &  99.2 & \textbf{22.8} & \textbf{14.0} & \textbf{30.8} &  99.2 & \textbf{27.6} \\ \hline
 \multirow{3}*{\textbf{DGCNN}} &  3D-Adv \cite{pcattack}  & 9.2 & 11.2 & 31.2 & 100 &  9.6 & 12.8 & 30.4 & 100 \\ & KNN \cite{robustshapeattack} & 7.2 &  9.6 & 14.0 &  99.6 &  6.8 & 10.0 & 11.2 &  99.6 \\ & AdvPC (Ours)  & \textbf{19.6} & \textbf{46.0} & \textbf{64.4} &  94.8 & \textbf{32.8} & \textbf{48.8} & \textbf{64.4} &  97.2 \\  
 \bottomrule
\end{tabular}}
\caption{\small \textbf{Transferability of Attacks}: We use norm-budgets (max $\ell_\infty$ norm allowed in the perturbation) of $\epsilon_\infty = 0.18$ and $\epsilon_\infty = 0.45$ . All the reported results are the untargeted Attack Success Rate (higher numbers are better attacks). \textbf{Bold} numbers indicate the most transferable attacks. Our attack consistently achieves better transferability than the other attacks for all networks, especially on DGCNN \cite{dgcn}. For reference, the classification accuracies on unperturbed samples for networks PN, PN++(MSG), PN++(SSG) and DGCNN are  92.8\%, 91.5\%, 91.5\%, and 93.7\%, respectively.
}
\label{tbl:transfer}
\end{table}
\begin{figure}[t]
\tabcolsep=0.03cm
\begin{tabular}{ccc}

\includegraphics[width=0.33\linewidth]{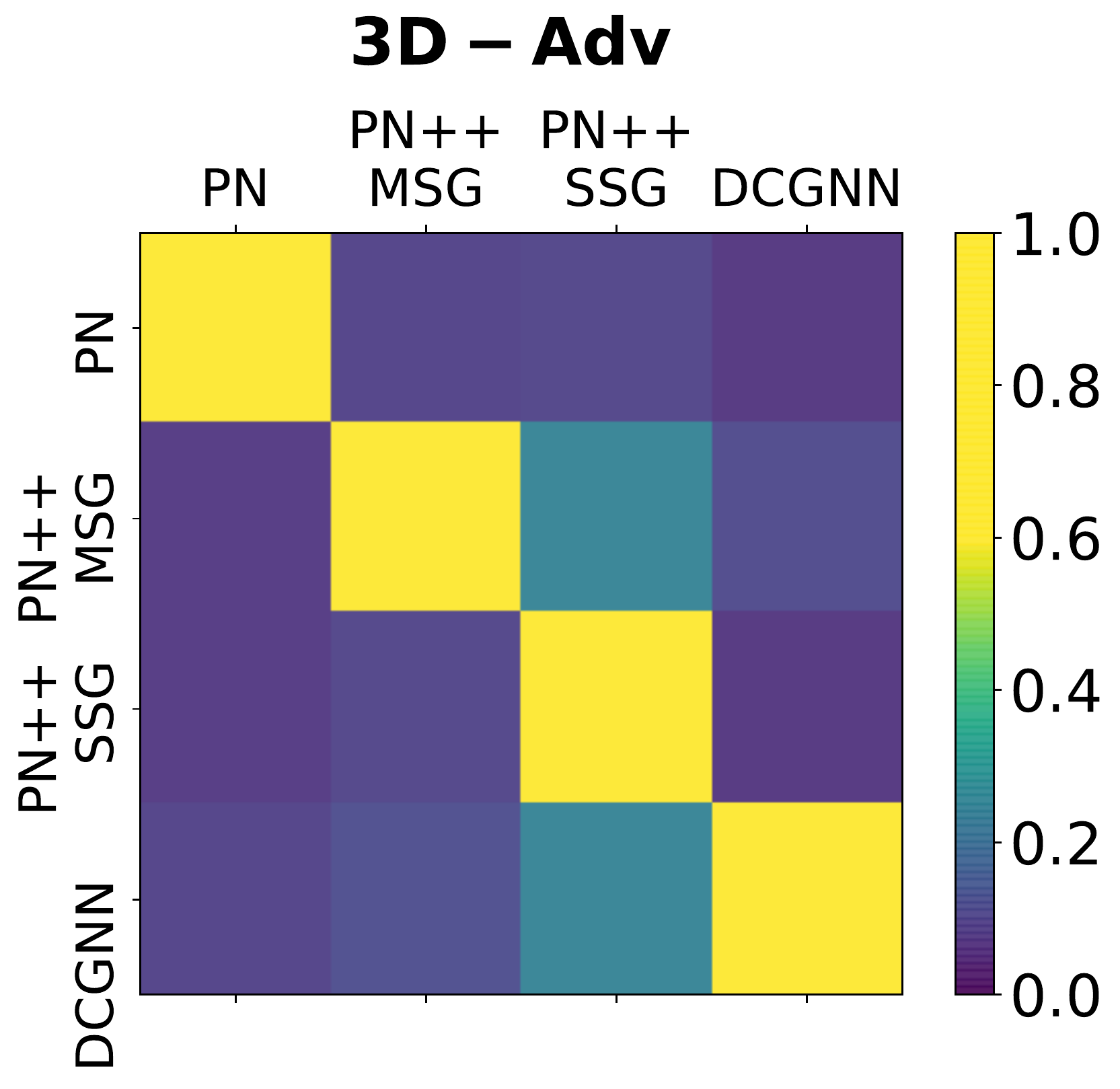} &
\includegraphics[width=0.33\linewidth]{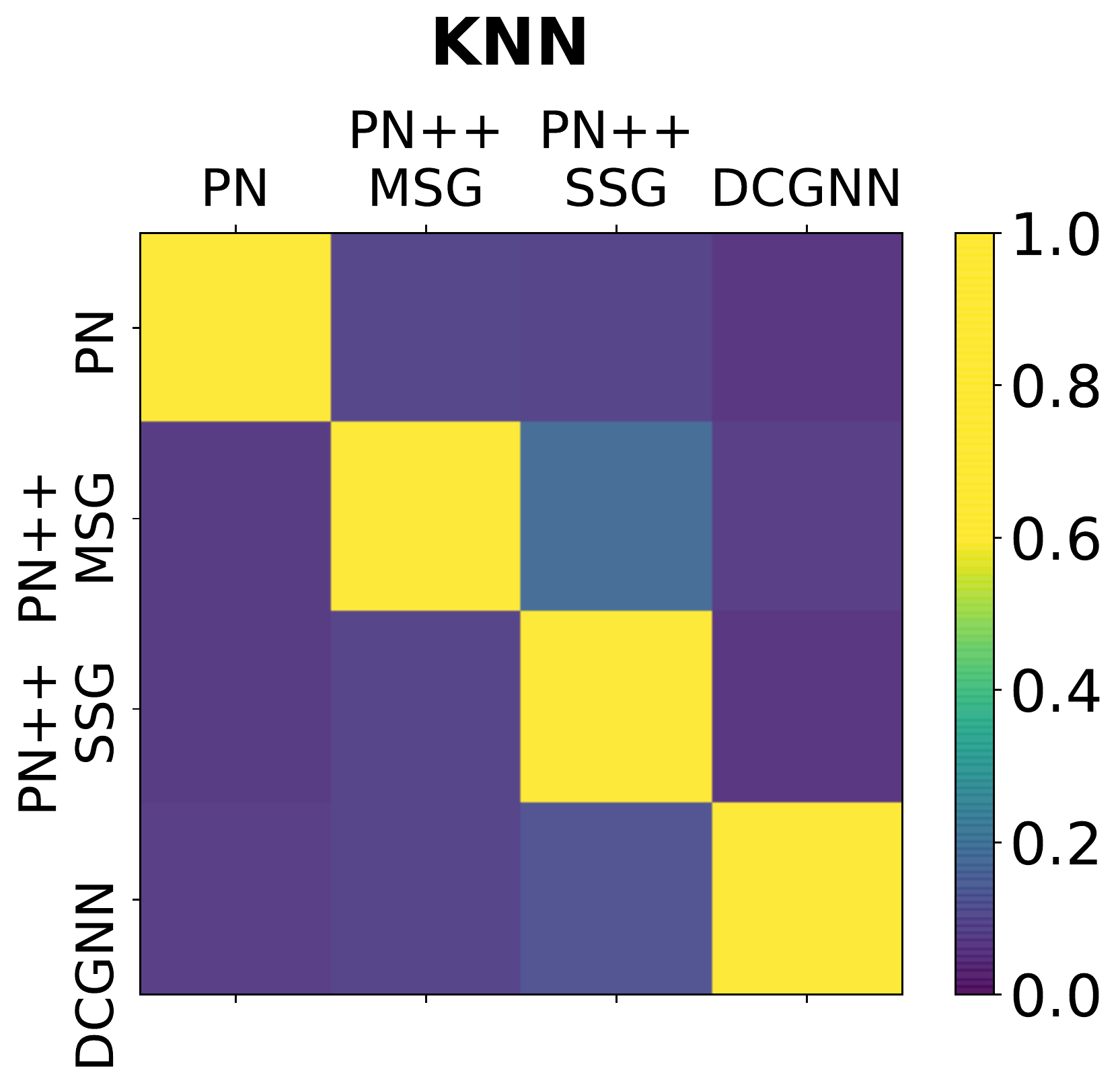} &
\includegraphics[width=0.33\linewidth]{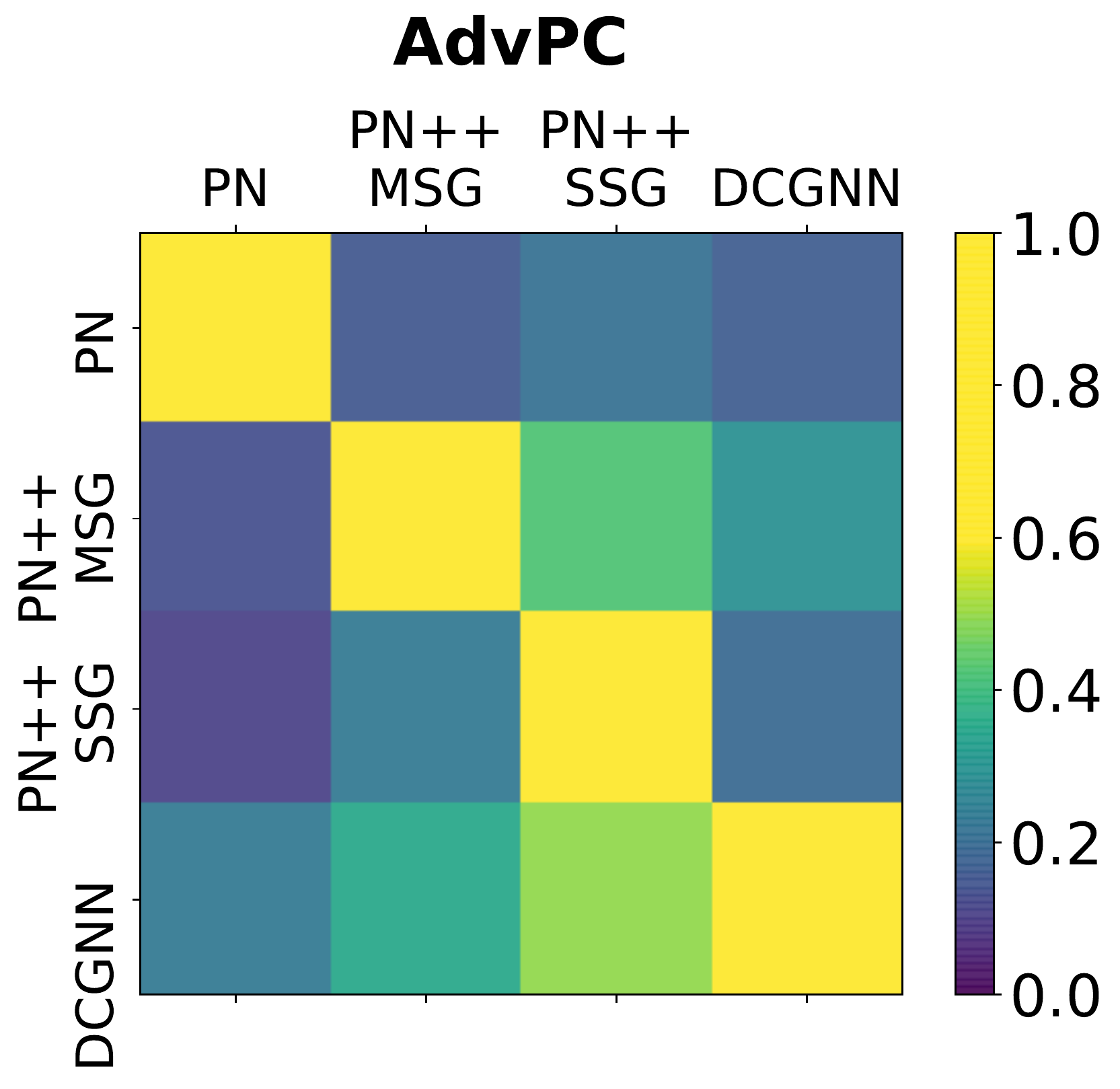}  \\
transferability: 11.5 \% &
transferability: 8.92 \% &
transferability: \textbf{24.9} \% \\
\end{tabular}

\caption{\small \textbf{Transferability Matrix}: Visualizing the overall transferability for 3D-Adv \cite{pcattack} (\textit{left}), KNN Attack \cite{robustshapeattack} (\textit{middle}), and our AdvPC (\textit{right}). Elements in the same row correspond to the same victim network used in the attack, while those in the same column correspond to the network that the attack is transferred to. Each matrix element measures the average success rate over the range of $\epsilon_\infty$ for the transfer network. We expect the diagonal elements of each transferability matrix (average success rate on the victim network) to have high values, since each attack is optimized on the same network it is transferred to. More importantly, brighter off-diagonal matrix elements indicate better transferability. 
We observe that our proposed AdvPC attack is more transferable than the other attacks and that DGCNN is a more transferable victim network than the other point cloud networks. 
The transferability score under each matrix is the average of the off-diagonal matrix values, which summarizes overall transferability for an attack.
}
\label{fig:transmatrix}
\end{figure}

\mysection{Attacking the Defenses}
We also analyze the success of our attacks against point cloud defenses. We compare AdvPC attacks and the baselines \cite{pcattack,robustshapeattack} against several defenses used in the point cloud literature: SOR, SRS, DUP-Net \cite{Deflecting}, and Adversarial Training \cite{pcattack}. We also add a newly trained AE (different from the one used in the AdvPC attack) to this list of defenses. %
For SRS, we use a drop rate of 10\%, while in SOR, we use the same parameters proposed in \cite{Deflecting}. We train DUP-Net on ModelNet40 with an up-sampling rate of 2. For Adversarial Training, all four networks are trained using a mix of the training data of ModelNet40 and adversarial attacks generated by \cite{pcattack}. While these experiments are for untargeted attacks, we perform similar experiments under targeted attacks and report the results in \supp for reference and completeness.

\subsection{Results}  \label{sec:results}
We present quantitative results that focus on two main aspects. First, we show the transferable power of AdvPC attacks to different point cloud networks. Second, we highlight the strength of AdvPC under different point cloud defenses.

\mysection{Transferability}
Table \ref{tbl:transfer} reports transferability results for $\epsilon_\infty = 0.18$ and $\epsilon_\infty = 0.45$ %
and compares AdvPC with the baselines \cite{pcattack,robustshapeattack}. The value $\epsilon_\infty = 0.18$ is chosen, since it allows the DGCNN attack to reach maximum success (see \secLabel{\ref{sec:sens}}), and the value $\epsilon_\infty = 0.45$ is arbitrarily chosen to be midway in the remaining range of $\epsilon_\infty$. It is clear that AdvPC attacks consistently beat the baselines when transferring between networks (up to 40\%). Our method shows substantial gains in the case of DGCNN. We also report transferability results for a range of $\epsilon_\infty$ values in \figLabel{\ref{fig:transferbility}} when the victim network is DGCNN, and the attacks transferred to all other networks. In \supp\!\!, we show the same plots when the victim network is taken to be PN and PN++. To represent all these transferability curves compactly, 
we aggregate their results into a Transferability Matrix. Every entry in this matrix measures the transferability from the victim network (\textbf{row}) to the transfer network (\textbf{column}), and it is computed as the average success rate of the attack evaluated on the transfer network across all $\epsilon_\infty$ values. This value reflects how good the perturbation is at fooling the transfer network overall. As such, we advocate the use of the transferability matrix as a standard mode of evaluation for future work on network-transferable attacks. In \figLabel{\ref{fig:transmatrix}}, we show the transferability matrices for our attack and the baselines. AdvPC transfers better overall, since it leads to higher (brighter) off-diagonal values in the matrix. %
Using the average of off-diagonal elements in this matrix as a single scalar measure of transferability, AdvPC achieves 24.9\% average transferability, as compared to 11.5\% for 3D-Adv \cite{pcattack} and 8.92\% for KNN Attack \cite{robustshapeattack}. We note that DGCNN \cite{dgcn} performs best in terms of transferability and is the hardest network to attack (for AdvPC and the baselines).  

\mysection{Attacking Defenses}
Since DGCNN performs the best in transferability, we use it to evaluate the resilience of our AdvPC attacks under different defenses. We use the five defenses described in \secLabel{\ref{sec:setup}} and report their results in Table \ref{tbl:breaking}. Our attack is more resilient than the baselines against all defenses. We note that the AE defense is very strong against all attacks compared to other defenses \cite{Deflecting}, which explains why AdvPC works very well against other defenses and transfers well to unseen networks. We also observe that our attack is strong against simple statistical defenses like \textbf{SRS} (38\% improvement over the baselines). We report results for other victim networks (PN and PN++) in the \supp, where AdvPC shows superior performance against the baselines under these defenses. 
\begin{table}[t]
\footnotesize
\centering
\setlength{\tabcolsep}{4pt} %
\renewcommand{\arraystretch}{1} %
\begin{tabular}{c|ccc|ccc} 
\toprule
 & \multicolumn{3}{c}{$\epsilon_\infty = 0.18$} & \multicolumn{3}{c}{$\epsilon_\infty = 0.45$} \\
\textbf{Defenses} & \textbf{\specialcell{3D-Adv\\ \cite{pcattack} }} & \textbf{\specialcell{KNN \\ \cite{robustshapeattack}}} & \textbf{\specialcell{AdvPC \\(ours)}}  & \textbf{\specialcell{3D-Adv \\\cite{pcattack}}} & \textbf{\specialcell{KNN \\ \cite{robustshapeattack}}} &  \textbf{\specialcell{AdvPC \\(ours)}}    \\
\midrule
No defense & \textbf{100} & 99.6 & 94.8  & \textbf{100} & 99.6 &  97.2\\ 
AE (newly trained) & 9.2 & 10.0 & \textbf{17.2} & 12.0 & 10.0 & \textbf{21.2} \\
Adv Training \cite{pcattack} & 7.2 &  7.6 & \textbf{39.6} &  8.8 &  7.2 & \textbf{42.4} \\
SOR \cite{Deflecting} &18.8 & 17.2 & \textbf{36.8} & 19.2 & 19.2 & \textbf{32.0} \\
DUP Net \cite{Deflecting}  & 28 & 28.8 &\textbf{43.6} & 28 & 31.2 & \textbf{37.2}\\ 
SRS \cite{Deflecting} & 43.2 & 29.2 & \textbf{80.0} & 47.6 & 31.2 & \textbf{85.6} \\
 \bottomrule
\end{tabular}
\caption{\small \textbf{Attacking Point Cloud Defenses:} We evaluate untargeted attacks using norm-budgets of $\epsilon_\infty = 0.18$ and $\epsilon_\infty = 0.45$ with DGCNN \cite{dgcn} as the victim network under different defenses for 3D point clouds. Similar to before, we report attack success rates (\textbf{higher} indicates better attack). AdvPC consistently outperforms the other attacks \cite{pcattack,robustshapeattack} for all defenses. %
Note that both the attacks \textit{and} evaluations are performed on DGCNN, which has an accuracy of 93.7\% without input perturbations (for reference).
}
\label{tbl:breaking}
\end{table}
\section{Analysis} \label{sec:analysis}
We perform several analytical experiments to further explore the results obtained in \secLabel{\ref{sec:results}}. We first study the effect of different factors that play a role in the %
transferability of our attacks. We also show some interesting insights related to the sensitivity of point cloud networks and the effect of the AE on the attacks.

\subsection{Ablation Study (hyperparameter $\gamma$)} \label{sec:ablation}
\begin{figure}[t]
\tabcolsep=0.03cm
\begin{tabular}{cc}
\includegraphics[width=0.5\columnwidth]{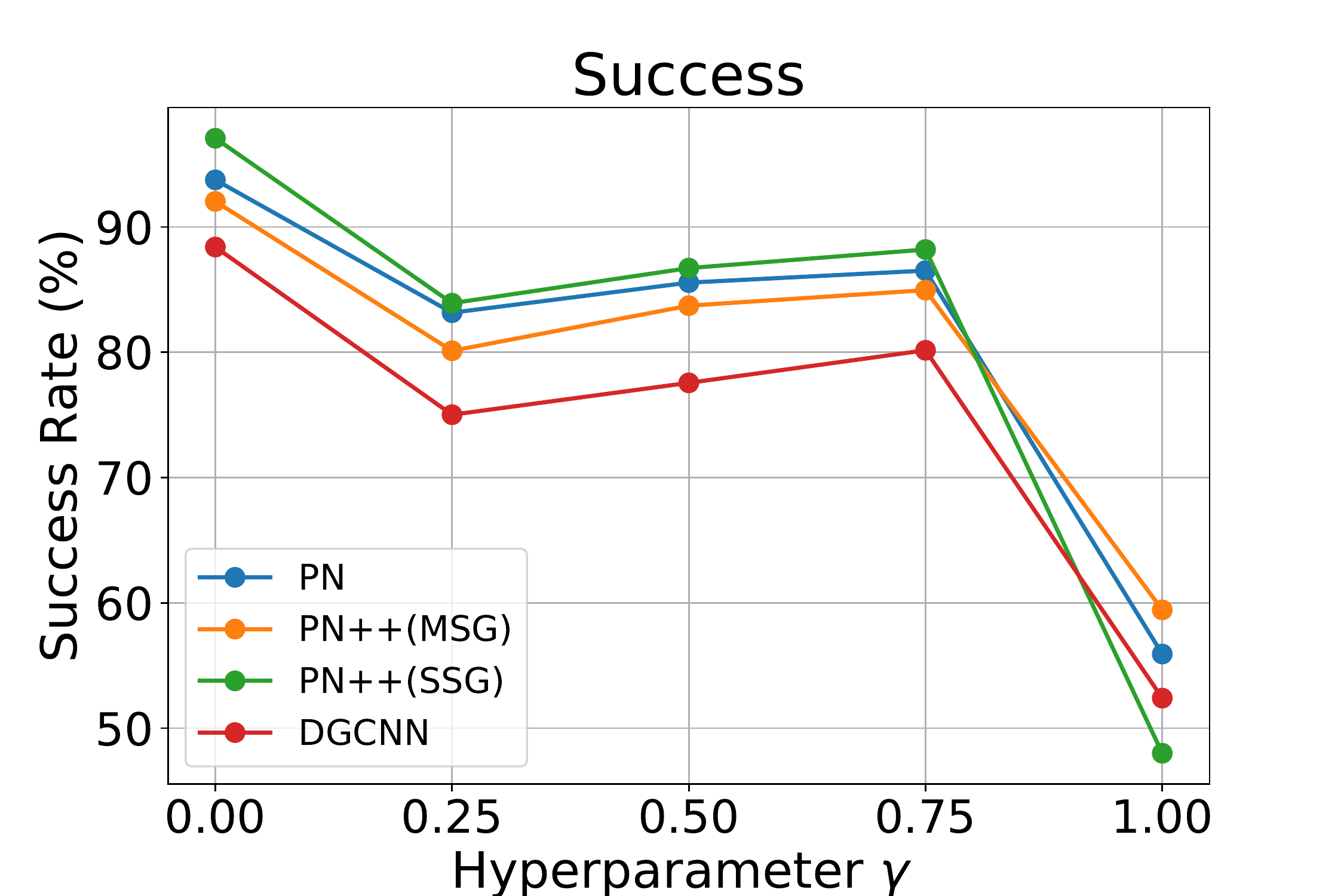} &
\includegraphics[width=0.5\columnwidth]{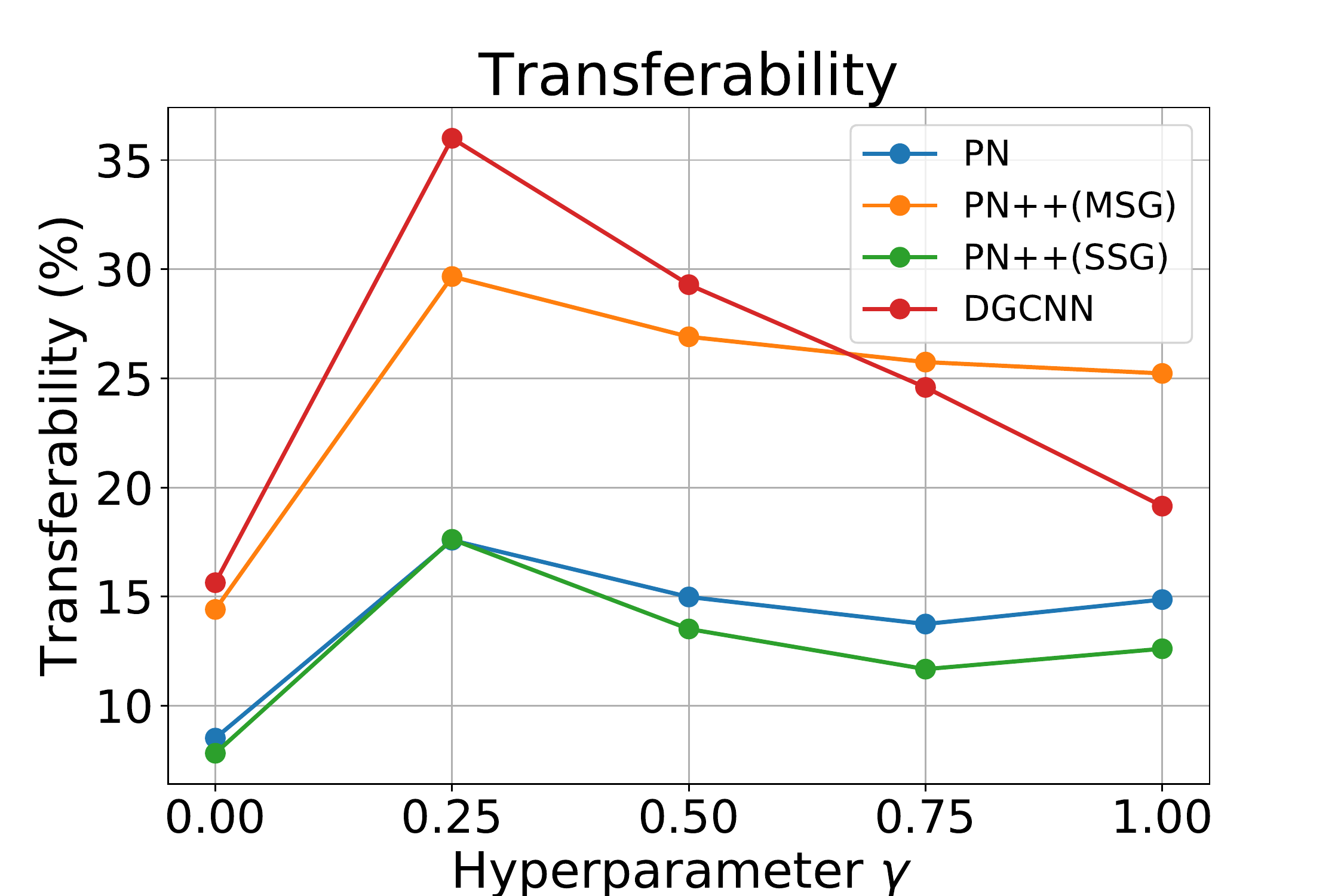} \\
\end{tabular}
\caption{\small \textbf{Ablation Study}: Studying the effect of changing AdvPC hyperparameter ($\gamma$) on the success rate of the attack (\textit{left}) and on its transferability (\textit{right}). The transferability score reported for each victim network is the average success rate on the transfer networks averaged across all different norm-budgets $\epsilon_\infty$. We note that as $\gamma$ increases, the success rate of the attack on the victim network drops, and the transferability varies with $\gamma$. We pick $\gamma=0.25$ in all of our experiments.}
\label{fig:gamma}
\end{figure}

Here, we study the effect of $\gamma$ used in \eqLabel{\ref{eq:final-objective}} on the performance of our attacks. 
While varying $\gamma$ between 0 and 1, we record the attack success rate on the victim network and report the transferability to all of the other three transfer networks (average success rate on the transfer networks). We present averaged results over all norm-budgets in \figLabel{\ref{fig:gamma}} for the four victim networks. %
One observation is that adding the AE loss with $\gamma>0$ tends to deteriorate the success rate, even though it improves transferability. We pick $\gamma=0.25$ in our experiments to balance success and transferability.   
\subsection{Network Sensitivity to Point Cloud Attacks} \label{sec:sens}
\begin{figure}[]
\tabcolsep=0.03cm
\begin{tabular}{ccc}
\includegraphics[width=0.33\columnwidth]{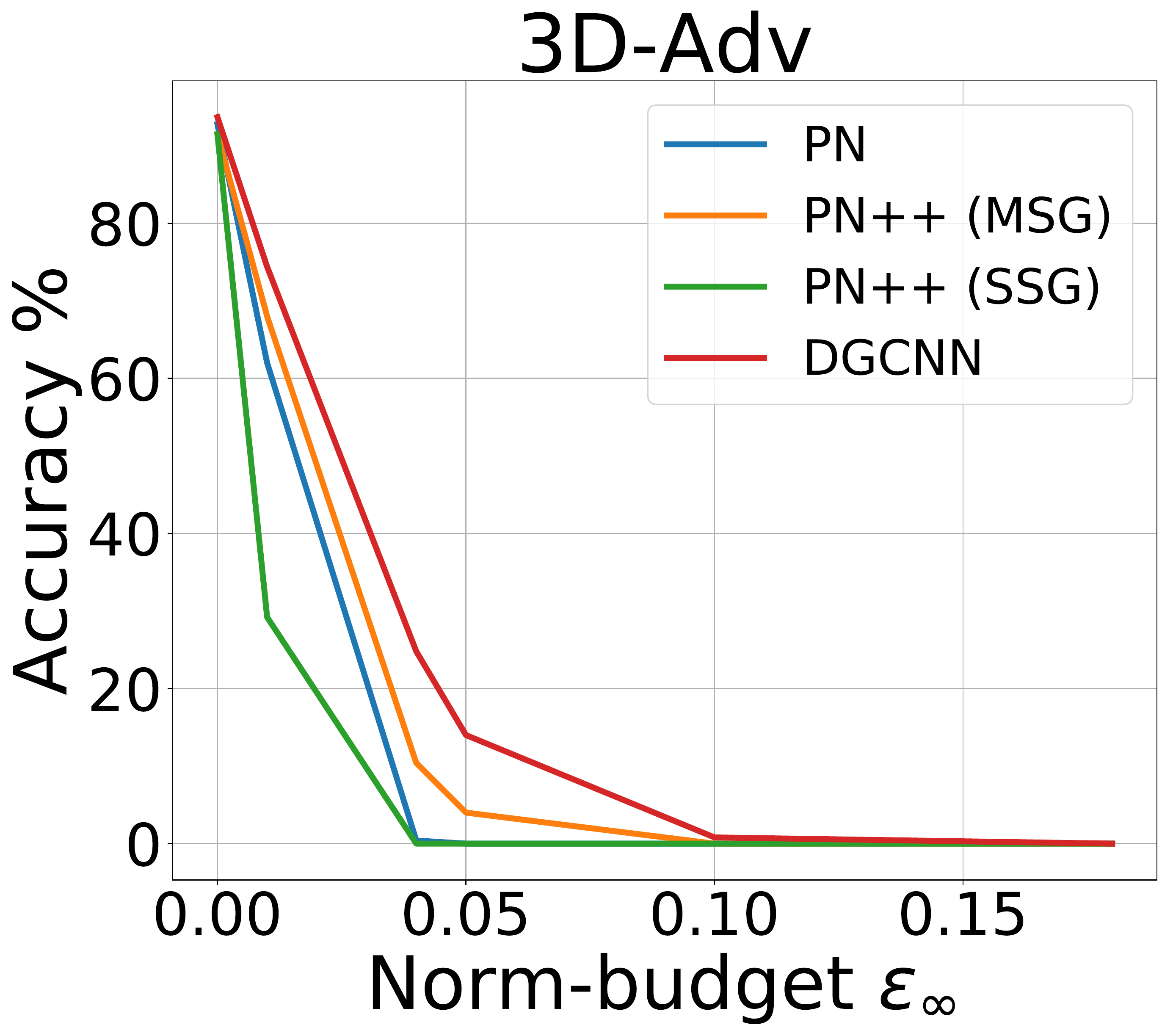} &
\includegraphics[width=0.33\columnwidth]{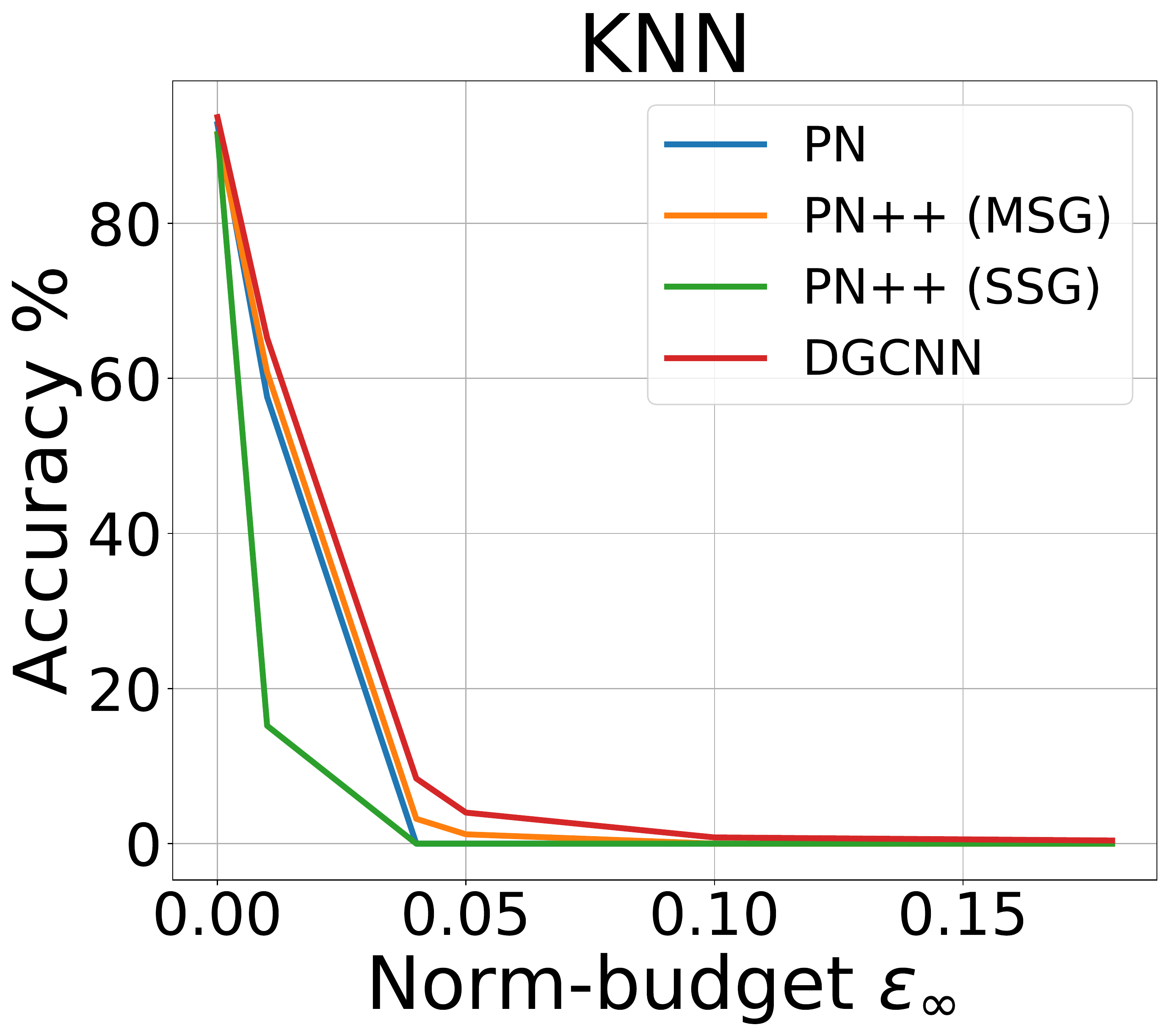} &
\includegraphics[width=0.33\columnwidth]{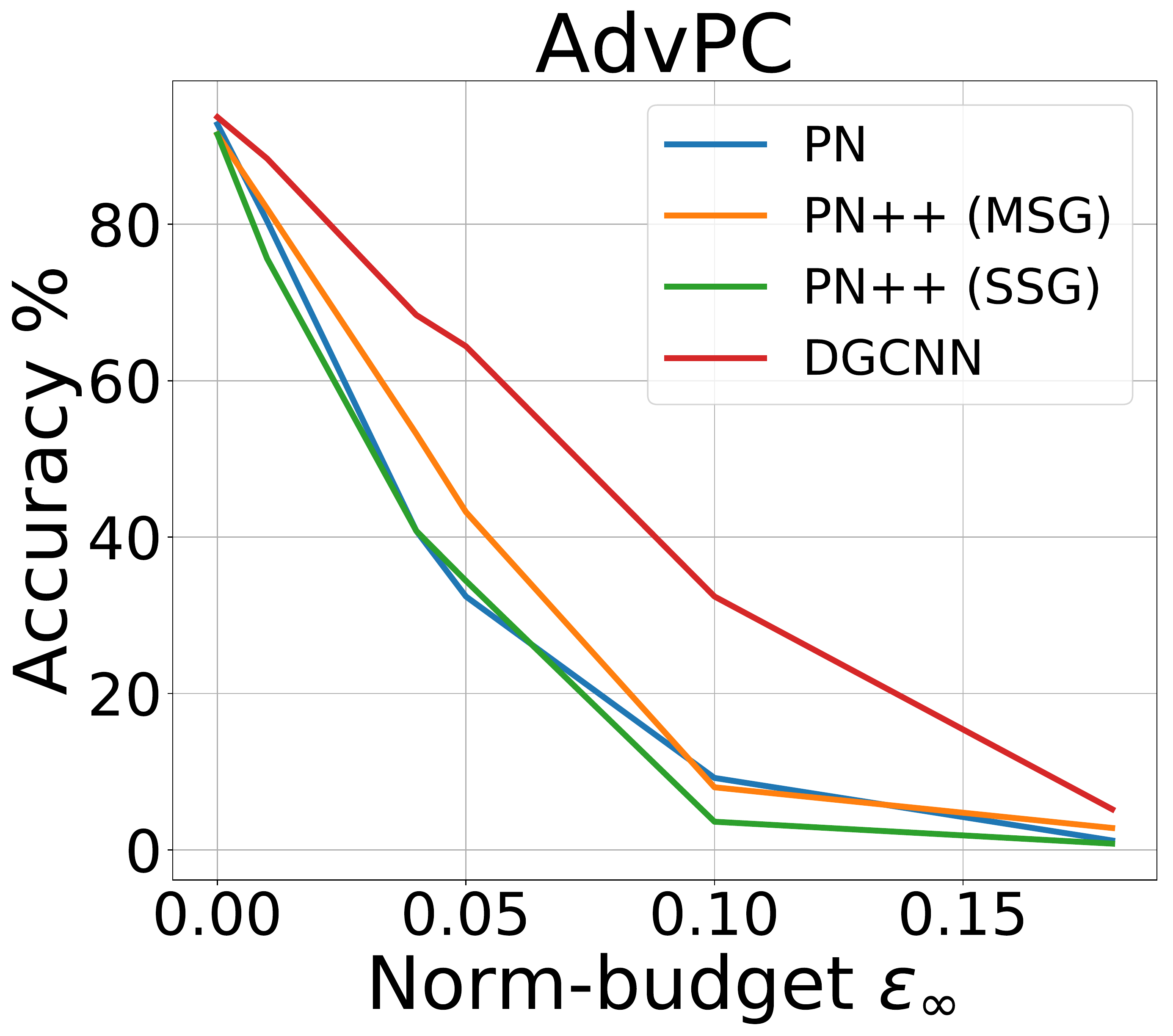} \\
\end{tabular}

\caption{\small \textbf{Sensitivity of Architectures}: We evaluate the sensitivity of each of the four networks for increasing norm-budget. For each network, we plot the classification accuracy under 3D-Adv perturbation \cite{pcattack} (\textit{left}), KNN Attack \cite{robustshapeattack} (\textit{middle}), and our AdvPC attack (\textit{right}). Overall, DGCNN \cite{dgcn} is affected the least by adversarial perturbation.}
\label{fig::sensitivity}
\end{figure}

\figLabel{\ref{fig::sensitivity}} plots the sensitivity of the various networks when they are subject to input perturbations of varying norm-budgets $\epsilon_\infty$. We measure the classification accuracy of each  network under our AdvPC attack ($\gamma=0.25$), 3D-Adv \cite{pcattack}, and KNN Attack \cite{robustshapeattack}. We observe that 
DGCNN \cite{dgcn} tends to be the most robust to adversarial perturbations in general. This might be explained by the fact that the convolution neighborhoods in DGCNN are dynamically updated across layers and iterations. This dynamic behavior in network structure may hinder the effect of the attack because gradient directions can change significantly from one iteration to another. This leads to failing attacks and higher robustness for DGCNN \cite{dgcn}.

\subsection{Effect of the Auto-Encoder (AE)} \label{sec:intrep}
\begin{figure}[t]
\tabcolsep=0.03cm
\resizebox{\textwidth}{!}{%
\begin{tabular}{cc|cc|cc|cc}
\toprule
\multicolumn{2}{c}{\specialcell{unperturbed \\ point cloud}}  & \multicolumn{2}{c}{3D-adv \cite{pcattack}} & \multicolumn{2}{c}{ KNN \cite{robustshapeattack}} & \multicolumn{2}{c}{AdvPC (ours)} \\ \hline
before AE & after AE & before AE & after AE  & before AE & after AE   & before AE & after AE \\ 
\includegraphics[trim={1cm 0 1.1cm 0},clip,width = 0.124\columnwidth]{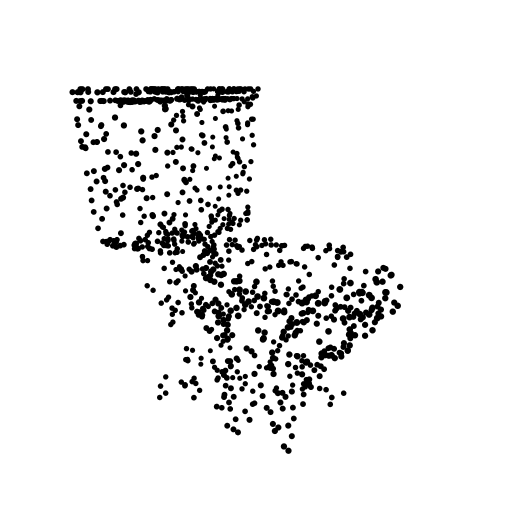} &
\includegraphics[trim={1cm 0 1.1cm 0},clip,width = 0.124\columnwidth]{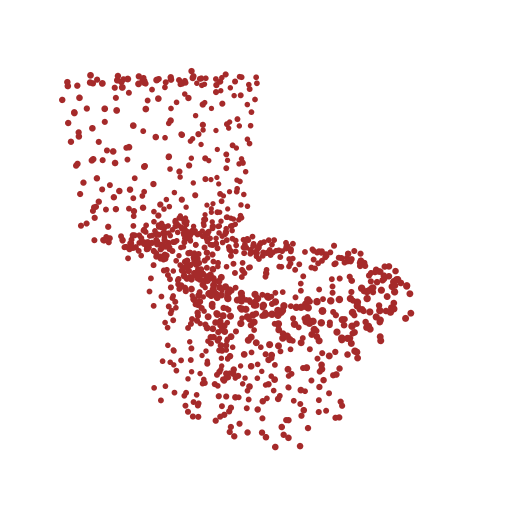} &
\includegraphics[trim={1cm 0 1.1cm 0},clip,width = 0.124\columnwidth]{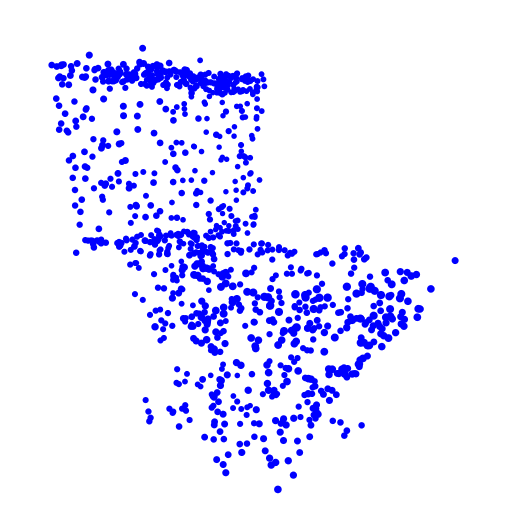} &
\includegraphics[trim={1cm 0 1.1cm 0},clip,width = 0.124\columnwidth]{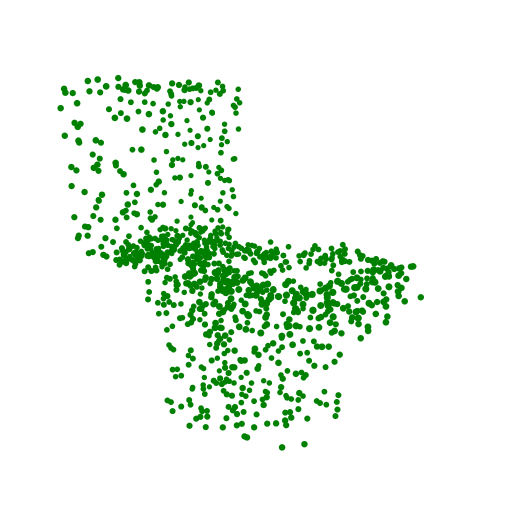} &
\includegraphics[trim={1cm 0 1.1cm 0},clip,width = 0.124\columnwidth]{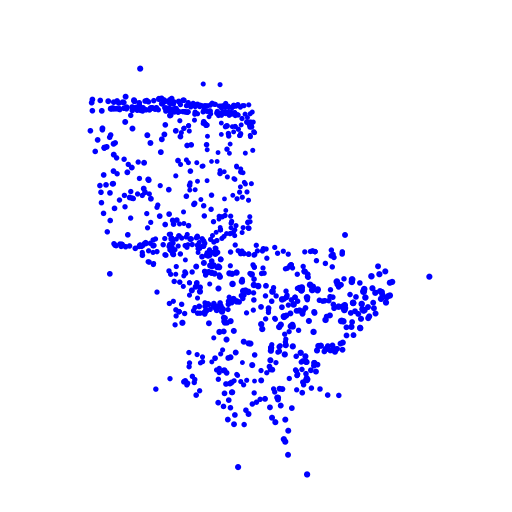} &
\includegraphics[trim={1cm 0 1.1cm 0},clip,width = 0.120\columnwidth]{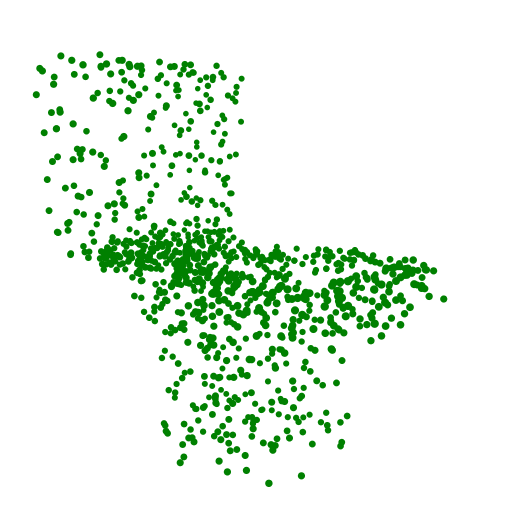} &
\includegraphics[trim={1cm 0 1.1cm 0},clip,width = 0.124\columnwidth]{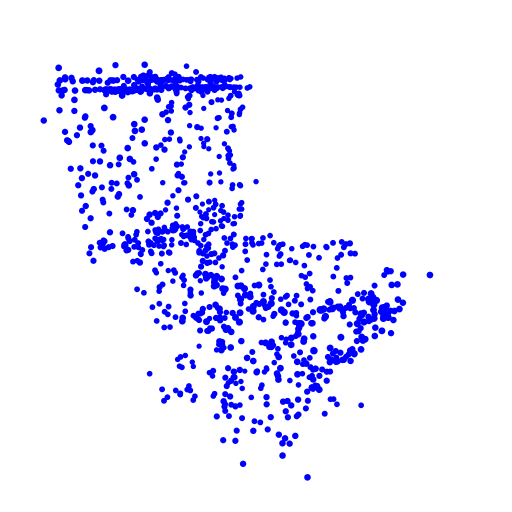} &
\includegraphics[trim={1cm 0 1.1cm 0},clip,width = 0.124\columnwidth]{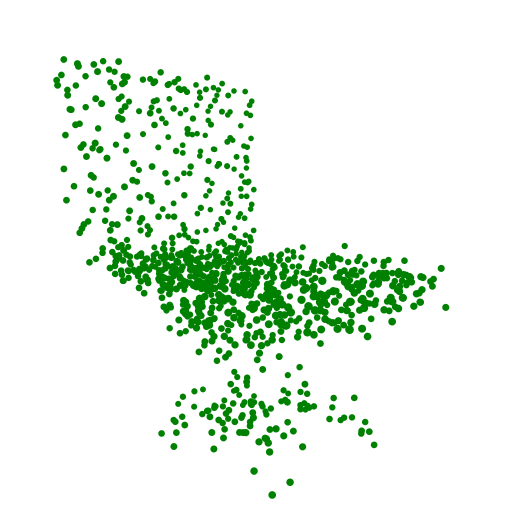} \\
PN: & PN: &PN:  & PN: &PN:  & PN:  &PN:  & PN: \\
\textbf{Toilet} \color{ForestGreen}\cmark & \textbf{Toilet} \color{ForestGreen}\cmark 
 & \textbf{Bed} \color{red}\xmark  & \textbf{Toilet} \color{ForestGreen}\cmark & \textbf{Bed} \color{red}\xmark  & \textbf{Toilet} \color{ForestGreen}\cmark & \textbf{Bed}  \color{red}\xmark  & \textbf{Chair}  \color{red}\xmark  \\
 \bottomrule
\end{tabular}
}
\caption{\small \textbf{Effect of the Auto-Encoder (AE):} The AE does not affect the unperturbed point cloud (classified correctly by PN before and after AE). The AE cleans the  point cloud perturbed by 3D-Adv and KNN \cite{pcattack,robustshapeattack}, which allows PN to predict the correct class label. However, our AdvPC attack can fool PN before and after AE reconstruction. Samples perturbed by AdvPC, if passed through the AE, transform into similar looking objects from different classes (Chair looks similar to Toilet).
}
\label{fig:comparisontoilet}
\end{figure}

In \figLabel{\ref{fig:comparisontoilet}}, we show an example of how AE reconstruction preserves the details of the unperturbed point cloud and does not change the classifier prediction. When a perturbed point cloud passes through the AE, it recovers a natural-looking shape. The AE's ability to reconstruct natural-looking 3D point clouds from various perturbed inputs might explain why it is a strong defense against attacks in Table \ref{tbl:breaking}. Another observation from \figLabel{\ref{fig:comparisontoilet}} is that: when we fix the target $t^\prime$ and do not enforce a specific incorrect target $t^{\prime\prime}$ (\ie untargeted attack setting) for the data adversarial loss on the reconstructed point cloud  in the AdvPC attack (\eqLabel{\ref{eq:final-objective}}), the optimization mechanism tends to pick $t^{\prime\prime}$ to be a \textit{similar} class to the correct one. For example, a \emph{Toilet} point cloud perturbed by AdvPC can be transformed into a \emph{Chair} (similar in appearance to a toilet), if reconstructed by the AE. %
This effect is not observed for the other attacks \cite{pcattack,robustshapeattack}, which do not consider the data distribution and optimize solely for the network.  %
For completeness, we tried replacing the AE with other 3D generative models from \cite{pc-ae} in our AdvPC attack, and we tried to use the learning approach in \cite{learnattack1,learnattack2} instead of optimization, but the attack success was less than satisfactory in both cases (refer to \supp\!\!).

\section{Conclusions} \label{sec:conclusion}
In this paper, we propose a new adversarial attack for 3D point clouds that utilizes a data adversarial loss to formulate network-transferable perturbations. Our attacks achieve better transferability to four popular point cloud networks than other 3D attacks, and they improve robustness against popular defenses. Future work would extend this attack to other 3D deep learning tasks, such as detection and segmentation, and integrate it into a robust training framework for point cloud networks. %

\mysection{Acknowledgments} This work was supported by the King Abdullah University of Science and Technology (KAUST) Office of Sponsored Research under Award No. RGC/3/3570-01-01.

\bibliographystyle{splncs04}
\bibliography{egbib}

\begin{thebibliography}{10}
\providecommand{\url}[1]{\texttt{#1}}
\providecommand{\urlprefix}{URL }
\providecommand{\doi}[1]{https://doi.org/#1}

\bibitem{pc-ae}
Achlioptas, P., Diamanti, O., Mitliagkas, I., Guibas, L.: Learning
  representations and generative models for 3d point clouds. International
  Conference on Machine Learning (ICML)  (2018)

\bibitem{strike}
Alcorn, M.A., Li, Q., Gong, Z., Wang, C., Mai, L., Ku, W.S., Nguyen, A.: Strike
  (with) a pose: Neural networks are easily fooled by strange poses of familiar
  objects. In: The IEEE Conference on Computer Vision and Pattern Recognition
  (CVPR) (2019)

\bibitem{lidar-adv}
Cao, Y., Xiao, C., Yang, D., Fang, J., Yang, R., Liu, M., Li, B.: Adversarial
  objects against lidar-based autonomous driving systems. CoRR
  \textbf{abs/1907.05418} (2019)

\bibitem{carlini}
Carlini, N., Wagner, D.: Towards evaluating the robustness of neural networks.
  In: IEEE Symposium on Security and Privacy (SP) (2017)

\bibitem{pc_engelmann2018}
{Engelmann}, F., {Kontogianni}, T., {Hermans}, A., {Leibe}, B.: Exploring
  spatial context for 3d semantic segmentation of point clouds. In: 2017 IEEE
  International Conference on Computer Vision Workshops (ICCVW). pp. 716--724
  (Oct 2017)

\bibitem{fast-sign}
Goodfellow, I., Shlens, J., Szegedy, C.: Explaining and harnessing adversarial
  examples. In: International Conference on Learning Representations (ICLR)
  (2015)

\bibitem{semantic-robustness}
Hamdi, A., Ghanem, B.: Towards analyzing semantic robustness of deep neural
  networks. CoRR  \textbf{abs/1904.04621} (2019)

\bibitem{sada}
Hamdi, A., Muller, M., Ghanem, B.: {SADA:} semantic adversarial diagnostic
  attacks for autonomous applications. In: AAAI Conference on Artificial
  Intelligence (2020)

\bibitem{pc_huang2018recurrent}
Huang, Q., Wang, W., Neumann, U.: Recurrent slice networks for 3d segmentation
  of point clouds. In: Proceedings of the IEEE Conference on Computer Vision
  and Pattern Recognition (CVPR). pp. 2626--2635 (2018)

\bibitem{adam}
Kingma, D.P., Ba, J.: Adam: {A} method for stochastic optimization. CoRR
  \textbf{abs/1412.6980} (2014)

\bibitem{projected-gradient}
Kurakin, A., Goodfellow, I.J., Bengio, S.: Adversarial machine learning at
  scale. CoRR  \textbf{abs/1611.01236} (2016)

\bibitem{pc_landrieu2019point}
Landrieu, L., Boussaha, M.: Point cloud oversegmentation with graph-structured
  deep metric learning pp. 7440--7449 (2019)

\bibitem{pc_landrieu2018large}
Landrieu, L., Simonovsky, M.: Large-scale point cloud semantic segmentation
  with superpoint graphs. In: Proceedings of the IEEE Conference on Computer
  Vision and Pattern Recognition (CVPR). pp. 4558--4567 (2018)

\bibitem{pc_li2018so}
Li, J., Chen, B.M., Hee~Lee, G.: So-net: Self-organizing network for point
  cloud analysis. In: Proceedings of the IEEE Conference on Computer Vision and
  Pattern Recognition (CVPR). pp. 9397--9406 (2018)

\bibitem{pc_li2018pointcnn}
Li, Y., Bu, R., Sun, M., Wu, W., Di, X., Chen, B.: Pointcnn: Convolution on
  x-transformed points. In: Advances in neural information processing systems
  (NIPS). pp. 820--830 (2018)

\bibitem{pgd-madry}
Madry, A., Makelov, A., Schmidt, L., Tsipras, D., Vladu, A.: Towards deep
  learning models resistant to adversarial attacks. In: International
  Conference on Learning Representations (ICLR) (2018)

\bibitem{universal-attack}
Moosavi-Dezfooli, S.M., Fawzi, A., Fawzi, O., Frossard, P.: Universal
  adversarial perturbations. In: The IEEE Conference on Computer Vision and
  Pattern Recognition (CVPR) (2017)

\bibitem{deepfool}
Moosavi-Dezfooli, S.M., Fawzi, A., Frossard, P.: Deepfool: A simple and
  accurate method to fool deep neural networks. In: The IEEE Conference on
  Computer Vision and Pattern Recognition (CVPR) (2016)

\bibitem{learnattack1}
Naseer, M.M., Khan, S.H., Khan, M.H., Shahbaz~Khan, F., Porikli, F.:
  Cross-domain transferability of adversarial perturbations. In: Advances in
  Neural Information Processing Systems (NeurIPS), pp. 12905--12915 (2019)

\bibitem{learnattack2}
Poursaeed, O., Katsman, I., Gao, B., Belongie, S.: Generative adversarial
  perturbations. In: Proceedings of the IEEE Conference on Computer Vision and
  Pattern Recognition (CVPR). pp. 4422--4431 (2018)

\bibitem{pointnet}
Qi, C.R., Su, H., Mo, K., Guibas, L.J.: Pointnet: Deep learning on point sets
  for 3d classification and segmentation. In: Proceedings of the IEEE
  Conference on Computer Vision and Pattern Recognition (CVPR). pp. 652--660
  (2017)

\bibitem{pointnet++}
Qi, C.R., Yi, L., Su, H., Guibas, L.J.: Pointnet++: Deep hierarchical feature
  learning on point sets in a metric space. In: Advances in neural information
  processing systems (NIPS). pp. 5099--5108 (2017)

\bibitem{first-attack}
Szegedy, C., Zaremba, W., Sutskever, I., Bruna, J., Erhan, D., Goodfellow,
  I.J., Fergus, R.: Intriguing properties of neural networks. CoRR
  \textbf{abs/1312.6199} (2013)

\bibitem{pc_tatarchenko2018tangent}
Tatarchenko, M., Park, J., Koltun, V., Zhou, Q.Y.: Tangent convolutions for
  dense prediction in 3d. In: Proceedings of the IEEE Conference on Computer
  Vision and Pattern Recognition (CVPR). pp. 3887--3896 (2018)

\bibitem{robustshapeattack}
Tsai, T., Yang, K., Ho, T.Y., Jin, Y.: Robust adversarial objects against deep
  learning models. In: AAAI Conference on Artificial Intelligence (2020)

\bibitem{autozoom}
Tu, C.C., Ting, P., Chen, P.Y., Liu, S., Zhang, H., Yi, J., Hsieh, C.J., Cheng,
  S.M.: Autozoom: Autoencoder-based zeroth order optimization method for
  attacking black-box neural networks. In: Proceedings of the AAAI Conference
  on Artificial Intelligence. vol.~33, pp. 742--749 (2019)

\bibitem{3d-obj-attk}
Tu, J., Ren, M., Manivasagam, S., Liang, M., Yang, B., Du, R., Cheng, F.,
  Urtasun, R.: Physically realizable adversarial examples for lidar object
  detection. In: Proceedings of the IEEE Conference on Computer Vision and
  Pattern Recognition (CVPR). pp. 13716--13725 (2020)

\bibitem{pc_wang2018sgpn}
Wang, W., Yu, R., Huang, Q., Neumann, U.: Sgpn: Similarity group proposal
  network for 3d point cloud instance segmentation. In: Proceedings of the IEEE
  Conference on Computer Vision and Pattern Recognition (CVPR). pp. 2569--2578
  (2018)

\bibitem{dgcn}
Wang, Y., Sun, Y., Liu, Z., Sarma, S.E., Bronstein, M.M., Solomon, J.M.:
  Dynamic graph cnn for learning on point clouds. ACM Transactions on Graphics
  (TOG)  (2019)

\bibitem{modelnet}
Wu, Z., Song, S., Khosla, A., Yu, F., Zhang, L., Tang, X., Xiao, J.: 3d
  shapenets: A deep representation for volumetric shapes. In: 2015 IEEE
  Conference on Computer Vision and Pattern Recognition (CVPR). pp. 1912--1920
  (2015)

\bibitem{pcattack}
Xiang, C., Qi, C.R., Li, B.: Generating 3d adversarial point clouds. In:
  Proceedings of the IEEE Conference on Computer Vision and Pattern Recognition
  (CVPR). pp. 9136--9144 (2019)

\bibitem{meshadv}
Xiao, C., Yang, D., Li, B., Deng, J., Liu, M.: Meshadv: Adversarial meshes for
  visual recognition. In: Proceedings of the IEEE Conference on Computer Vision
  and Pattern Recognition (CVPR). pp. 6898--6907 (2019)

\bibitem{pc_ye20183d}
Ye, X., Li, J., Huang, H., Du, L., Zhang, X.: 3d recurrent neural networks with
  context fusion for point cloud semantic segmentation. In: European Conference
  on Computer Vision (ECCV). pp. 415--430. Springer (2018)

\bibitem{punet}
Yu, L., Li, X., Fu, C.W., Cohen-Or, D., Heng, P.A.: Pu-net: Point cloud
  upsampling network. In: Proceedings of IEEE Conference on Computer Vision and
  Pattern Recognition (CVPR) (2018)

\bibitem{physicalattack}
Zeng, X., Liu, C., Wang, Y.S., Qiu, W., Xie, L., Tai, Y.W., Tang, C.K., Yuille,
  A.L.: Adversarial attacks beyond the image space. In: The IEEE Conference on
  Computer Vision and Pattern Recognition (CVPR) (2019)

\bibitem{gan-attack}
Zhao, Z., Dua, D., Singh, S.: Generating natural adversarial examples. In:
  International Conference on Learning Representations (ICLR) (2018)

\bibitem{pointdrop}
Zheng, T., Chen, C., Yuan, J., Li, B., Ren, K.: Pointcloud saliency maps. In:
  The IEEE International Conference on Computer Vision (ICCV) (2019)

\bibitem{Deflecting}
Zhou, H., Chen, K., Zhang, W., Fang, H., Zhou, W., Yu, N.: Dup-net: Denoiser
  and upsampler network for 3d adversarial point clouds defense. In: The IEEE
  International Conference on Computer Vision (ICCV) (2019)

\end{thebibliography}
\clearpage
\appendix

\section{Background on Point Cloud Distances}
We define a point cloud $\mathcal{X} \in \mathbb{R}^{N \times 3}$, as a set of $N$ 3D points, where each point  $\mathbf{x}_i \in \mathbb{R}^{3}$ is represented by its 3D coordinates $(x_i, y_i, z_i)$. In this work, we focus solely on the perturbations of the input. This means we modify each point $\mathbf{x}_i$ by a perturbation variable. Formally, we define the perturbed point set $\mathcal{X}^{\prime} = \mathcal{X} + \boldsymbol{\Delta}$, where $\boldsymbol{\Delta} \in \mathbb{R}^{N \times 3}$ is the perturbation parameter we are optimizing for. Consequently, each pair ($\mathbf{x}_i, \mathbf{x}^{\prime}_i$) are in correspondence.
\subsection{Trivial Distances ($\ell_p$)}
The most commonly used distance metric in adversarial attacks in the image domain is $\ell_p$. Unlike image domain where every pixel corresponds to the perturbed pixel in adversarial attacks, in point clouds adversarial attacks by adding, removing, or transforming the point cloud destroys the correspondence relationship to the unperturbed sample. Hence, it becomes infeasible to accurately calculate the $ \ell_p$ metric for the attack. In our paper, we focus on adversarial perturbations, which preserves the matching between the unperturbed sample and the perturbed sample. This property of preservation of matching points allows us to measure the $\ell_p$ norms of the attack exactly, which allow for standard evaluation similar to the one in the image domain.Here we assume the two point-sets are equal in size and are aligned , \ie for $\mathbf{x}_{i} \in \mathcal{X}  ~~ ,~~ \mathbf{x}_{i}^{\prime}  = \mathbf{x}_{i} + \boldsymbol{\delta}_i~ ,~ i \in 1,2,...,N  $

\begin{equation} 
\label{eq:sup-lp}
\mathcal{D}_{\ell_p}\left(\mathcal{X}, \mathcal{X}^{\prime}\right)=\left(\sum_{i}\left\|\boldsymbol{\delta}_i\right\|_{p}^{p}\right)^{\frac{1}{p}}
\end{equation}

For our attacks, we use the $\ell_2$ and $\ell_\infty$ distances, defined in (\ref{eq:sup-l2}) and (\ref{eq:sup-linfty}) respectively. The $\ell_2$ distance measures the energy of the perturbation, while $\ell_\infty$ represents the maximum allowed perturbation of each $\mathbf{\delta}_i \in \boldsymbol{\Delta}$.

\mysection{$\ell_2$ distance,}
The $\ell_2$ measures the energy of the perturbation performed on the unperturbed point cloud . Its calculation is similar to calculating the Frobenius norm of the matrix $\mathbf{X}$ that represent the point set perturbation variable $\boldsymbol{\Delta}$ such that each row of $\mathbf{X}$ is a point $ \mathbf{\delta}_i \in \boldsymbol{\Delta}$. The $\ell_2$ distance between two point sets can be measured as follows  
\begin{equation} 
\label{eq:sup-l2}
\mathcal{D}_{\ell_2}\left(\mathcal{X}, \mathcal{X}^{\prime}\right)=   
\left(\sum_{i}\left\|\boldsymbol{\delta}_i\right\|_{2}^{2}\right)^{\frac{1}{2}} = \left\| \boldsymbol{\Delta} \right\|_{\text{F}}
\end{equation}

\mysection{$\ell_\infty$ distance,}
The $\ell_\infty$ represents the max allowed perturbation at any dimension to every single point $ \mathbf{\delta}_i$ in the perturbation set  $\boldsymbol{\Delta}$ . This distance between two point sets can be measured as follows :
\begin{equation} 
\label{eq:sup-linfty}
\mathcal{D}_{\ell_\infty}\left(\mathcal{X}, \mathcal{X}^{\prime}\right)=  \max _{i} 
\left\|\boldsymbol{\delta}_i\right\|_{\infty}
\end{equation}

\subsection{Non-trivial Distances}
Other point cloud distances that are commonly used in the literature do not require the two sets to be in a known correspondence (like the strict $\ell_p$). These distance metrics include the following: Chamfer Distances, Hausdorff Distance, and Earth Mover Distance. In what follows, we formally present each of these metrics.

\mysection{Chamfer Distance (CD)}
This is a common distance to compare 2 point sets. CD measures the average distance between closest point pairs of 2 different point clouds. We define CD in \eqLabel{\ref{eq:sup-chamfer}}.
\begin{equation} 
\label{eq:sup-chamfer}
\mathcal{D}_{CD}\left(\mathcal{X}, \mathcal{X}^{\prime}\right)=\frac{1}{\left\|\mathcal{X}^{\prime}\right\|_{0}} \sum_{\mathbf{x}_{i}^{\prime} \in \mathcal{X}^{\prime}} \min _{\mathbf{x}_{i} \in \mathcal{X}}\left\|\mathbf{x}_{i} -\mathbf{x}_{i}^{\prime}\right\|_{2}^{2}
\end{equation}

\mysection{Hausdorff distance (HD)}
With HD, we compute the largest distance in the set of containing $\mathbf{x} \in \mathcal{X}$ and its closest point $\mathbf{x}^{\prime} \in \mathcal{X}^{\prime}$. We define HD as follows:
\begin{equation} \label{eq:sup-hausdorff}
\mathcal{D}_{H}\left(\mathcal{X}, \mathcal{X}^{\prime}\right)=\max _{\mathbf{x}_{i}^{\prime} \in \mathcal{X}^{\prime}} \min _{\mathbf{x}_{i} \in \mathcal{X}}\left\|\mathbf{x}_{i} -\mathbf{x}_{i}^{\prime}\right\|_{2}^{2}
\end{equation}

\mysection{Earth Mover Distance (EMD)}
The EMD measures
the total \emph{effort} performed in the optimal transport scheme that transforms the first point set to the other. It is defined as follows: 
\begin{equation} \label{eq:sup-emd}
\mathcal{D}_{E M D}\left(\mathcal{X}, \mathcal{X}^{\prime}\right)=\min _{\phi: \mathcal{X} \rightarrow \mathcal{X}^{\prime}} \sum_{i}\|\mathbf{x}_{i}^{\prime} -\phi(\mathbf{x}_{i})\|_{2},
\end{equation}
where $\phi: \mathcal{X} \rightarrow \mathcal{X}^{\prime} $ is a bijection transform.

\clearpage

\section{Our Full Formulation} \label{sec:sup-full-formulation}
\begin{figure*}[t]
\begin{center}
   \includegraphics[width=1\linewidth]{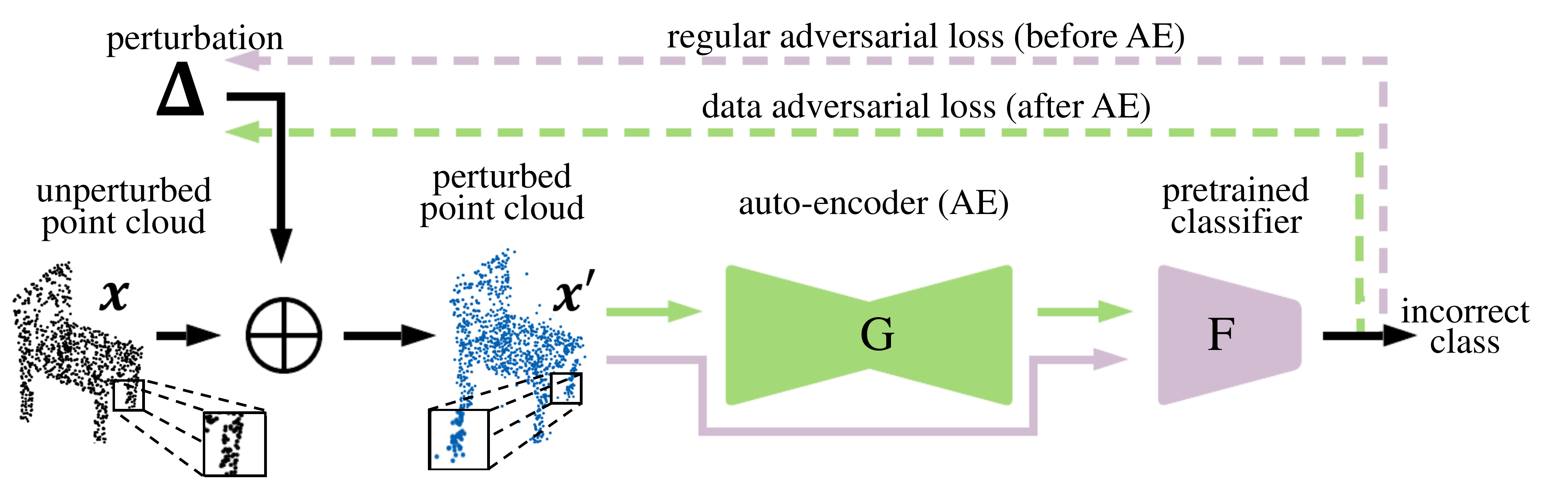} %
     \end{center}
   \caption{\small \textbf{AdvPC Attack Pipeline:} We optimize for the constrained perturbation variable $\boldsymbol{\Delta}$ to generate the perturbed sample $\mathcal{X}^{\prime} = \mathcal{X} + \boldsymbol{\Delta}$. The perturbed sample fools a trained classifier $\mathbf{F}$ (\ie $\mathbf{F}(\mathcal{X}^{\prime})$ is incorrect), and at the same time, if the perturbed sample is reconstructed by an Auto-Encoder (AE) $\mathbf{G}$, it too fools the classifier (\ie $\mathbf{F}(\mathbf{G}(\mathcal{X}^{\prime}))$ is incorrect). The AdvPC loss for network $\mathbf{F}$ is defined in \eqLabel{\ref{eq:sup-final-objective}} and has two parts: network adversarial loss (\textit{\textit{purple}}) and data adversarial loss (\textit{green}). Dotted lines are gradients flowing to the perturbation variable $\boldsymbol{\Delta}$.}
     \vspace{-12pt}

\label{fig:sup-:pipeline}
\end{figure*}

The pipeline of AdvPC is illustrated in \figLabel{\ref{fig:sup-:pipeline}}. It consists of an Auto-Encoder (AE) $\mathbf{G}$, which is trained to reconstruct 3D point clouds, and a point cloud classifier $\mathbf{F}$.  
We seek to find a perturbation variable $\boldsymbol{\Delta}$ added to the input $\mathcal{X}$ to fool $\mathbf{F}$  before \emph{and} after it passes through the AE for reconstruction. The setup makes the attack less dependent on the victim network and more dependent on the data (leveraged by the AE). As such, we expect this strategy to generalize to different networks.
Next, we describe the main components of our pipeline: 3D point cloud input, AE, and point cloud classifier, and then we present our attack setup and loss. %
\subsection{AdvPC Attack Pipeline}
\mysection{3D Point Clouds ($\mathcal{X}$)}
We define a point cloud $\mathcal{X} \in \mathbb{R}^{N \times 3}$, as a set of $N$ 3D points, where each point  $\mathbf{x}_i \in \mathbb{R}^{3}$ is represented by its 3D coordinates $(x_i, y_i, z_i)$. 

\mysection{Point Cloud Networks ($\mathbf{F}$)}
We focus on 3D point cloud classifiers with a feature max pooling layer as detailed in \eqLabel{\ref{eq:sup-pcnet}}, where $h_{\text{mlp}}$ and $h_{\text{conv}}$ are MLP and Convolutional ($1\times1$ or edge) layers respectively. This produces a K-class classifier $\mathbf{F}$.
\begin{equation}
\label{eq:sup-pcnet}
\mathbf{F}(\mathcal{X})=h_{\text{mlp}} ( \max _{\mathbf{x}_{i} \in \mathcal{X}}\left\{h_{\text{conv}}\left(\mathbf{x}_{i}\right)\right\})
\end{equation}
Here, $\mathbf{F} : ~ \mathbb{R}^{N \times 3} \rightarrow \mathbb{R}^K $ produces the logits layer of the classifier with size $K$. 
For our attacks, we take $\mathbf{F}$ to be one of the following widely used networks in the literature: PointNet \cite{pointnet}, PointNet++ \cite{pointnet++} in single-scale form (SSG)  and multi-scale form (MSG), and DGCNN \cite{dgcn}. %
\secLabel{\ref{sec:sup-sens}} delves deep into the differences between them in terms of their sensitivities to adversarial perturbations.

\mysection{Point Cloud Auto-Encoder ($\mathbf{G}$)}
An AE learns a representation of the data and acts as an effective defense against adversarial attacks. It ideally projects a perturbed point cloud onto the natural manifold of inputs. Any AE architecture in point clouds can be used in our pipeline, but we select the one in \cite{pc-ae} because of its simple structure and effectiveness in recovering from adversarial perturbation. 
The AE $\mathbf{G}$ consists of an encoding part, $\mathbf{g}_{\text{encode}}:\mathbb{R}^{N \times 3} \xrightarrow{}\mathbb{R}^{q} $ (similar to \eqLabel{\ref{eq:sup-pcnet}}),
and an MLP decoder, $\mathbf{g}_{\text{mlp}}:\mathbb{R}^{q} \xrightarrow{}\mathbb{R}^{N \times 3} $, to produce a point cloud. It can be described formally as: $\mathbf{G}(.) = \mathbf{g}_{\text{mlp}}\big(\mathbf{\mathbf{g}_{\text{encode}}(\mathcal{.})} \big)$
We train the AE with the Chamfer loss as in \cite{pc-ae} on the same data used to train $\mathbf{F}$, such that it can reliably encode and decode 3D point clouds. We freeze the AE weights during the optimization of the adversarial perturbation on the input.  
We show in \secLabel{\ref{sec:sup-def-unt}} how the AE acts as an effective defense against previous point cloud adversarial perturbations. Since the AE learns how naturally occurring point clouds look like, the gradients updating the attack, which is also tasked to fool the reconstructed sample after the AE, actually become more dependent on the data and less on the victim network. The enhanced data dependency of our attack results in the success of our attacks on unseen transfer networks besides the success on the victim network.
As such, the proposed composition allows the crafted attack to successfully attack the victim classifier, as well as, fool transfer classifiers that operate on a similar input data manifold. 
Furthermore, since many of the available defenses rely on natural statistics of 3D point clouds \cite{Deflecting}, we show that attacking the classifier after AE reconstruction can also lead to perturbations resilient to these defenses. 

\subsection{AdvPC Attack Loss} 
\begin{figure}[]
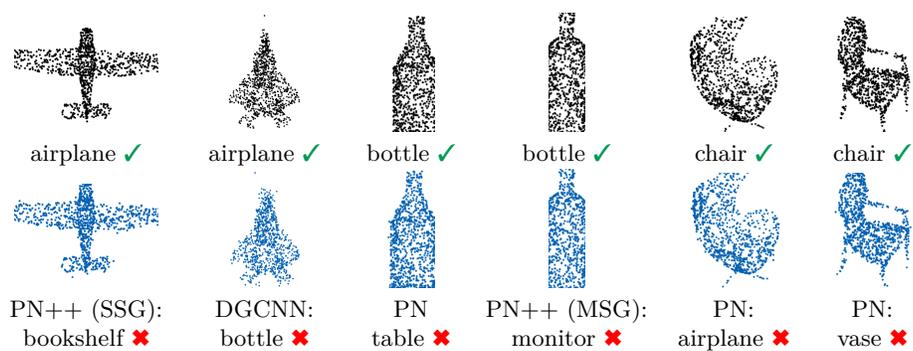

\tabcolsep=0.2cm
\begin{tabular}{cccccc}
\includegraphics[trim={1.5cm 2.5cm 1.5cm 3.4cm},clip, width = 0.75in]{images/met/E3/FQTY_v_0_t_4_b_0_n_0_orig.png} &
 \includegraphics[trim={0cm 1cm 0cm 1.5cm},clip, width = 0.75in]{images/met/E3/0I63_v_0_t_5_b_1_n_4_orig.png} & 
\includegraphics[trim={5.5cm 7cm 6cm 1.5cm},clip, width = 0.25in]{images/met/E3/8RS8_v_5_t_0_b_1_n_1_orig.png} &
\includegraphics[trim={6cm 7cm 6cm 1.5cm},clip, width = 0.19in]{images/met/E3/NOLG_v_5_t_37_b_0_n_2_orig.png} & 
\includegraphics[trim={4cm 6cm 4cm 1.5cm},clip, width = 0.55in]{images/met/E3/8RS8_v_8_t_0_b_4_n_3_orig.png} &
\includegraphics[trim={4cm 6cm 4cm 1.5cm},clip, width = 0.55in]{images/met/E3/8RS8_v_8_t_5_b_4_n_0_orig.png} \\
airplane \color{ForestGreen}\cmark &
airplane \color{ForestGreen}\cmark &
bottle \color{ForestGreen}\cmark &
bottle \color{ForestGreen}\cmark &
chair \color{ForestGreen}\cmark &
chair \color{ForestGreen}\cmark \\
\includegraphics[trim={1.5cm 2.5cm 1.5cm 3.4cm},clip, width = 0.75in]{images/met/E3/FQTY_v_0_t_2_b_0_n_0_adv.png} &
 \includegraphics[trim={0cm 1cm 0cm 1.5cm},clip, width = 0.75in]{images/met/E3/0I63_v_0_t_5_b_1_n_4_adv.png} &
\includegraphics[trim={5.5cm 7cm 6cm 1.5cm},clip, width = 0.25in]{images/met/E3/8RS8_v_5_t_0_b_1_n_1_adv.png} &
\includegraphics[trim={6cm 7cm 6cm 1.5cm},clip, width = 0.19in]{images/met/E3/NOLG_v_5_t_37_b_0_n_2_adv.png} &
\includegraphics[trim={4cm 6cm 4cm 1.5cm},clip, width = 0.55in]{images/met/E3/8RS8_v_8_t_0_b_4_n_3_adv.png} &
\includegraphics[trim={4cm 6cm 4cm 1.5cm},clip, width = 0.55in]{images/met/E3/8RS8_v_8_t_5_b_4_n_0_adv.png} \\
PN++ (SSG):&
DGCNN: &
PN &
PN++ (MSG):&
PN:&
PN:\\
bookshelf  \color{red}\xmark &
bottle \color{red}\xmark &
table \color{red}\xmark &
monitor  \color{red}\xmark &
airplane \color{red}\xmark &
vase \color{red}\xmark \\

\end{tabular}
\caption{\small \textbf{Examples of AdvPC Targeted Attacks:} Adversarial attacks are generated for victim networks PointNet, PointNet ++ (MSG/SSG) and DGCNN using AdvPC. The unperturbed point clouds are in black (\textit{top}) while the perturbed examples are in blue (\textit{bottom}). The network predictions are shown under each point cloud. The wrong prediction of each perturbed point cloud matches the target of the AdvPC attack. %
}
\label{fig:sup-qualitative}
\vspace{-8pt}
\end{figure}

\mysection{Soft Constraint Loss}
In AdvPC attacks, like the ones in \figLabel{\ref{fig:sup-qualitative}}, we focus solely on perturbations of the input. We modify each point $\mathbf{x}_i$ by a an addictive perturbation variable $\delta_i$. Formally, we define the perturbed point set $\mathcal{X}^{\prime} = \mathcal{X} + \boldsymbol{\Delta}$, where $\boldsymbol{\Delta} \in \mathbb{R}^{N \times 3}$ is the perturbation parameter we are optimizing for. Consequently, each pair ($\mathbf{x}_i, \mathbf{x}^{\prime}_i$) are in correspondence. Adversarial attacks are commonly formulated as in \eqLabel{\ref{eq:sup-adv-attack}}, where the goal is to find an input perturbation $\boldsymbol{\Delta}$ that successfully fools $\mathbf{F}$ into predicting an incorrect label $t^{\prime}$, while 
keeping $\mathcal{X^{\prime}}$ and $\mathcal{X}$ close under distance metric  $\mathcal{D}\colon \mathbb{R}^{N\times3} \times \mathbb{R}^{N\times3} \rightarrow \mathbb{R}$. 
\begin{equation} 
\label{eq:sup-adv-attack}
\min_{\boldsymbol{\Delta}} ~~\mathcal{D}\left(\mathcal{X}, \mathcal{X}^{\prime}\right) \quad \text { s.t. } \left[\argmax_{i}~\mathbf{F}\left(\mathcal{X}^{\prime}\right)_{i}\right]=t^{\prime}
\end{equation}
The formulation in \eqLabel{\ref{eq:sup-adv-attack}} can describe targeted attacks (if $t^\prime$ is specified before the attack) or untargeted attacks (if $t^\prime$ is any label other than the true label of $\mathcal{X}$). We adopt the following choice of $t^\prime$ for untargeted attacks: $t^\prime = \left[\argmax_{i\neq \text{true}}~\mathbf{F}\left(\mathcal{X}^{\prime}\right)_{i}\right]$.We present the results of both targeted and untargeted attacks in this supplementary.
As pointed out in \cite{carlini}, it is difficult to directly solve \eqLabel{\ref{eq:sup-adv-attack}}. Instead, previous works like \cite{pcattack,robustshapeattack} have used the well-known C\&W formulation, giving rise to the commonly known soft constraint attack:
\begin{equation} 
\label{eq:sup-attack-soft}
\min_{\boldsymbol{\Delta}} ~~f_{t^{\prime}}\left(\mathbf{F}(\mathcal{X}^{\prime})\right)+ ~\lambda~ \mathcal{D}\left(\mathcal{X}, \mathcal{X}^{\prime}\right)
\end{equation}
where $\mathcal{D}\left(\mathcal{X}, \mathcal{X}^{\prime}\right)$ can be any of the distances proposed in \eqLabel{\ref{eq:sup-l2},\ref{eq:sup-chamfer},\ref{eq:sup-emd}}, while  $f_{t^{\prime}}\left(\mathbf{F}(\mathcal{X}^{\prime})\right)$ is the targeted adversarial loss function defined on the network $\mathbf{F}$ to move it to target $t^{\prime}$ as in \eqLabel{\ref{eq:sup-adv-loss}}.  %
\begin{equation} 
\label{eq:sup-adv-loss}
f_{t^{\prime}}\left(\mathbf{F}(\mathcal{X}^{\prime})\right)=\max\left(\max _{i \neq t^{\prime}}\left(\mathbf{F}\left(\mathcal{X}^{\prime}\right)_{i}\right)-\mathbf{F}\left(\mathcal{X}^{\prime}\right)_{t^{\prime}} + \kappa,0\right),
\end{equation}
where $\kappa$ is a loss margin. The 3D-Adv attack \cite{pcattack} uses $\ell_2$ for $\mathcal{D}\left(\mathcal{X}, \mathcal{X}^{\prime}\right)$ while KNN attack \cite{robustshapeattack} uses Chamfer Distance.

\mysection{Hard Constraint Loss}
An alternative to \eqLabel{\ref{eq:sup-adv-attack}} is to put $\mathcal{D}\left(\mathcal{X}, \mathcal{X}^{\prime}\right)$ as a hard constraint, where the objective can be minimized using Projected Gradient Descent (PGD) \cite{projected-gradient,pgd-madry} as follows. 

\begin{equation} 
\label{eq:sup-attack-hard}
\min_{\boldsymbol{\Delta}} ~~ f_{t^{\prime}}\left(\mathbf{F}(\mathcal{X}^{\prime})\right) ~~~~ s.t.~~ \mathcal{D}\left(\mathcal{X}, \mathcal{X}^{\prime}\right) \leq \epsilon 
\end{equation}

Using a hard constraint sets a limit to the amount of added perturbation in the attack. This limit is defined by $\epsilon$ in \eqLabel{\ref{eq:sup-attack-hard}}. Having this bound ensures fair comparisons between different attacks schemes. We can do these comparisons by measuring the effectiveness of these attacks at different levels of $\epsilon$.
Using PGD, the above optimization in \eqLabel{\ref{eq:sup-attack-hard}} with $\ell_p$ distance $\mathcal{D}_{\ell_p}\left(\mathcal{X}, \mathcal{X}^{\prime}\right)$ can be solved by iteratively projecting the perturbed sample onto the $\ell_p$ sphere of size $\epsilon_p$ such that:  
\begin{equation} \label{eq:sup-pert-hard}
\boldsymbol{\Delta}_{t+1} =  \Pi_{p}\left( \boldsymbol{\Delta}_{t} - \eta \nabla _{\boldsymbol{\Delta}_{t}}f_{t^{\prime}}\left(\mathbf{F}(\mathcal{X}^{\prime})\right),\epsilon_p \right)
\end{equation}
Here, $\Pi_{p}\left(\boldsymbol{\Delta},\epsilon_p\right)$  projects the perturbation $\boldsymbol{\Delta}$ onto the $\ell_p$ sphere of size $\epsilon_p$, and $\eta$ is a step size. The two most commonly used $\ell_p$ distance metrics in the literature are $\ell_2$, which measures the energy of the perturbation, and $\ell_\infty$, which measures the maximum point perturbation of each $\boldsymbol{\delta}_i \in \boldsymbol{\Delta}$. Our experiments use the $\ell_2$ distance defined as in \eqLabel{\ref{eq:sup-l2}}.
while the projection of $\boldsymbol{\Delta}$ onto the $\ell_2$ sphere of size $\epsilon_2$ is:
\begin{equation} \label{eq:sup-project-two}
\Pi_{2}\left(\boldsymbol{\Delta},\epsilon_2\right) = \frac{\epsilon_2}{\max\left( \left\| \boldsymbol{\Delta} \right\|_{\text{F}},\epsilon_2\right)} \boldsymbol{\Delta}
\end{equation}
On the other hand, the $\ell_\infty$ projection formulation is as follows:
\begin{equation} \label{eq:sup-project-infty}
\Pi_{\infty}\left(\boldsymbol{\Delta},\epsilon_{\infty}\right) = \text{SAT}_{\epsilon_{\infty}}(\boldsymbol{\delta}_{i}), \quad \forall \boldsymbol{\delta}_{i} \in \boldsymbol{\Delta},
\end{equation}
here $\text{SAT}_{    \zeta}\left(\boldsymbol{\delta}_{i}\right)$ is the element-wise saturation function that takes every element of vector $\boldsymbol{\delta}_{i}$ and limit its range in $[-\zeta,\zeta]$.

\mysection{Data Adversarial Loss}
The objectives in \eqLabel{\ref{eq:sup-adv-attack}, \ref{eq:sup-attack-hard}} focus solely on the network $\mathbf{F}$. We also want to add more focus on the data in crafting our attacks. We do so by fooling $\mathbf{F}$ using both the perturbed input $\mathcal{X}^{\prime}$ and the AE reconstruction $\mathbf{G}(\mathcal{X}^{\prime})$. Our new objective becomes:
\begin{align} 
 \label{eq:sup-pre-final-objective}
 \min_{\boldsymbol{\Delta}}  ~~\mathcal{D}\left(\mathcal{X}, \mathcal{X}^{\prime}\right) \quad
    \text { s.t. } [\argmax _{i}~\mathbf{F}\left(\mathcal{X}^{\prime}\right)_{i}]= t^{\prime};~~ [\argmax _{i}~\mathbf{F}\left(\mathbf{G}(\mathcal{X}^{\prime})\right)_{i}] = t^{\prime\prime}
\end{align}
Here, $t^{\prime\prime}$ is any incorrect label $ t^{\prime\prime} \neq \argmax _{i}\mathbf{F}\left(\mathcal{X}\right)_{i}$ and $t^{\prime}$ is just like \eqLabel{\ref{eq:sup-adv-attack}}. The second constraint ensures that the prediction of the perturbed sample after the AE %
differs from the true label of the unperturbed sample. %
Similar to \eqLabel{\ref{eq:sup-adv-attack}}, this objective is hard to optimize, so we follow similar steps as in \eqLabel{\ref{eq:sup-attack-hard}} and optimize the following objective for AdvPC using PGD (using $\ell_p$ as the distance metric): 

\begin{equation} 
\label{eq:sup-final-objective}
\begin{aligned}
\min_{\boldsymbol{\Delta}} ~~ (1 - \gamma) ~f_{t^{\prime}}\left(\mathbf{F}(\mathcal{X}^{\prime})\right) + \gamma ~ f_{t^{\prime\prime}}\left(\mathbf{F}\left(\mathbf{G}(\mathcal{X}^{\prime})\right)\right)   \quad
& s.t. ~~ \mathcal{D}_{\ell_p}\left(\mathcal{X}, \mathcal{X}^{\prime}\right) \leq \epsilon_p 
\end{aligned}
\end{equation}
Here,$f$ is as in \eqLabel{\ref{eq:sup-adv-loss}}, while $\gamma$ is a hyper-parameter that trades off the attack's success before and after the AE . %
When $\gamma = 0$, the formulation in \eqLabel{\ref{eq:sup-final-objective}} becomes \eqLabel{\ref{eq:sup-attack-hard}}.. We use PGD to solve \eqLabel{\ref{eq:sup-final-objective}} as follows.
\begin{equation}   \label{eq:sup-final-pert-hard}
\begin{aligned}
\boldsymbol{\Delta}_{t+1} =  \Pi_{p}\Big( \boldsymbol{\Delta}_{t} &- \eta (1 - \gamma) \nabla _{\boldsymbol{\Delta}_{t}} ~f_{t^{\prime}}\big(\mathbf{F}(\mathcal{X}^{\prime})\big) \\ &-  \eta ~ \gamma ~ \nabla _{\boldsymbol{\Delta}_{t}} ~ f_{t^{\prime \prime }}\big(\mathbf{F}\left(\mathbf{G}(\mathcal{X}^{\prime})\right)\big),\epsilon_p \Big)
\end{aligned}
\end{equation}
Where $\Pi_{p}$ is the projection to $\ell_p$ as in \eqLabel{\ref{eq:sup-project-two},\ref{eq:sup-project-infty}}

We follow the same procedure as in \cite{pcattack} when solving the optimization in \eqLabel{\ref{eq:sup-final-objective}} by keeping a record of any $\boldsymbol{\Delta}$ that satisfies the constraints in \eqLabel{\ref{eq:sup-pre-final-objective}} and by trying different initializations for $\boldsymbol{\Delta}$. If we achieve the constraints in \eqLabel{\ref{eq:sup-pre-final-objective}} in one of the optimizations' initializations, we try smaller hard norms in the following initialization in order to find a better solution ( smaller norm). The is the exactly the Binary Search followed by \cite{pcattack} to find the best hyperparameter $\lambda$ in \eqLabel{\ref{eq:sup-attack-soft}} that will result in the smallest norm perturbation that succeeds in the attack on that specific sample.

\clearpage
\section{Qualitative Results }
\vspace{-10pt}
\vspace{-8pt}
\begin{figure}[b]
\tabcolsep=0cm

\begin{tabular}{cccccc}

\includegraphics[trim={1cm 1cm 1cm 1.5cm},clip,width = 0.75in]{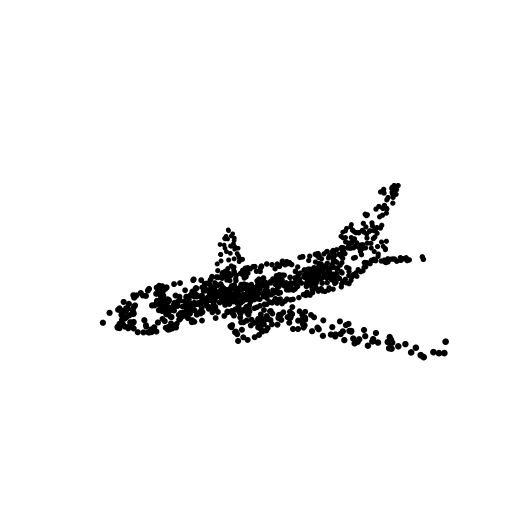} &
\includegraphics[trim={1cm 1cm 1cm 1.5cm},clip,width = 0.75in]{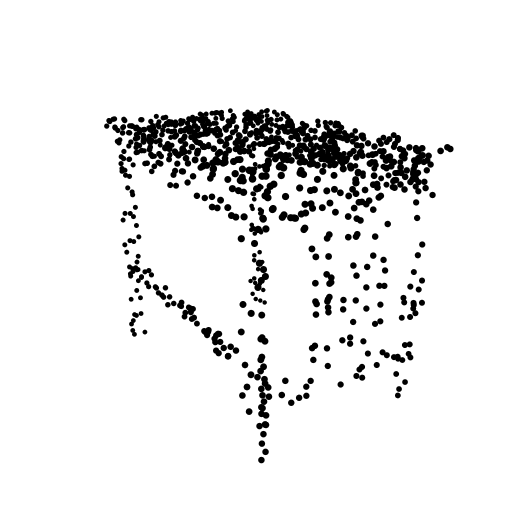} &
\includegraphics[trim={1cm 1cm 1cm 1.5cm},clip,width = 0.75in]{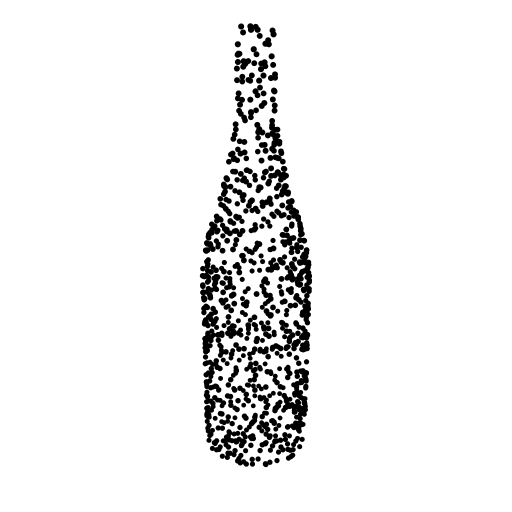} &
\includegraphics[trim={1cm 1cm 1cm 1.5cm},clip,width = 0.75in]{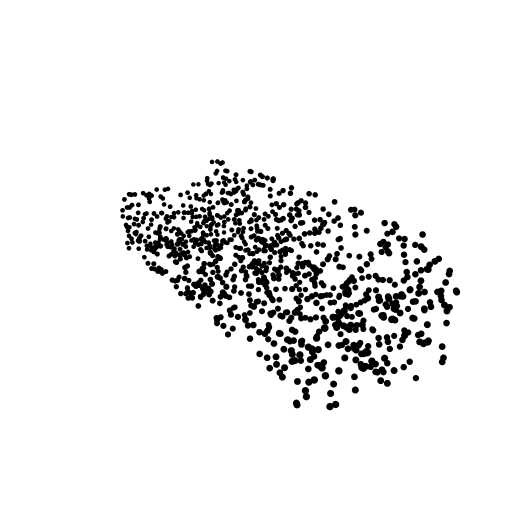} &
\includegraphics[trim={1cm 1cm 1cm 1.5cm},clip,width = 0.75in]{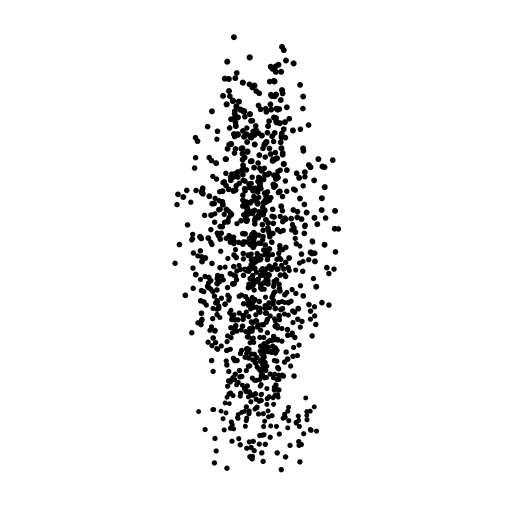} &
\includegraphics[trim={1cm 1cm 1cm 1.5cm},clip,width = 0.75in]{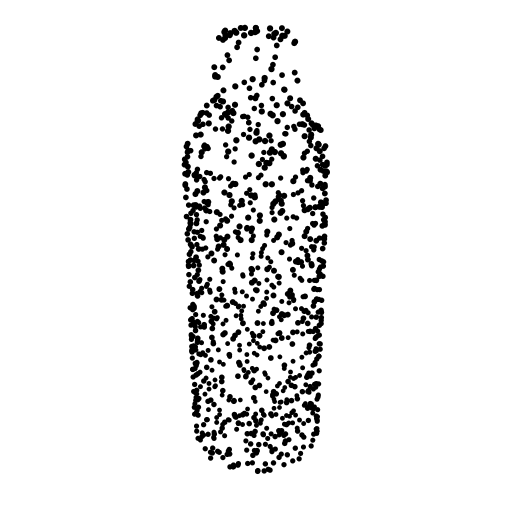}  \\

airplane \color{ForestGreen}\cmark &
table \color{ForestGreen}\cmark &
bottle \color{ForestGreen}\cmark &
sofa \color{ForestGreen}\cmark &
vase \color{ForestGreen}\cmark &
bottle \color{ForestGreen}\cmark \\

\includegraphics[trim={0cm 1cm 1cm 1.5cm},clip,width = 0.75in]{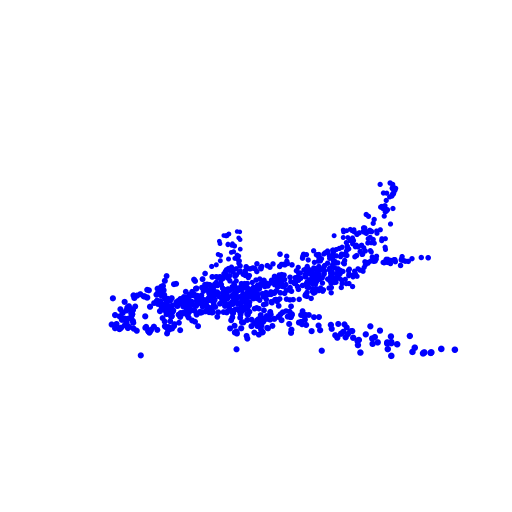} &
\includegraphics[trim={0cm 1cm 1cm 1.5cm},clip,width = 0.75in]{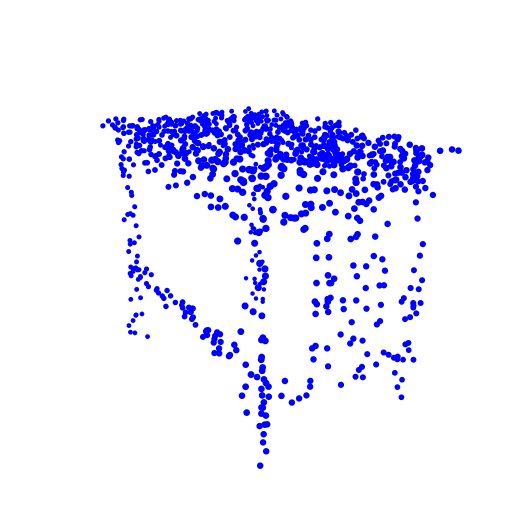} &
\includegraphics[trim={1cm 1cm 1cm 1.5cm},clip,width = 0.75in]{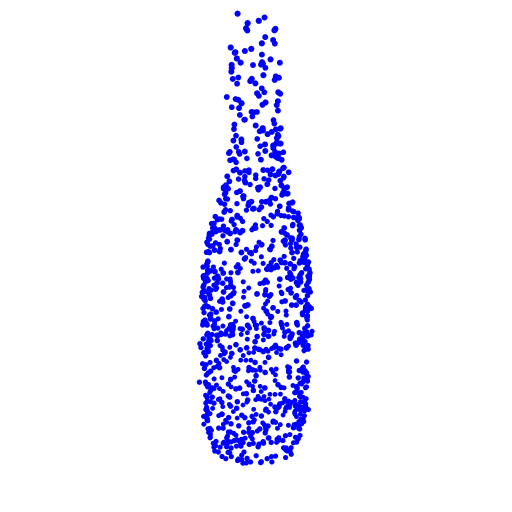} &
\includegraphics[trim={1cm 1cm 1cm 1.5cm},clip,width = 0.75in]{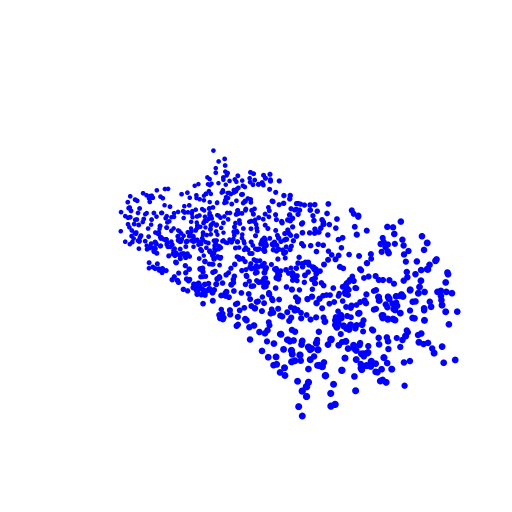} &
\includegraphics[trim={1cm 1cm 1cm 1.5cm},clip,width = 0.75in]{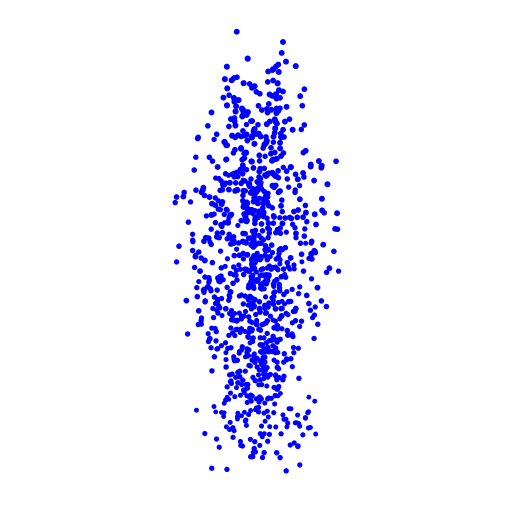} &
\includegraphics[trim={1cm 1cm 1cm 1.5cm},clip,width = 0.75in]{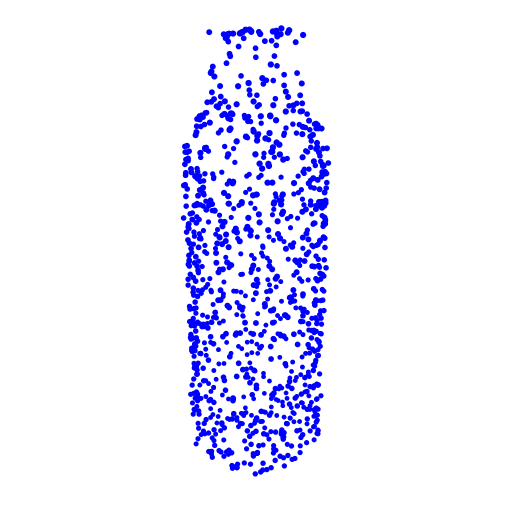}  \\

PN++ (SSG):&
PN: &
PN++ (MSG) &
PN: &
PN++ (SSG):&
DGCNN:\\

sofa  \color{red}\xmark &
stand \color{red}\xmark &
sofa \color{red}\xmark &
bench  \color{red}\xmark &
sofa \color{red}\xmark &
sofa \color{red}\xmark \\

\end{tabular}
\caption{\small \textbf{Examples of AdvPC Untargeted Attacks:} Adversarial attacks are generated for victim networks PointNet, PointNet ++ (MSG/SSG) and DGCNN using AdvPC. The unperturbed point clouds are in black (\textit{top}) while the perturbed examples are in blue (\textit{bottom}). The network predictions are shown under each point cloud.%
}
\label{fig:sup-qualitative_unt}
\vspace{-8pt}
\end{figure}

\clearpage

\section{Experiments Setup} \label{sec:sup-setup}
\subsection{Dataset and Networks}
We use ModelNet40 \cite{modelnet} to train the classifier network ($\mathbf{F}$) and the AE network ($\mathbf{G}$), as well as test our attacks. ModelNet40 contains 12,311 CAD models from 40 different classes. CAD models are divided into 9,843 for training and 2,468 for testing. Similar to previous work \cite{Deflecting,pcattack,pointdrop}, we sample 1,024 points from each object. We train the $\mathbf{F}$ victim networks: PointNet\cite{pointnet}, PointNet++ in both Single-Scale (SSG) and Multi-scale  (MSG) \cite{pointnet++} settings, and DGCNN \cite{dgcn}. %
For a fair comparison, we adopt the subset of ModelNet40 detailed in \cite{pcattack} to perform and evaluate our attacks against their work (we call this the attack set). In the attack set, 250 examples are chosen from 10 ModelNet40 classes. In untargeted attacks we only perform the attack once per test sample and report the average results. However, in targeted attacks (like the ones in \secLabel{\ref{sec:sup-def-tar}}) we evaluate the attacks on all the possible targets for each sample and report the average results as followed by c\cite{pcattack}.  

\subsection{Adversarial Attack Methods}
We compare AdvPC against the state-of-the-art baselines 3D-Adv \cite{pcattack} and KNN-Attack \cite{robustshapeattack}. For all attacks, We use Adam optimizer \cite{adam} with learning rate $\eta = 0.01$, and perform 2 different initializations for the optimization of $\boldsymbol{\Delta}$ (as followed by \cite{pcattack}). The number of iterations for the attack optimization for all the networks is 200. We set the loss margin $\kappa=30$ in \eqLabel{\ref{eq:sup-adv-loss}} for both 3D-Adv \cite{pcattack} and AdvPC and $\kappa=15$ for KNN-Attack \cite{robustshapeattack} (as they suggest in their paper). For other hyperparameters of \cite{pcattack,robustshapeattack}, we follow what they report in their works. We pick $\gamma =0.25$ in \eqLabel{\ref{eq:sup-final-objective}} for AdvPC because it strikes a balance between the success of the attack and its transferability (refer to \secLabel{\ref{sec:sup-ablation}} for details). In all of the attacks, we follow the same procedure as \cite{pcattack}, where the best attack that satisfies the objective during the optimization is reported. In this supplementary, we perform both the $\ell_\infty$ and $\ell_2$ modes of attacks for the baselines and for AdvPC. For fair comparisons between the attack methods on the same norm-budgets, we add the following to all the attacks. For $\ell_\infty$ attacks, we add the hard projection $\Pi_{\infty}\left(\boldsymbol{\Delta},\epsilon_{\infty}\right)$ (from \eqLabel{\ref{eq:sup-project-infty}}, while for $\ell_2$ attack, we add the hard projection $\Pi_{2}\left(\boldsymbol{\Delta},\epsilon_{2}\right)$ (from \eqLabel{\ref{eq:sup-project-two}}. This insures that all the attacks have the same norm-budgets $\epsilon_{\infty}$ or $\epsilon_{2}$ ( depending on the attack mode).

\subsection{Transferability}
For the constrained $\ell_\infty$ metric, we measure their success rate at different norm-budgets $\epsilon_\infty$ taken to be in the range $[0,0.75]$, whereas norm-budgets $\epsilon_2$ is taken in the range $[0,7]$ . These ranges are chosen because they enables the attacks to reach 100\% success on the victim network, as well as offer an opportunity for transferability to other networks. We compare AdvPC against the state-of-the-art baselines \cite{pcattack,robustshapeattack} under these norm-budgets (\eg see \figLabel{\ref{fig:sup-transferbility-I},\ref{fig:sup-transferbility-2}}). The exact norms used for $\epsilon_\infty$ and $\epsilon_2$ are $\{0.01, 0.04, 0.05, 0.1, 0.18, 0.28, 0.35, 0.45, 0.6, 0.75\}$ and $\{0.1, 0.22, 0.48, 0.72, 1.0, 1.5, 1.8,  2.8, 4.0, 7.0\}$ respectively. At exactly $\epsilon_\infty=\epsilon_2 = 0$, we get the classification accuracies on unperturbed samples for networks PN, PN++(MSG), PN++(SSG) and DGCNN to be .92.8\%, 91.5\%, 91.5\%, and 93.7\% respectively. To measure the success of the attack, we measure the percentage of samples out of all attacked samples that the victim network misclassified. We also measure transferability from each victim network to the transfer networks. For each pair of networks, we optimize the attack on one network (victim) and measure the success rate of this optimized attack when applied as input to the other network (transfer). We report these success rates for all network pairs. No defenses are used in the transferability experiment. All the attacks performed in this section are untargeted attacks (following the convention for transferability experiments \cite{pcattack}). The results are reported in Sections \ref{sec:sup-transfer-spec},\ref{sec:sup-transfer-all}, and \ref{sec:sup-transfer-mat}

\subsection{Attacking the Defenses}
We also analyze the success of our attacks against point cloud defenses. We compare AdvPC attacks and the baselines \cite{pcattack,robustshapeattack} against several defenses used in the point cloud literature: SOR, SRS, DUP-Net \cite{Deflecting}, and Adversarial Training \cite{pcattack}. We also add a newly trained AE (different from the one used in the AdvPC attack) to this list of defenses. %
For SRS, we use a drop rate of 10\%, while in SOR, we use the same parameters proposed in \cite{Deflecting}. We train DUP-Net on ModelNet40 with an up-sampling rate of 2. For Adversarial Training, all four networks are trained using a mix of the training data of ModelNet40 and adversarial attacks generated by \cite{pcattack}. We always report the success rate as 1-accuracy of the victim networks on the perturbed data for that specific norm-budget. The results for untargeted attacks ($\ell_\infty$ and $\ell_2$) are reported in \secLabel{\ref{sec:sup-def-unt}}, while for targeted attacks, the defense results ($\ell_\infty$ and $\ell_2$) are reported in \secLabel{\ref{sec:sup-def-tar}}.

\clearpage
\section{Full Transferability Results } 

\subsection{Transferability on Specific Norms} \label{sec:sup-transfer-spec}
\begin{table}[]
\footnotesize
\setlength{\tabcolsep}{0.5pt} %
\renewcommand{\arraystretch}{1.1} %
\centering
\vspace{-2pt}
\resizebox{\textwidth}{!}{%
\begin{tabular}{cc|cccc|cccc} 
\toprule
- & - &\multicolumn{4}{c|}{$\epsilon_\infty = 0.18$} &\multicolumn{4}{c}{$\epsilon_\infty = 0.45$}\\
\specialcell{\textbf{Victim}\\\textbf{Network}} & \textbf{Attack}  &\textbf{PN} & \textbf{\specialcell{ PN++ \\(MSG)}} & \textbf{\specialcell{ PN\scriptsize++\\ (SSG)}} & \textbf{DGCNN}  &\textbf{PN} & \textbf{\specialcell{ PN++ \\(MSG)}} & \textbf{\specialcell{ PN++\\ (SSG)}} & \textbf{DGCNN}   \\
\midrule
 \multirow{3}*{\textbf{PN}} & 3D-Adv \cite{pcattack}  &  100 &  8.4 & 10.4 &  6.8 & 100 &  8.8 &  9.6 &  8.0 \\
 & KNN \cite{robustshapeattack} & 100 &  9.6 & 10.8 &  6.0 & 100 &  9.6 &  8.4 &  6.4 \\
 & AdvPC (Ours)  &  98.8 & \textbf{20.4} & \textbf{27.6} & \textbf{22.4} &  98.8 & \textbf{18.0} & \textbf{26.8} & \textbf{20.4} \\ \hline
 \multirow{3}*{\textbf{\specialcell{ PN++ \\(MSG)}}} & 3D-Adv \cite{pcattack} & 6.8 & 100 & 28.4 & 11.2 &  7.2 & 100 & 29.2 & 11.2 \\ & KNN \cite{robustshapeattack} & 6.4 & 100 & 22.0 &  8.8 &  6.4 & 100 & 23.2 &  7.6 \\ & AdvPC (Ours)  & \textbf{13.2} &  97.2 & \textbf{54.8} & \textbf{39.6} & \textbf{18.4} &  98.0 & \textbf{58.0} & \textbf{39.2} \\ \hline
 \multirow{3}*{\textbf{\specialcell{ PN++\\ (SSG)}}} &  3D-Adv \cite{pcattack}  & 7.6 &  9.6 & 100 &  6.0 &  7.2 & 10.4 & 100 &  7.2 \\ & KNN \cite{robustshapeattack} &  6.4 &  9.2 & 100 &  6.4 &  6.8 &  7.6 & 100 &  6.0 \\ & AdvPC (Ours)  & \textbf{12.0} & \textbf{27.2} &  99.2 & \textbf{22.8} & \textbf{14.0} & \textbf{30.8} &  99.2 & \textbf{27.6} \\ \hline
 \multirow{3}*{\textbf{DGCNN}} &  3D-Adv \cite{pcattack}  & 9.2 & 11.2 & 31.2 & 100 &  9.6 & 12.8 & 30.4 & 100 \\ & KNN \cite{robustshapeattack} & 7.2 &  9.6 & 14.0 &  99.6 &  6.8 & 10.0 & 11.2 &  99.6 \\ & AdvPC (Ours)  & \textbf{19.6} & \textbf{46.0} & \textbf{64.4} &  94.8 & \textbf{32.8} & \textbf{48.8} & \textbf{64.4} &  97.2 \\  
 \bottomrule
\end{tabular}}
\caption{\small \textbf{Transferability of Attacks under $\ell_\infty$ Norms}: We use norm-budgets (max $\ell_\infty$ norm allowed in the perturbation) of $\epsilon_\infty = 0.18$ and $\epsilon_\infty = 0.45$ . All the reported results are the untargeted Attack Success Rate (higher numbers are better attacks). \textbf{Bold} numbers indicate the most transferable attacks. Our attack consistently achieves better transferability than the other attacks for all networks, especially on DGCNN \cite{dgcn}. For reference, the classification accuracies on unperturbed samples for networks PN, PN++(MSG), PN++(SSG) and DGCNN are  92.8\%, 91.5\%, 91.5\%, and 93.7\%, respectively.
}
\label{tbl:transfer-Inf}
\end{table}

\begin{table}[]
\footnotesize
\setlength{\tabcolsep}{0.5pt} %
\renewcommand{\arraystretch}{1.1} %
\centering
\vspace{-2pt}
\resizebox{\textwidth}{!}{%
\begin{tabular}{cc|cccc|cccc} 
\toprule
- & - &\multicolumn{4}{c|}{$\epsilon_2 = 1.8$} &\multicolumn{4}{c}{$\epsilon_2 = 4.0$}\\
\specialcell{\textbf{Victim}\\\textbf{Network}} & \textbf{Attack}  &\textbf{PN} & \textbf{\specialcell{ PN++ \\(MSG)}} & \textbf{\specialcell{ PN\scriptsize++\\ (SSG)}} & \textbf{DGCNN}  &\textbf{PN} & \textbf{\specialcell{ PN++ \\(MSG)}} & \textbf{\specialcell{ PN++\\ (SSG)}} & \textbf{DGCNN}   \\
\midrule
 \multirow{3}*{\textbf{PN}} & 3D-Adv \cite{pcattack}  &  100 &  8.4 &  8.8 &  7.2 & 100 &  7.6 &  9.6 &  6.0 \\
 & KNN \cite{robustshapeattack} & 100 &  9.2 &  8.4 &  7.2 & 100 &  8.8 &  8.8 &  7.6 \\
 & AdvPC (Ours)  &  98.0 & \textbf{17.2} & \textbf{28.0} & \textbf{22.0} &  98.8 & \textbf{16.0} & \textbf{19.6} & \textbf{15.6} \\ \hline
 \multirow{3}*{\textbf{\specialcell{ PN++ \\(MSG)}}} & 3D-Adv \cite{pcattack} & 6.8 & 100 & 32.4 & 14.8 &  7.6 & 100 & 28.0 & 14.0 \\ & KNN \cite{robustshapeattack} &  7.2 & 100 & 22.8 &  8.4 &  6.8 & 100 & 22.8 &  8.4 \\ & AdvPC (Ours)  &  \textbf{13.2} &  94.8 & \textbf{53.2} & \textbf{33.2} & \textbf{22.8} &  98.4 & \textbf{55.2} & \textbf{44.0} \\ \hline
 \multirow{3}*{\textbf{\specialcell{ PN++\\ (SSG)}}} &  3D-Adv \cite{pcattack}  & 6.8 &  8.8 & 100 &  8.0 &  7.2 & 10.4 & 100 &  7.2 \\ & KNN \cite{robustshapeattack} &  6.8 &  8.8 & 100 &  7.6 &  6.4 &  8.4 & 100 &  6.4 \\ & AdvPC (Ours)  & \textbf{10.8}\textbf{} & \textbf{27.6} &  96.4 & \textbf{26.8} & \textbf{10.0} & \textbf{25.6} &  98.8 & \textbf{23.6} \\ \hline
 \multirow{3}*{\textbf{DGCNN}} &  3D-Adv \cite{pcattack}  & 10.8 & 14.4 & 39.6 & 100 & 10.8 & 14.0 & 32.4 & 100 \\ & KNN \cite{robustshapeattack} &  7.2 & 11.2 & 13.6 & 100 &  6.8 &  8.4 & 11.2 &  99.6 \\ & AdvPC (Ours)  & \textbf{20.8} & \textbf{32.4} & \textbf{52.4} &  85.2 & \textbf{38.8} & \textbf{48.4} & \textbf{63.2} &  98.8 \\  
 \bottomrule
\end{tabular}}
\caption{\small \textbf{Transferability of Attacks under $\ell_2$ Norms}: We use norm-budgets (max $\ell_2$ norm allowed in the perturbation) of $\epsilon_2 = 1.8$ and $\epsilon_2 = 4.0$ . All the reported results are the untargeted Attack Success Rate (higher numbers are better attacks). \textbf{Bold} numbers indicate the most transferable attacks. Our attack consistently achieves better transferability than the other attacks for all networks, especially on DGCNN \cite{dgcn}. For reference, the classification accuracies on unperturbed samples for networks PN, PN++(MSG), PN++(SSG) and DGCNN are  92.8\%, 91.5\%, 91.5\%, and 93.7\%, respectively.
}
\label{tbl:transfer-2}
\end{table}

\clearpage
\subsection{Transferability on Different Norms} \label{sec:sup-transfer-all}
\begin{figure}[]
\tabcolsep=0.03cm
\begin{tabular}{c|ccc}
\toprule
\textbf{victim network} & \multicolumn{3}{c}{\textbf{transfer network}}\\
\hline 
\includegraphics[width=0.25\linewidth]{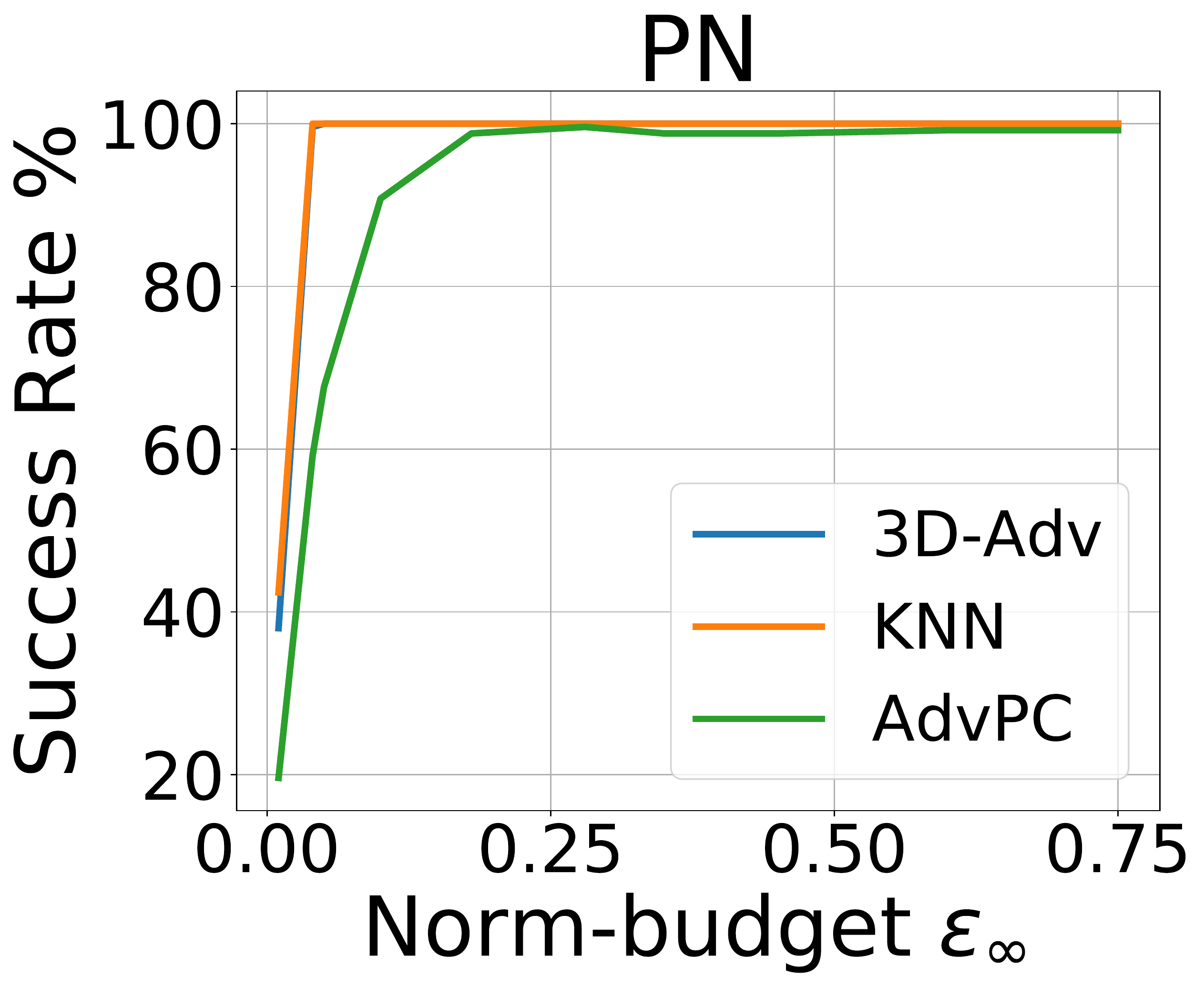} &
 \includegraphics[width=0.25\linewidth]{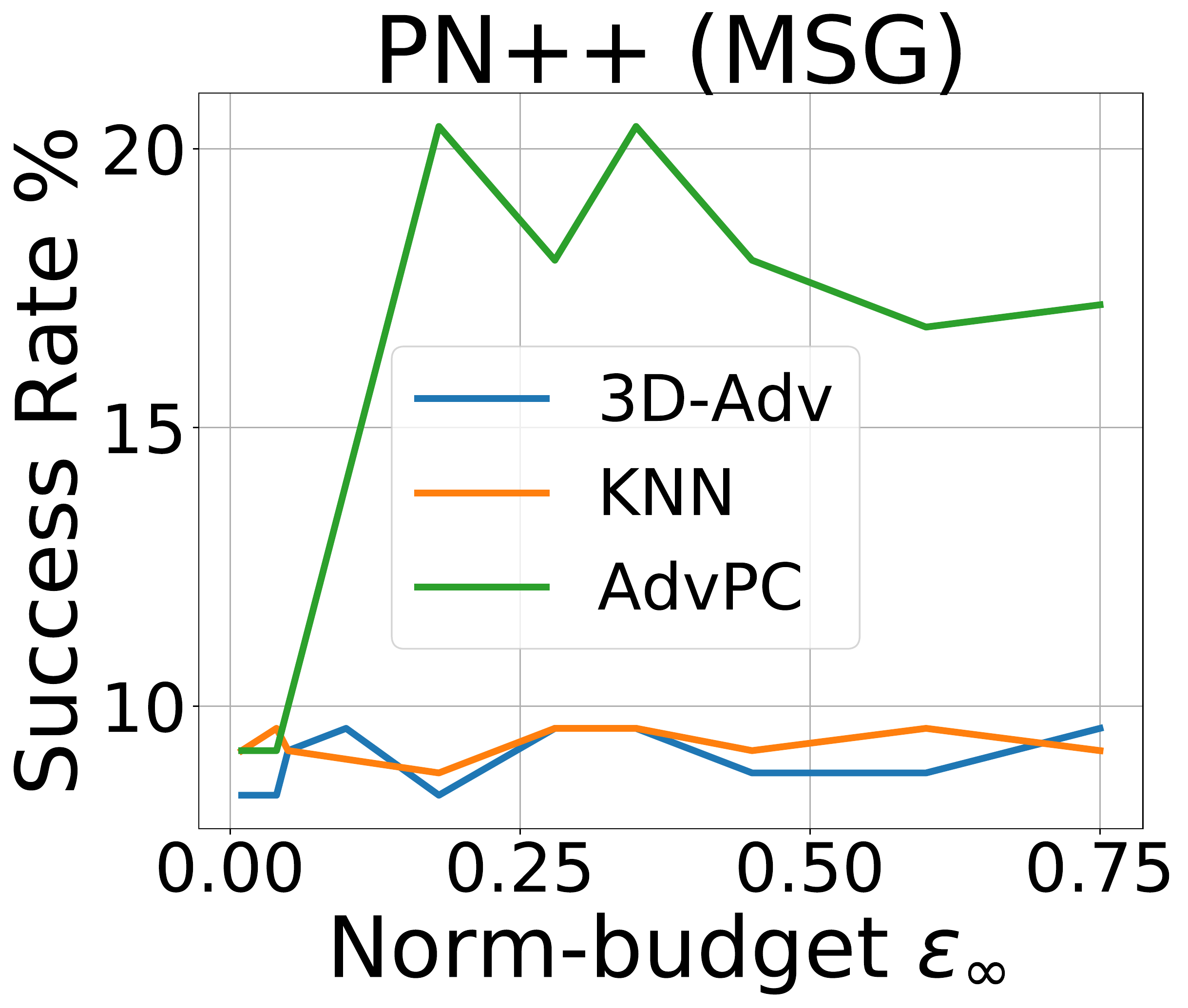} &
\includegraphics[width=0.25\linewidth]{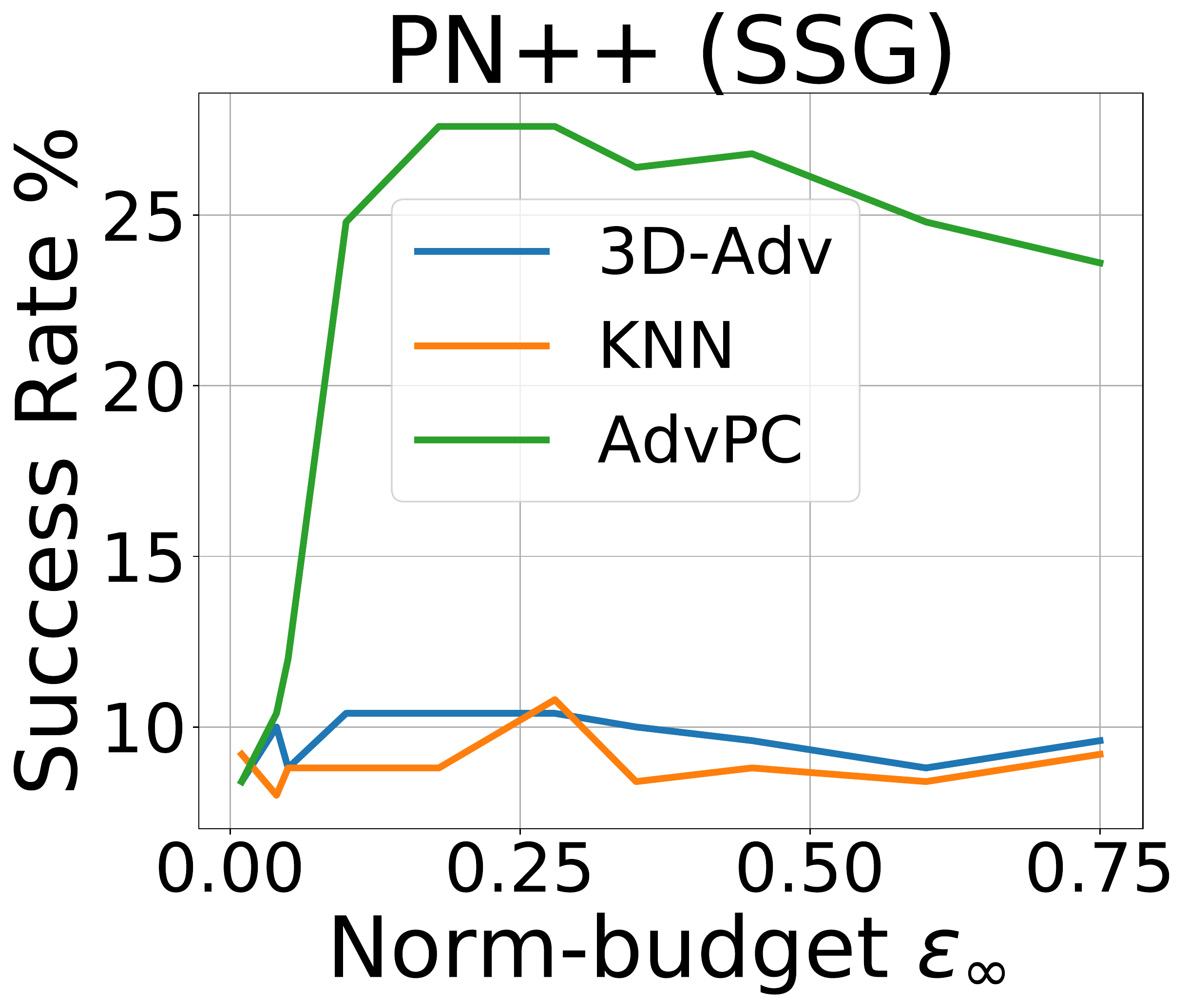} &
\includegraphics[width=0.25\linewidth]{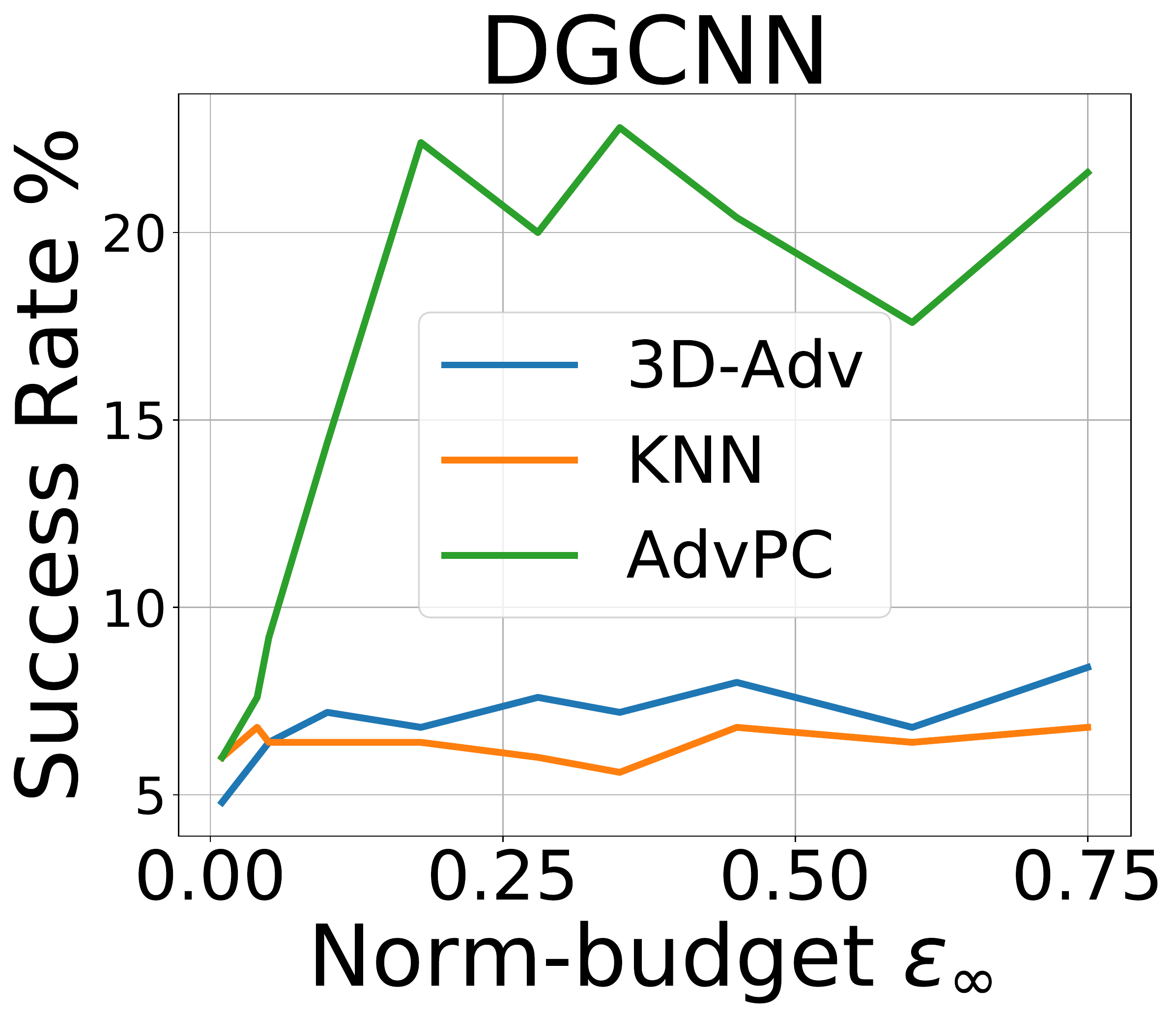} \\
\hline 
\includegraphics[width=0.25\linewidth]{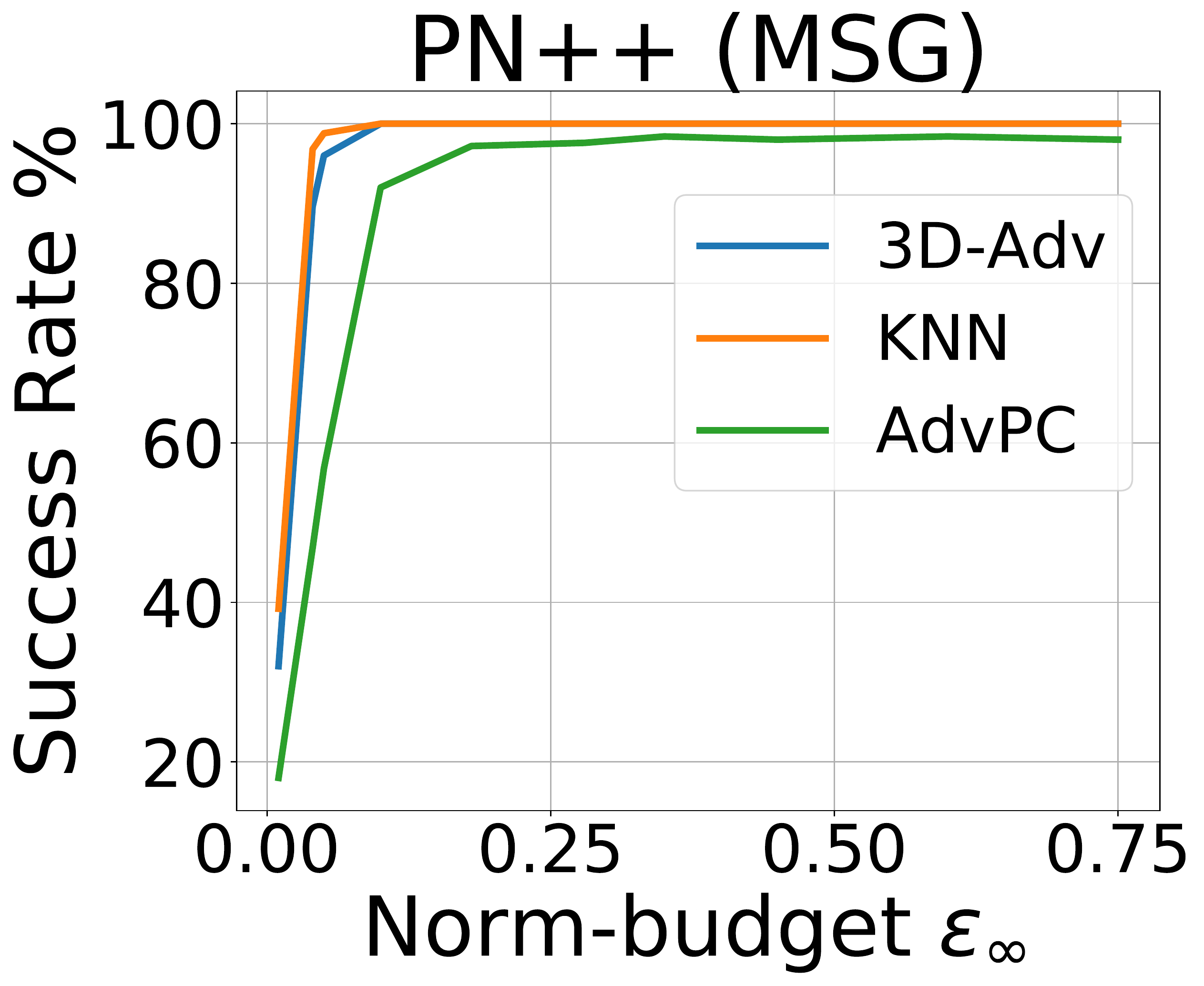} &
 \includegraphics[width=0.25\linewidth]{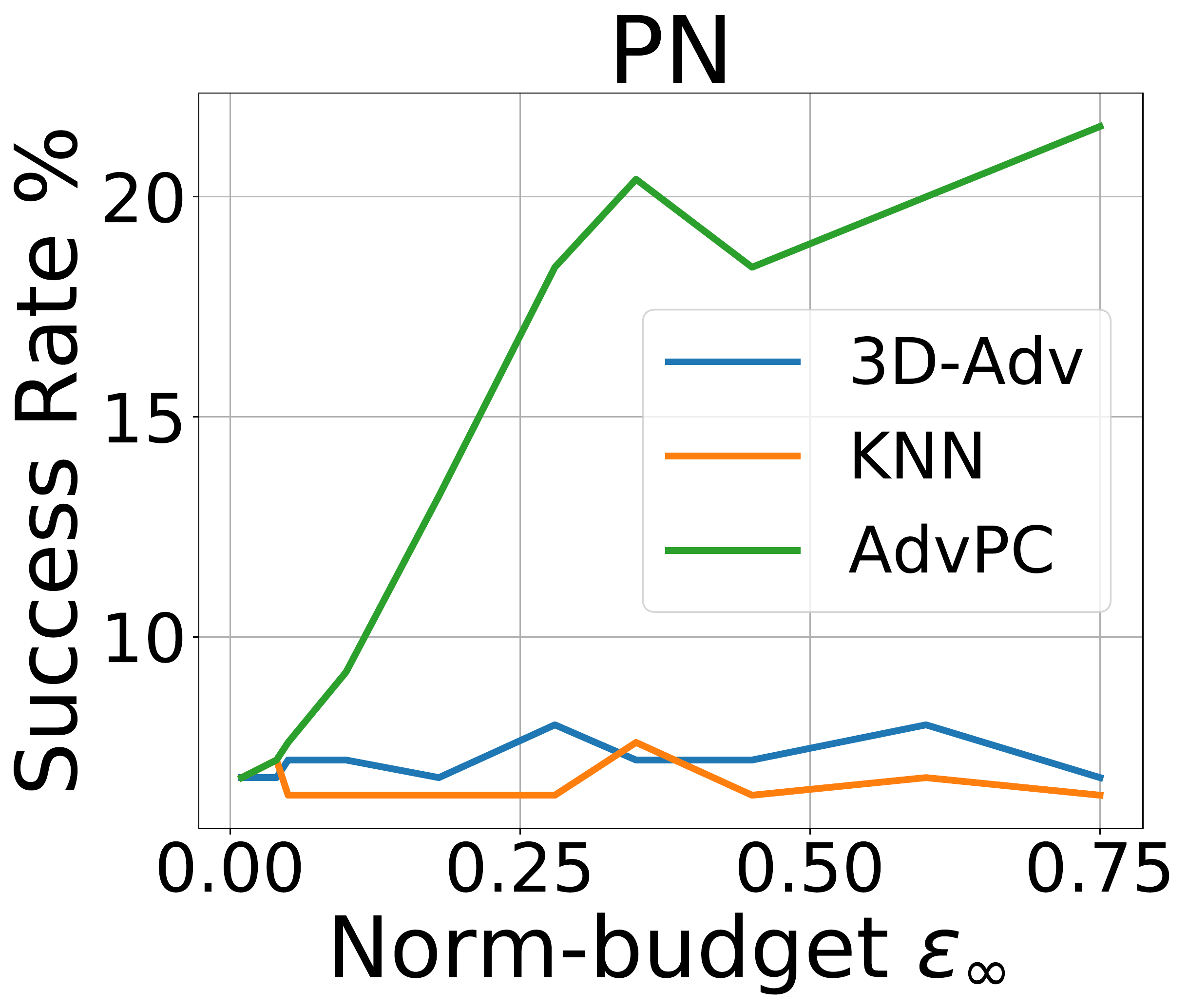} &
\includegraphics[width=0.25\linewidth]{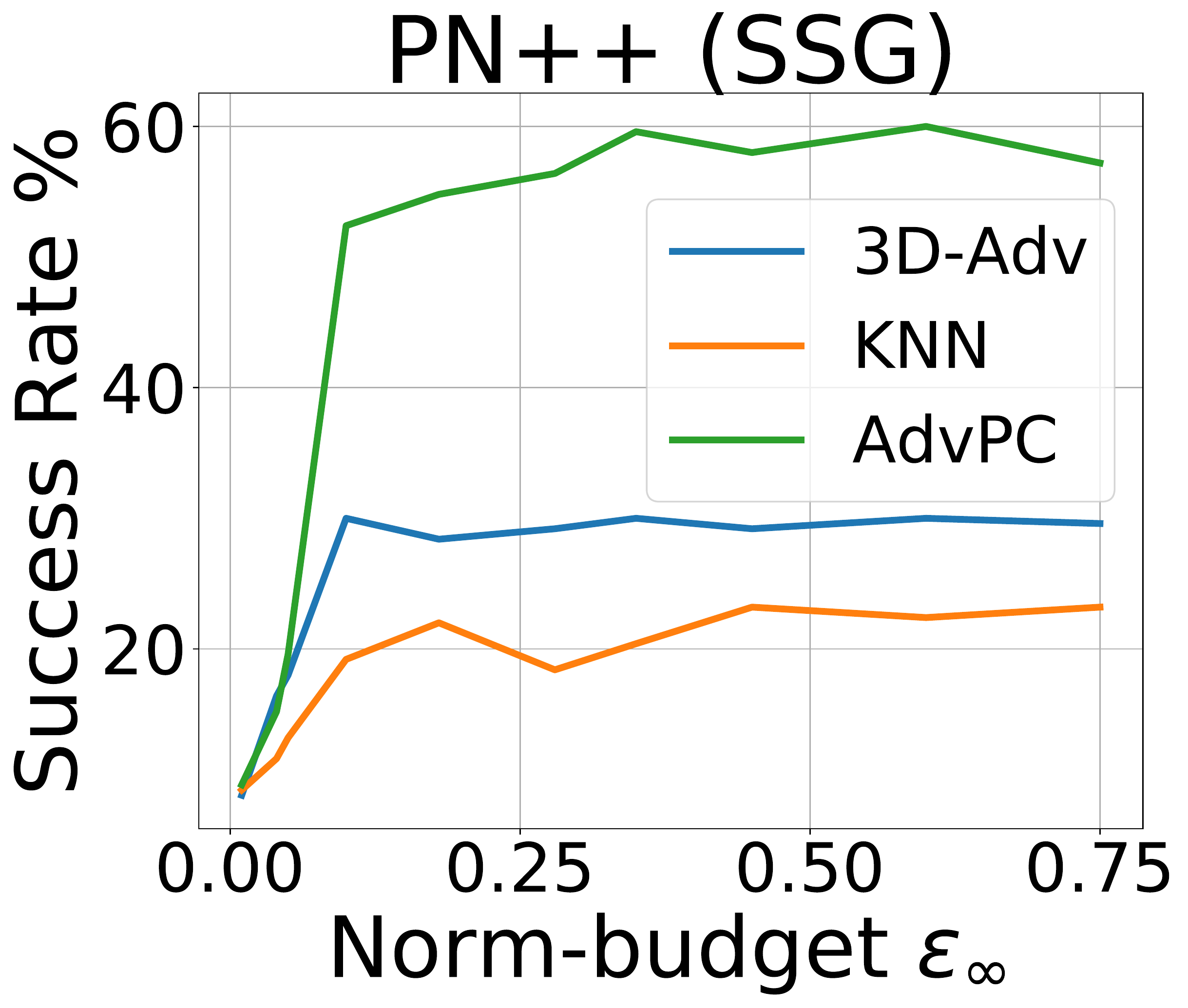} &
\includegraphics[width=0.25\linewidth]{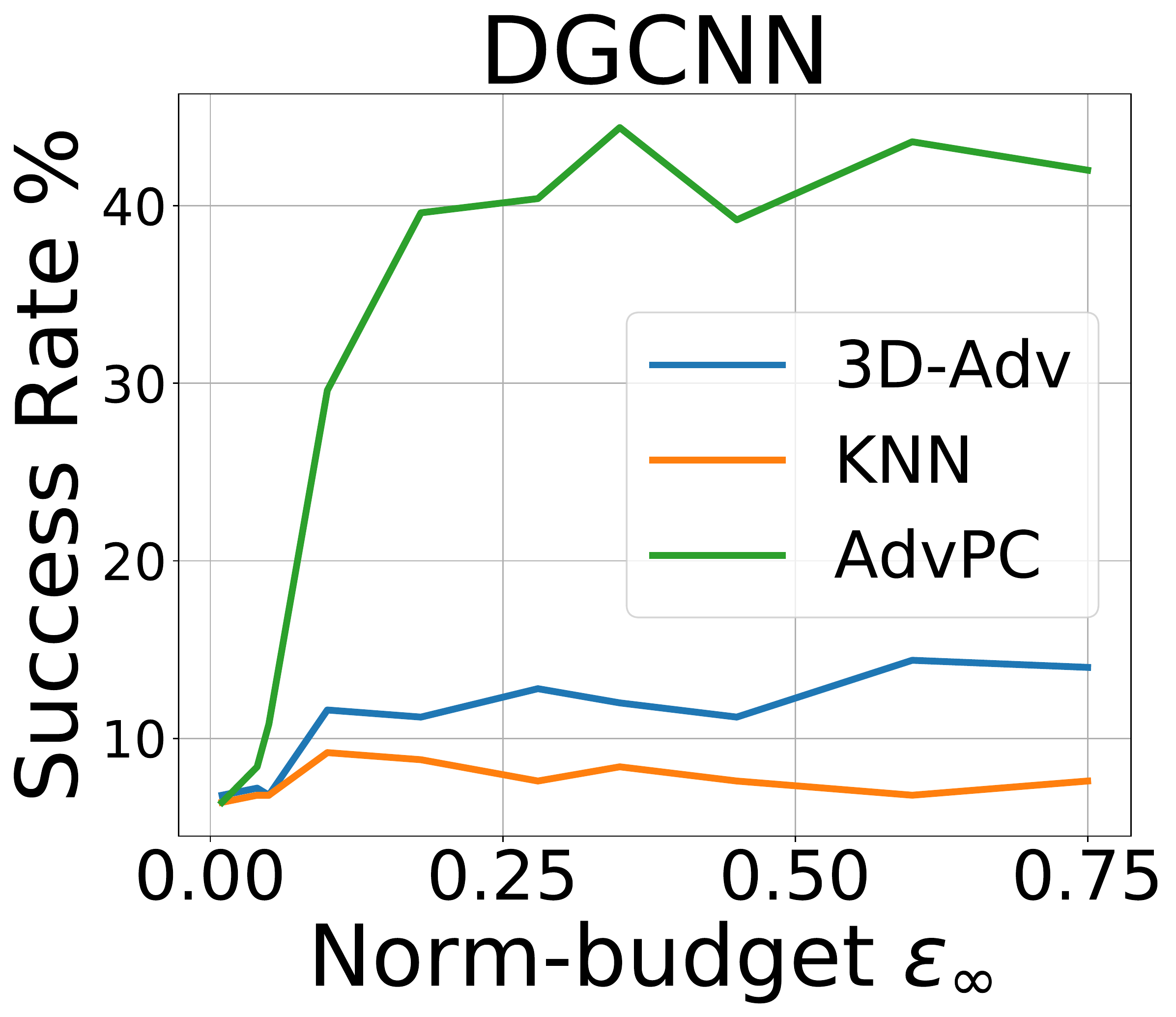}\\
\hline 
\includegraphics[width=0.25\linewidth]{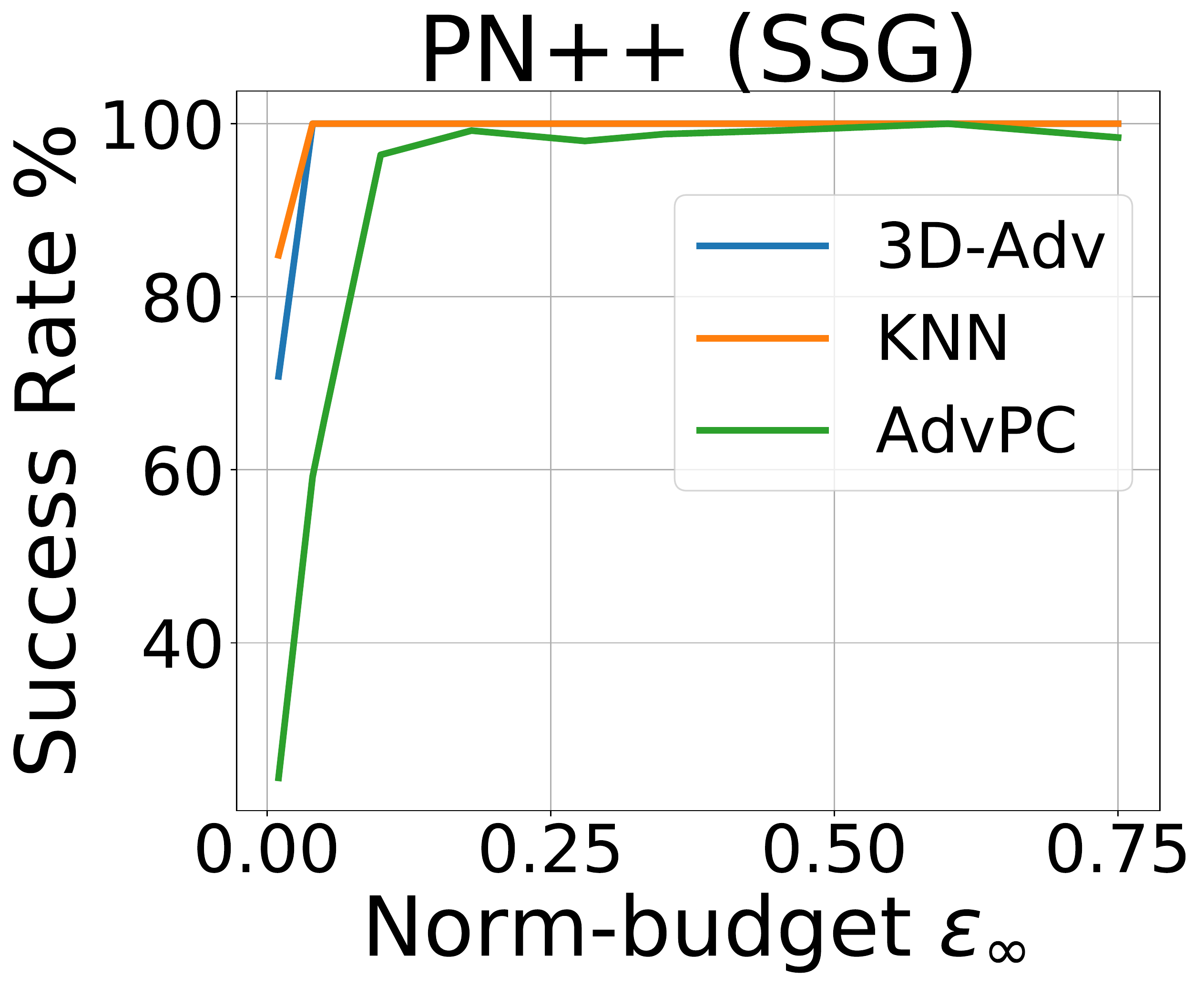} &
 \includegraphics[width=0.25\linewidth]{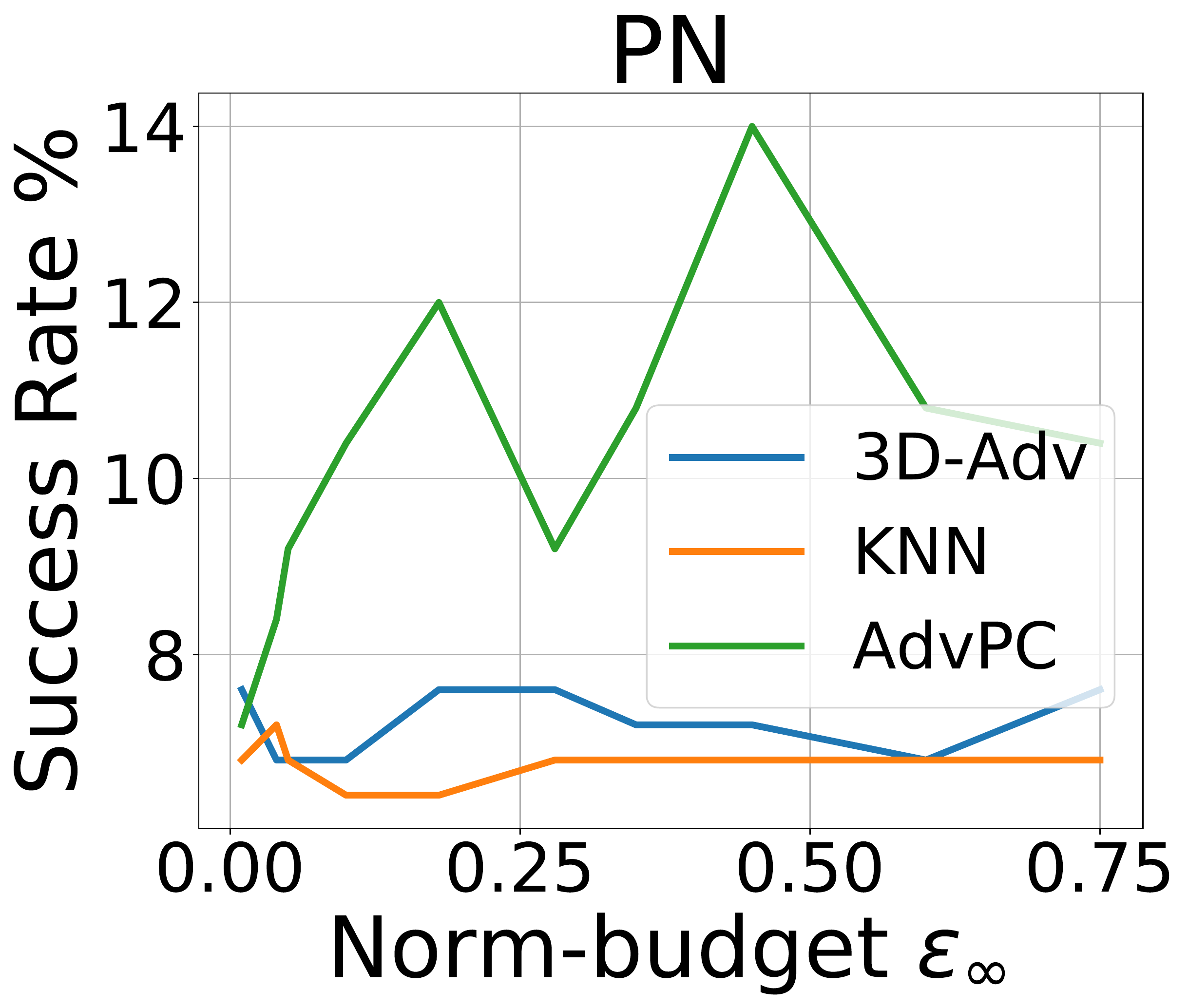} &
\includegraphics[width=0.25\linewidth]{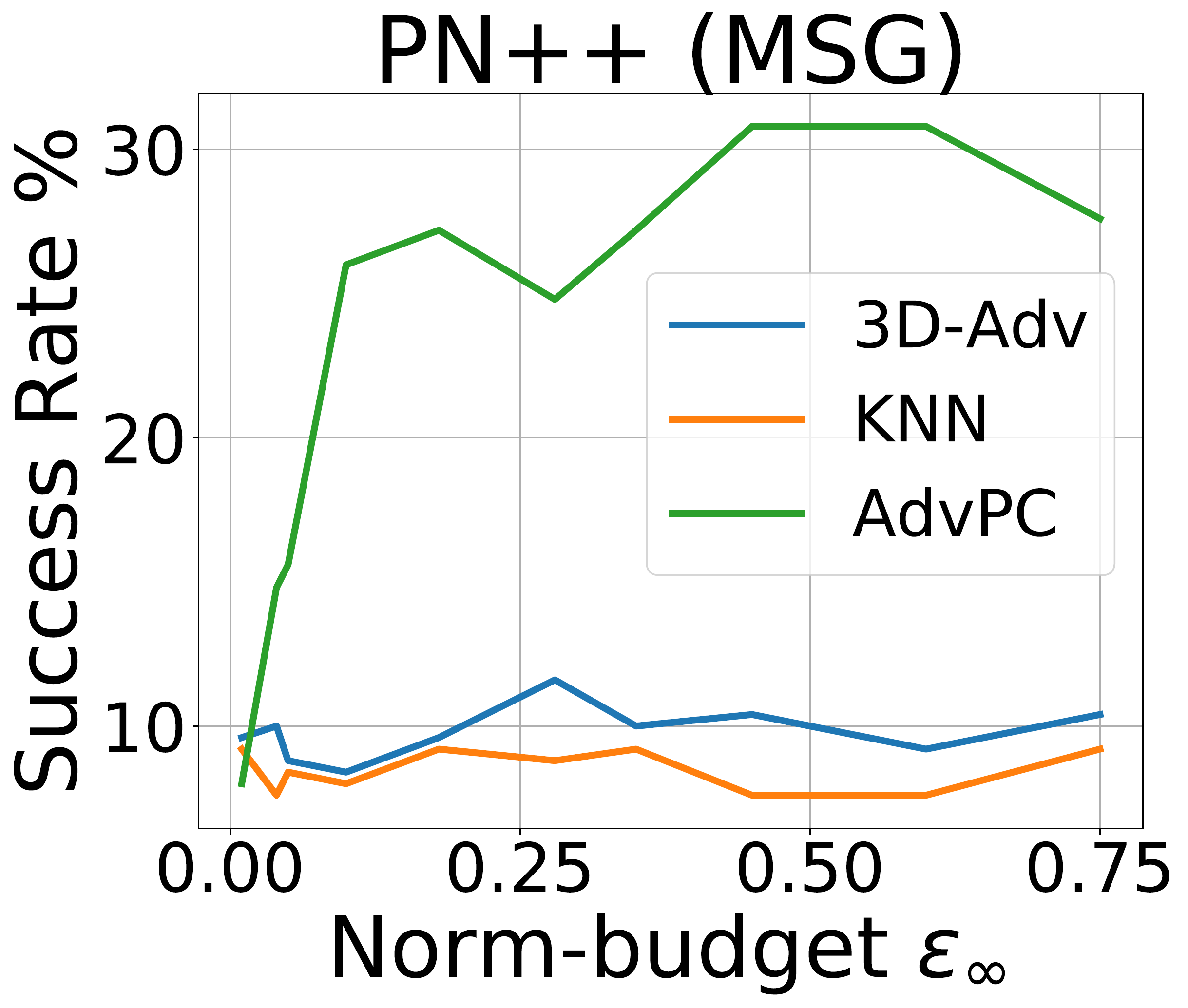} &
\includegraphics[width=0.25\linewidth]{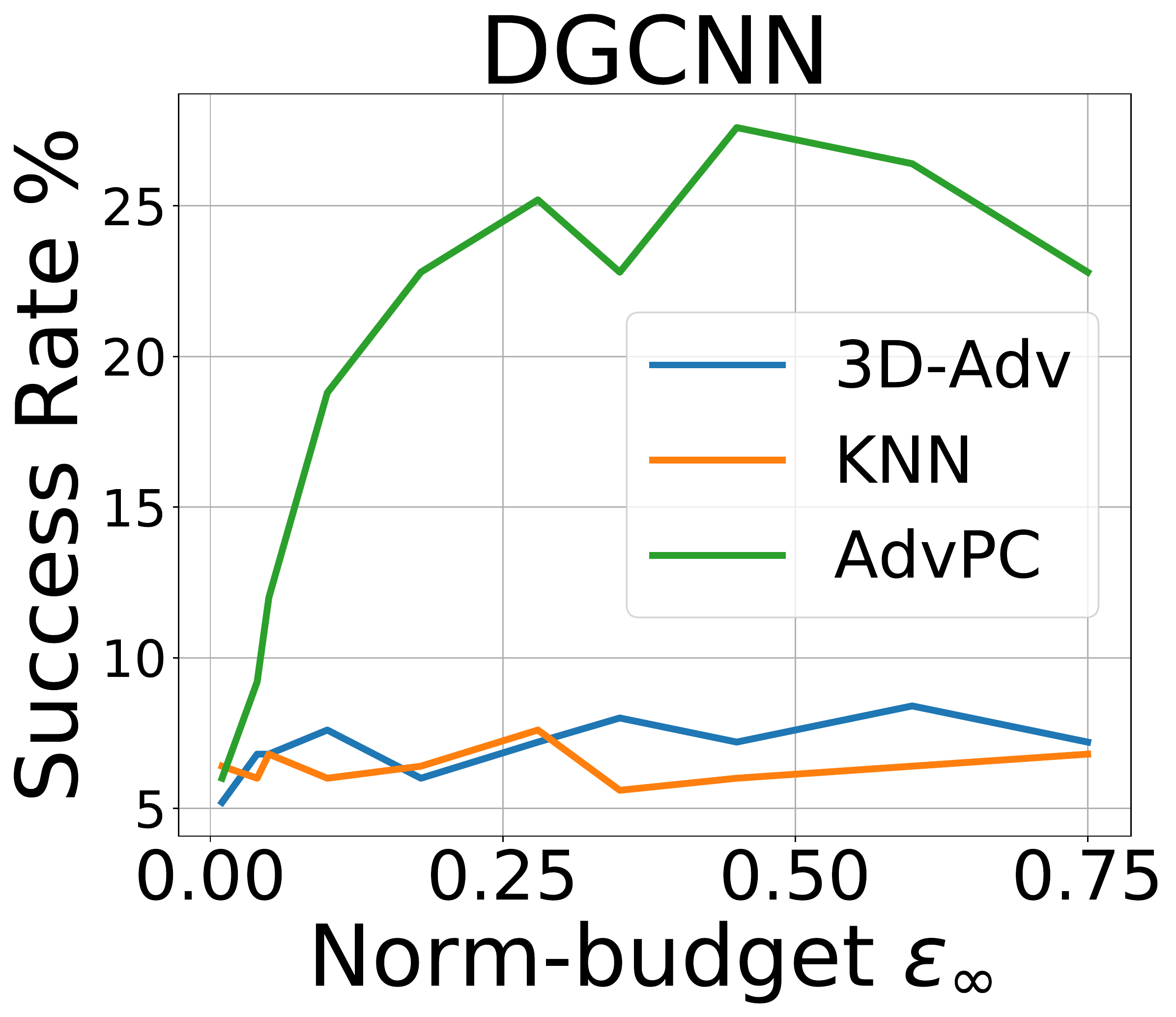} \\
\hline 
\includegraphics[width=0.25\linewidth]{images/transferbility/DGCNN_Succ_ECCV.pdf} &
 \includegraphics[width=0.25\linewidth]{images/transferbility/PN_Succ_ECCV.pdf} &
\includegraphics[width=0.25\linewidth]{images/transferbility/PN2_Succ_ECCV.pdf} &
\includegraphics[width=0.25\linewidth]{images/transferbility/PN1_Succ_ECCV.pdf} \\
\bottomrule

\end{tabular}

\caption{\small \textbf{Transferability Across Different $\epsilon_\infty$ Norm-Budgets}: Here, the attacks are optimized using different \textbf{$\epsilon_\infty$} norm-budgets. We report the attack success on all victim networks and the success of these attacks on each transfer network. We note that our AdvPC transfers better to the other networks across different  $\epsilon_\infty$ as compared to the baselines 3D-adv\cite{pcattack} and KNN attack \cite{robustshapeattack}.}
\label{fig:sup-transferbility-2}
\vspace{-8pt}
\end{figure}

\begin{figure}[]
\tabcolsep=0.03cm
\begin{tabular}{c|ccc}
\toprule
\textbf{victim network} & \multicolumn{3}{c}{\textbf{transfer network}}\\
\hline 
\includegraphics[width=0.24\linewidth]{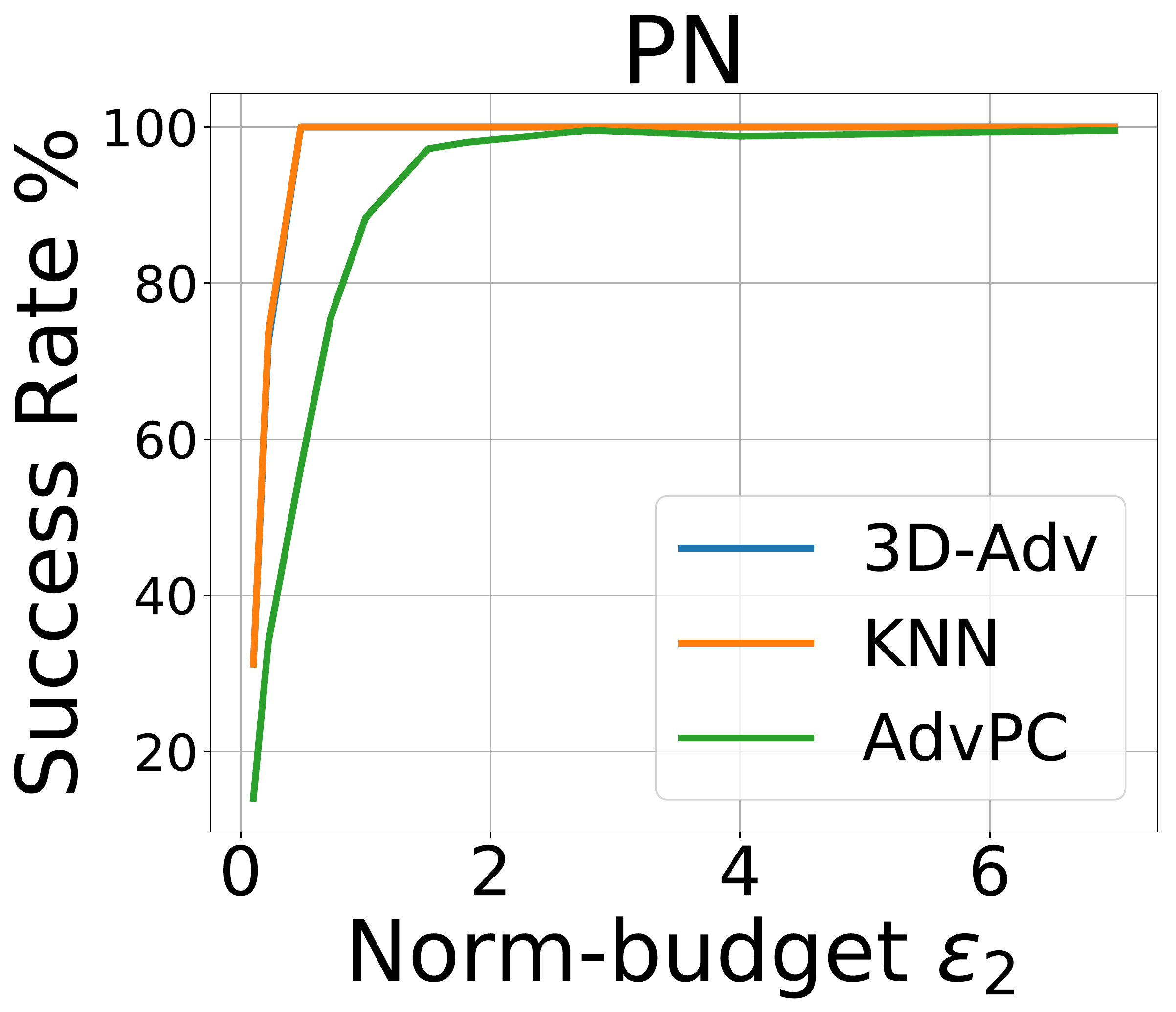} &
 \includegraphics[width=0.25\linewidth]{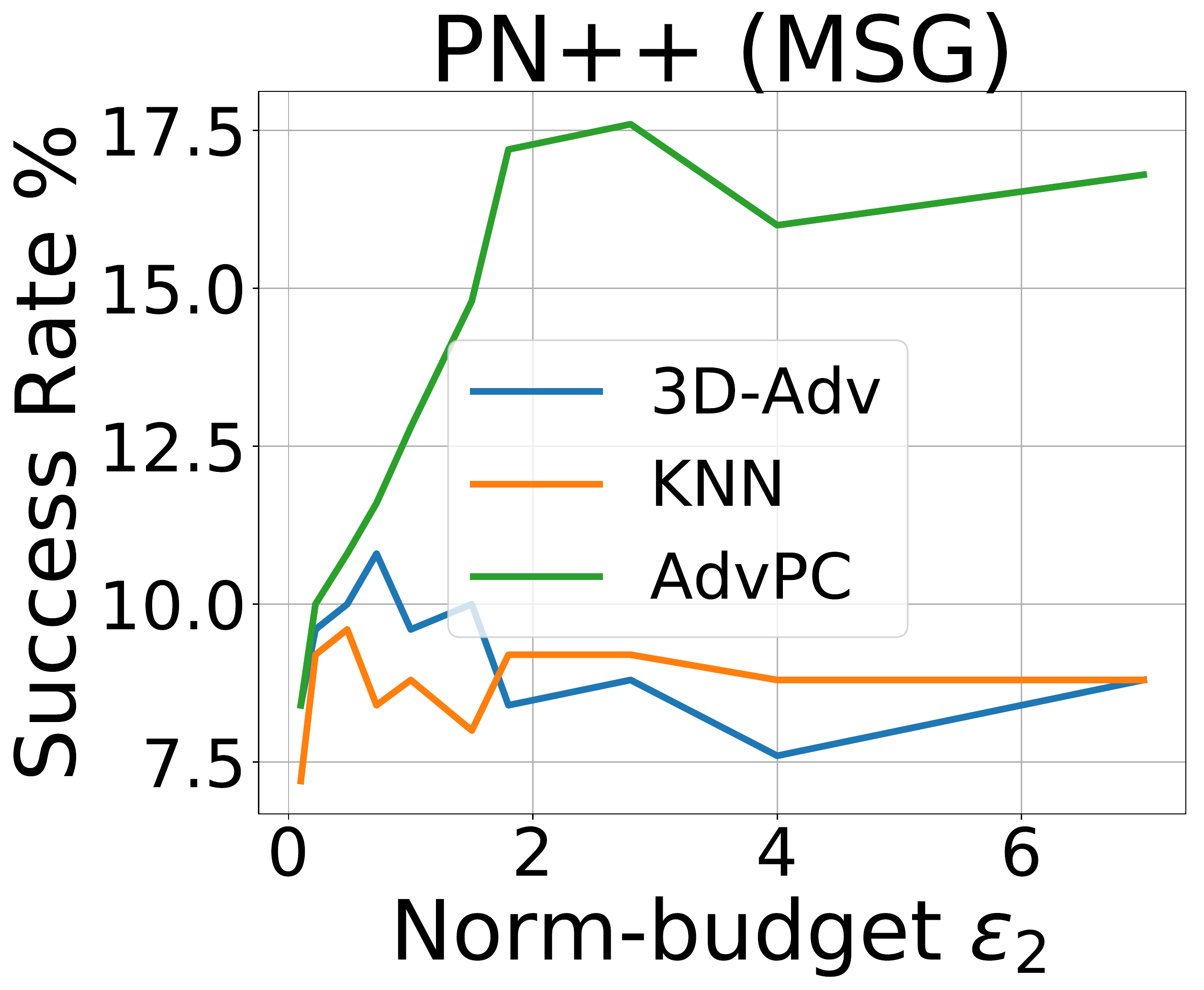} &
\includegraphics[width=0.25\linewidth]{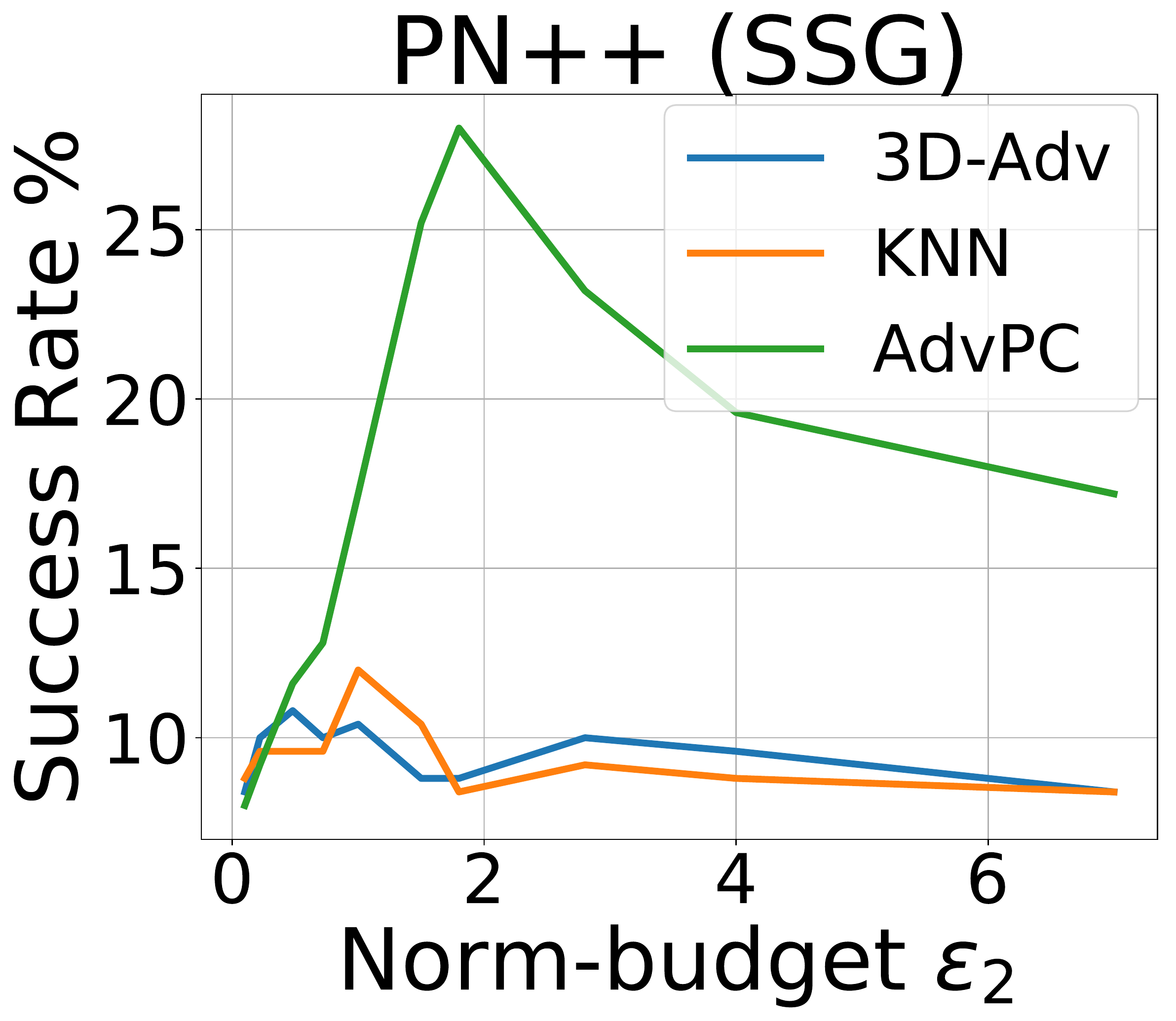} &
\includegraphics[width=0.25\linewidth]{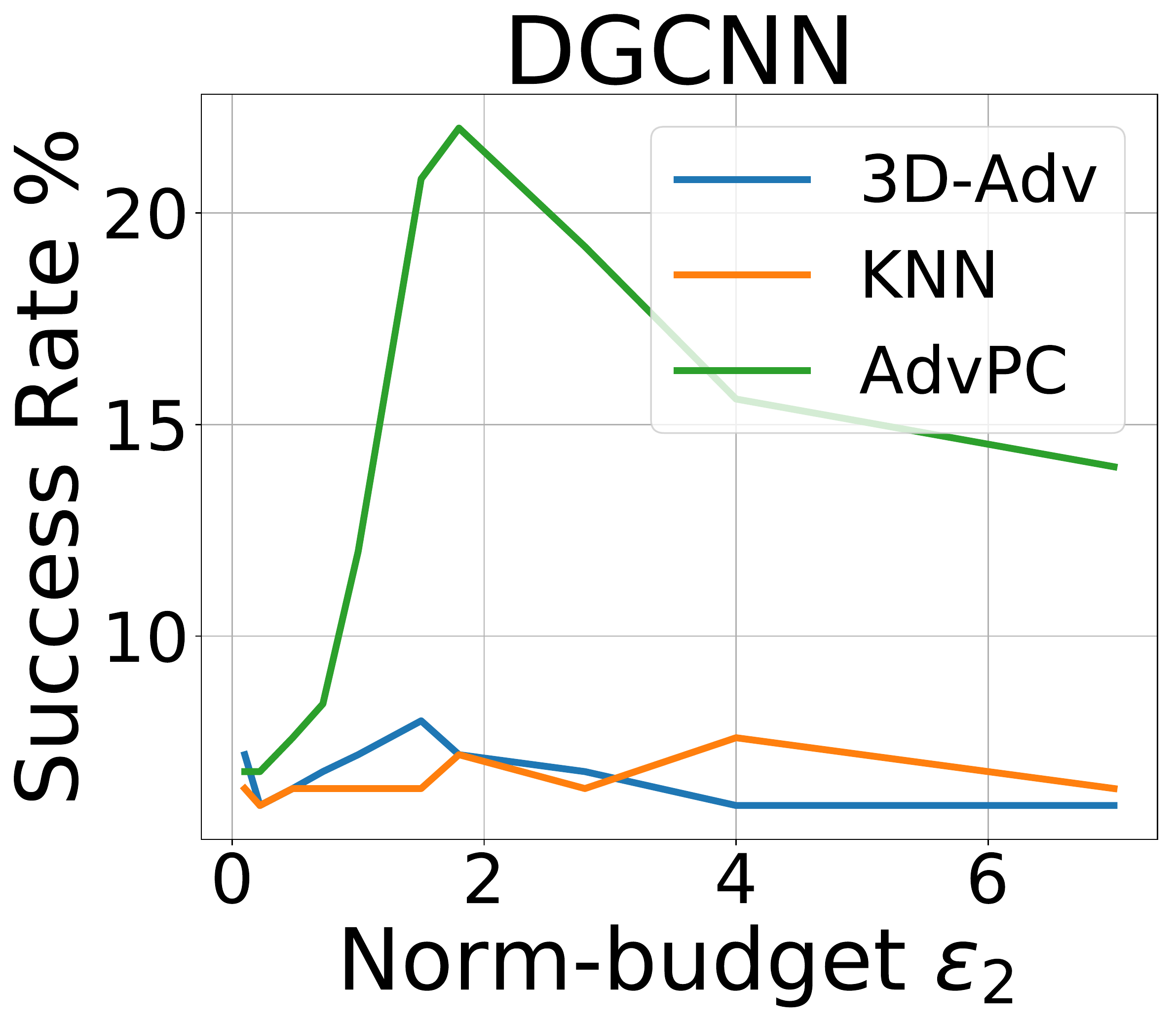} \\
\hline 
\includegraphics[width=0.24\linewidth]{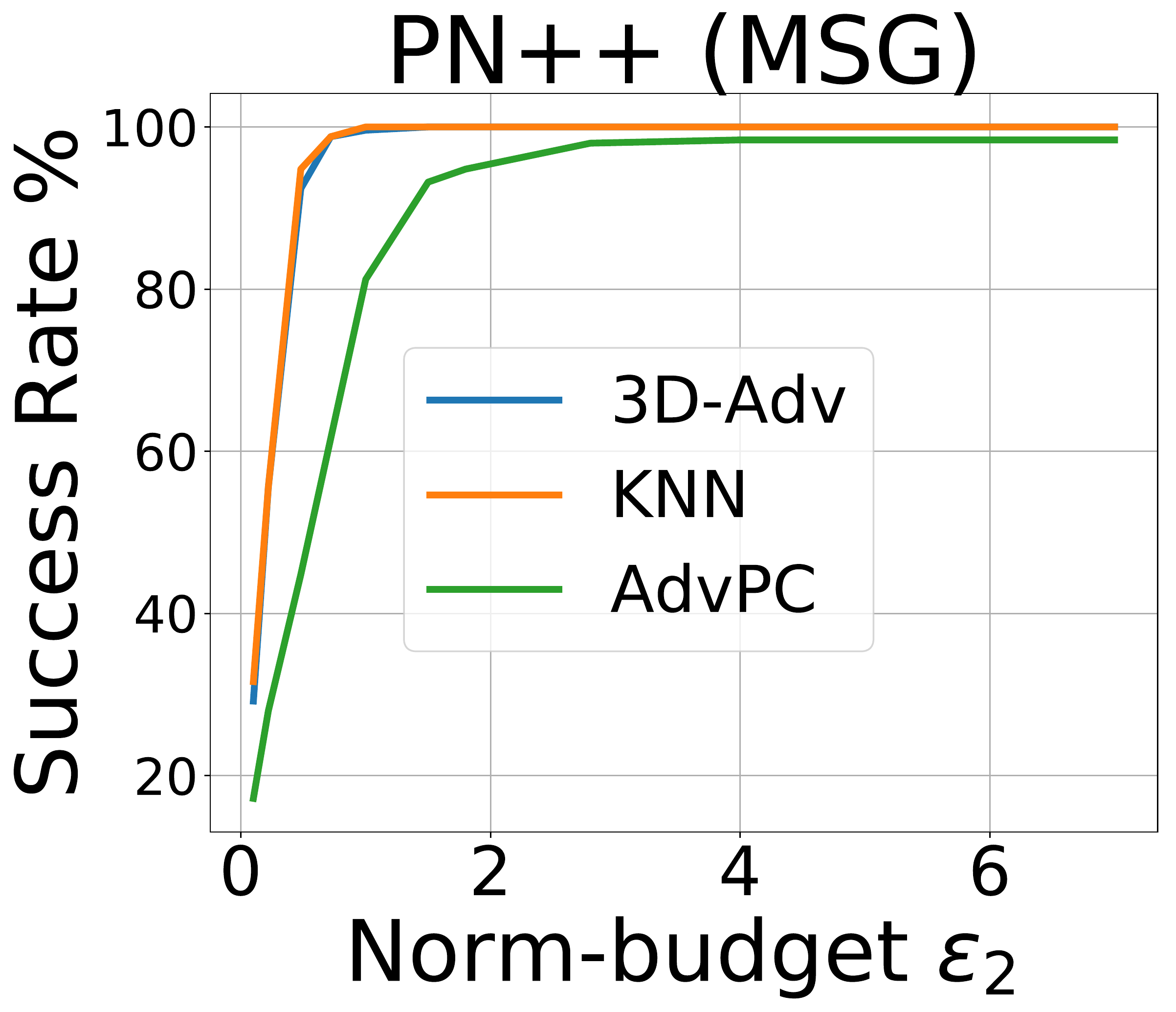} &
 \includegraphics[width=0.25\linewidth]{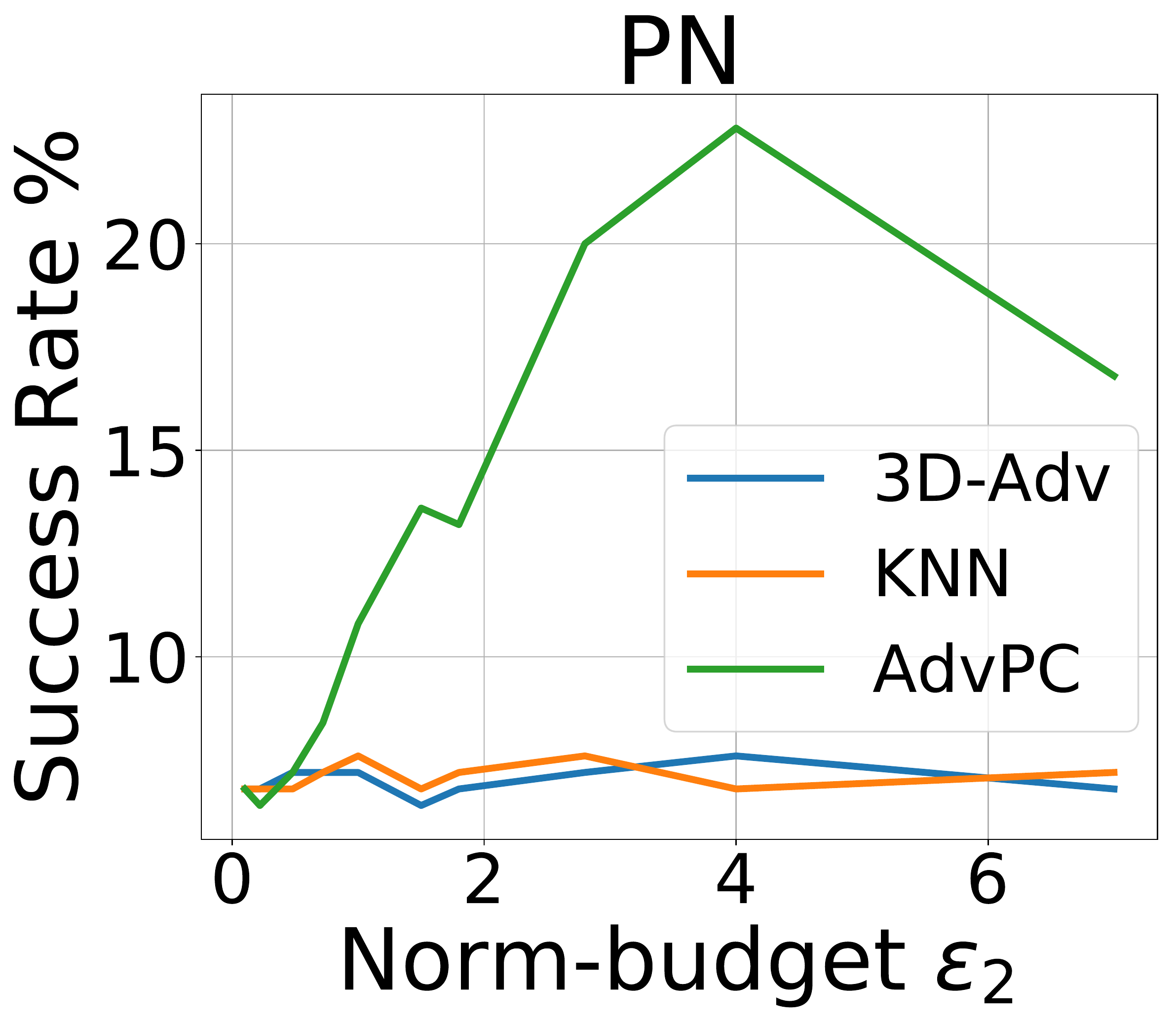} &
\includegraphics[width=0.25\linewidth]{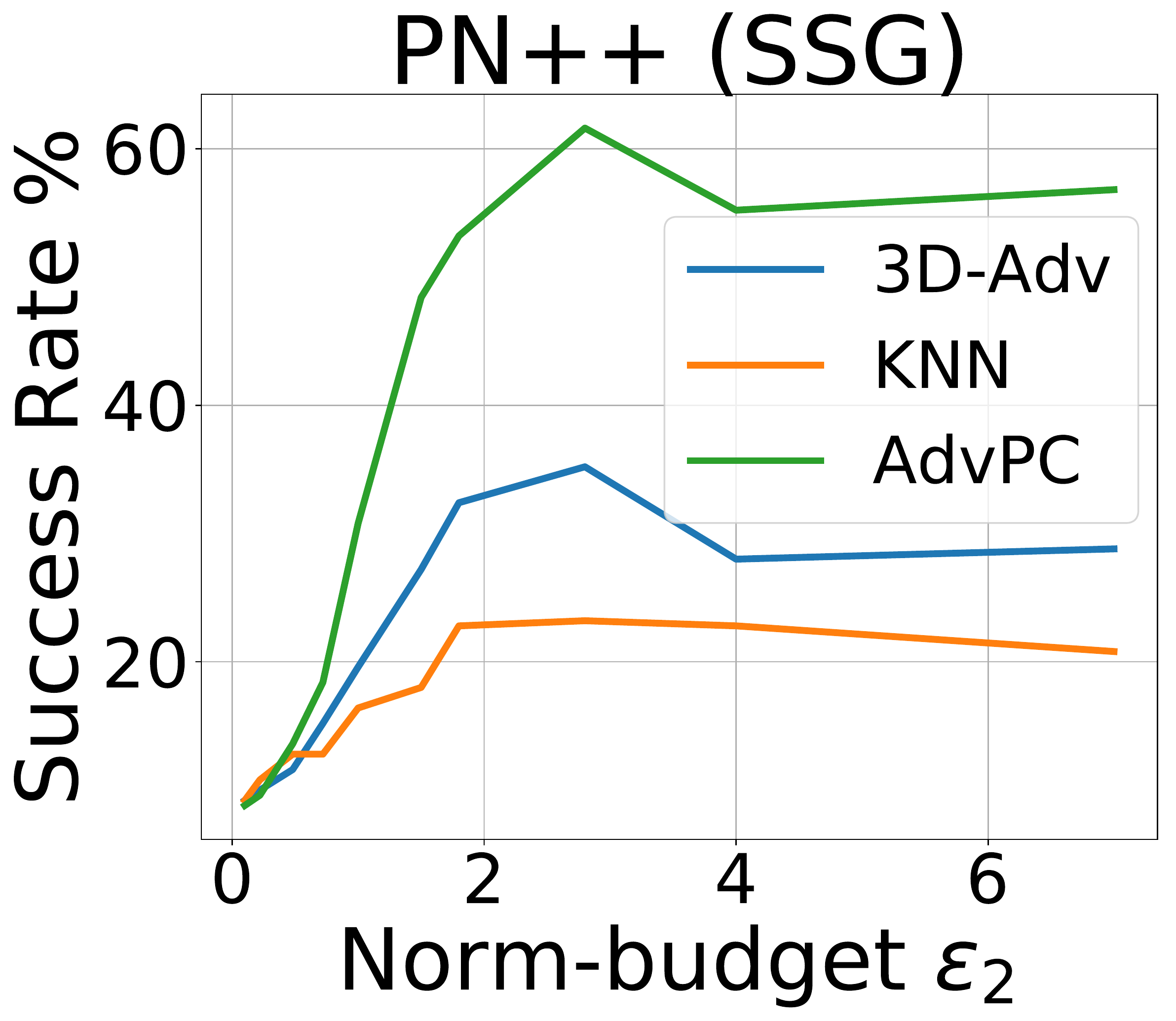} &
\includegraphics[width=0.25\linewidth]{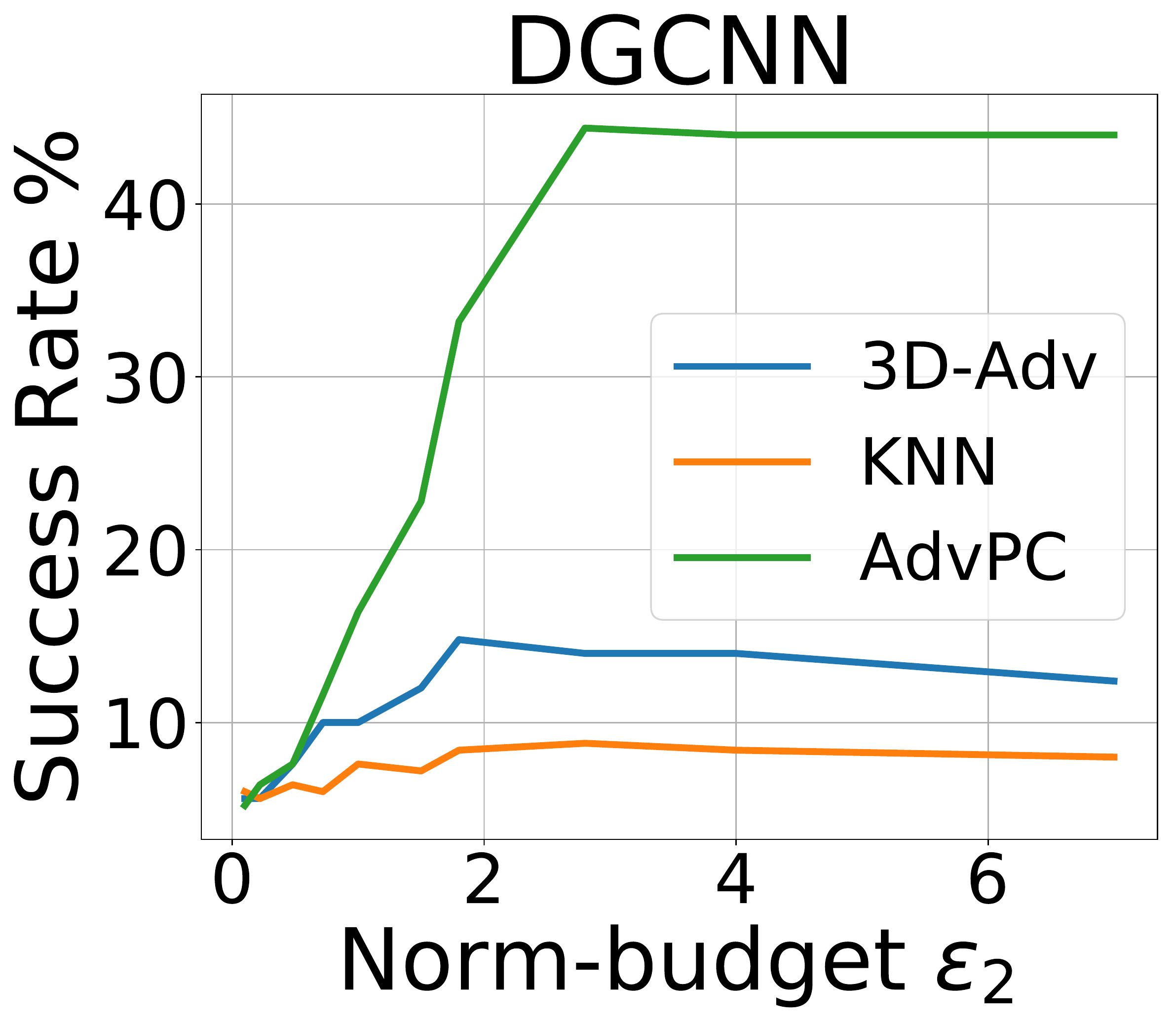}\\
\hline 
\includegraphics[width=0.24\linewidth]{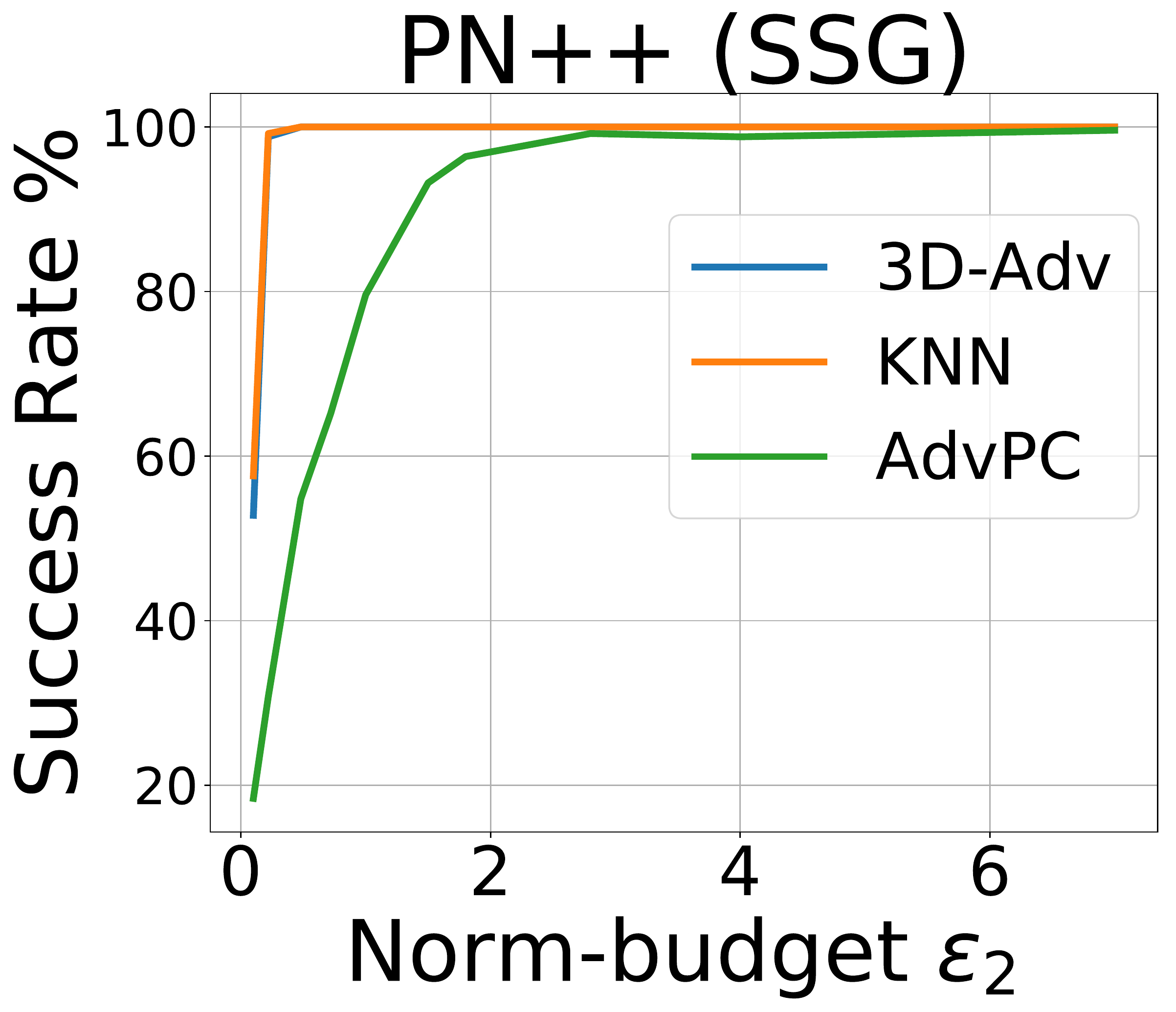} &
 \includegraphics[width=0.25\linewidth]{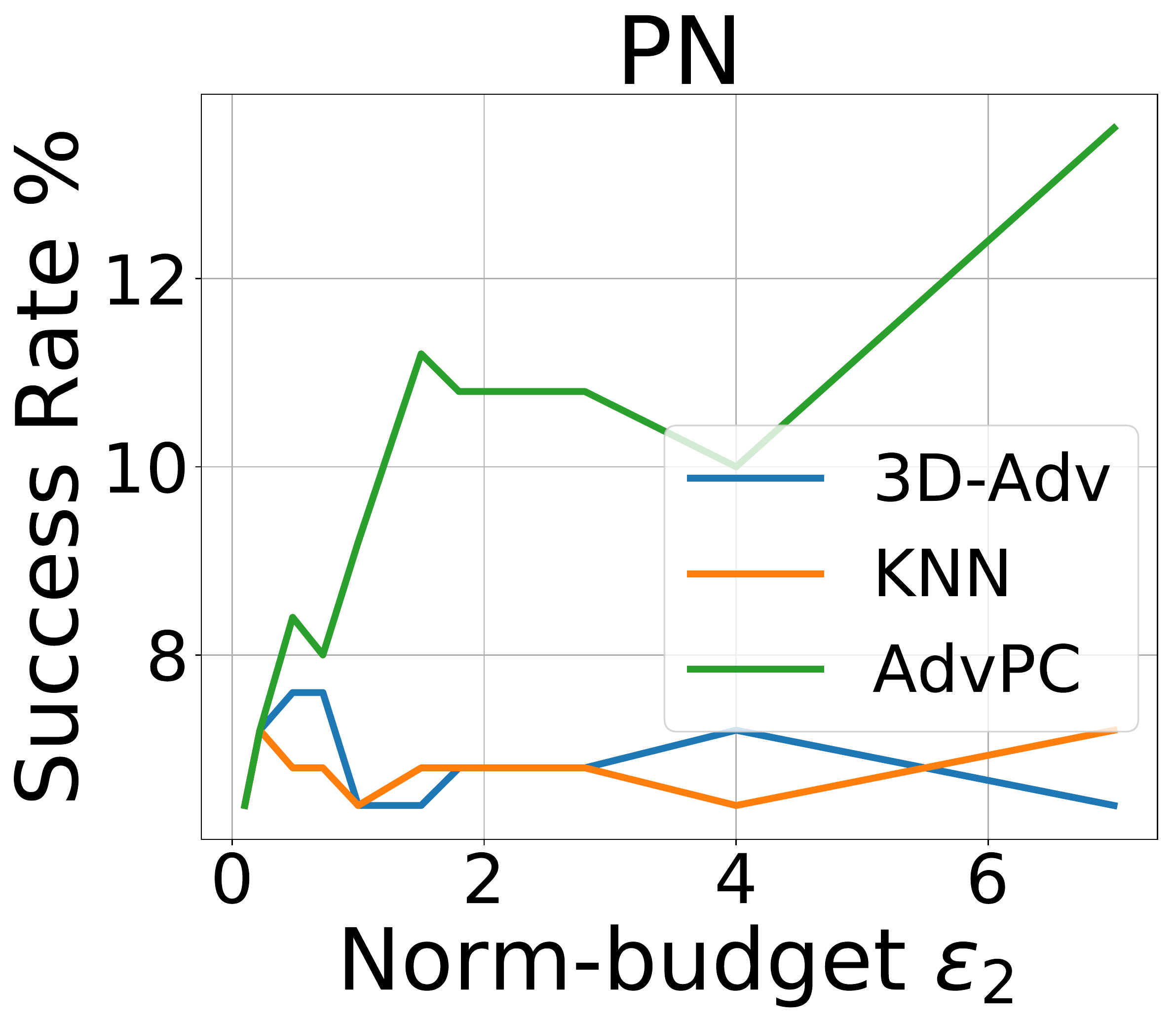} &
\includegraphics[width=0.25\linewidth]{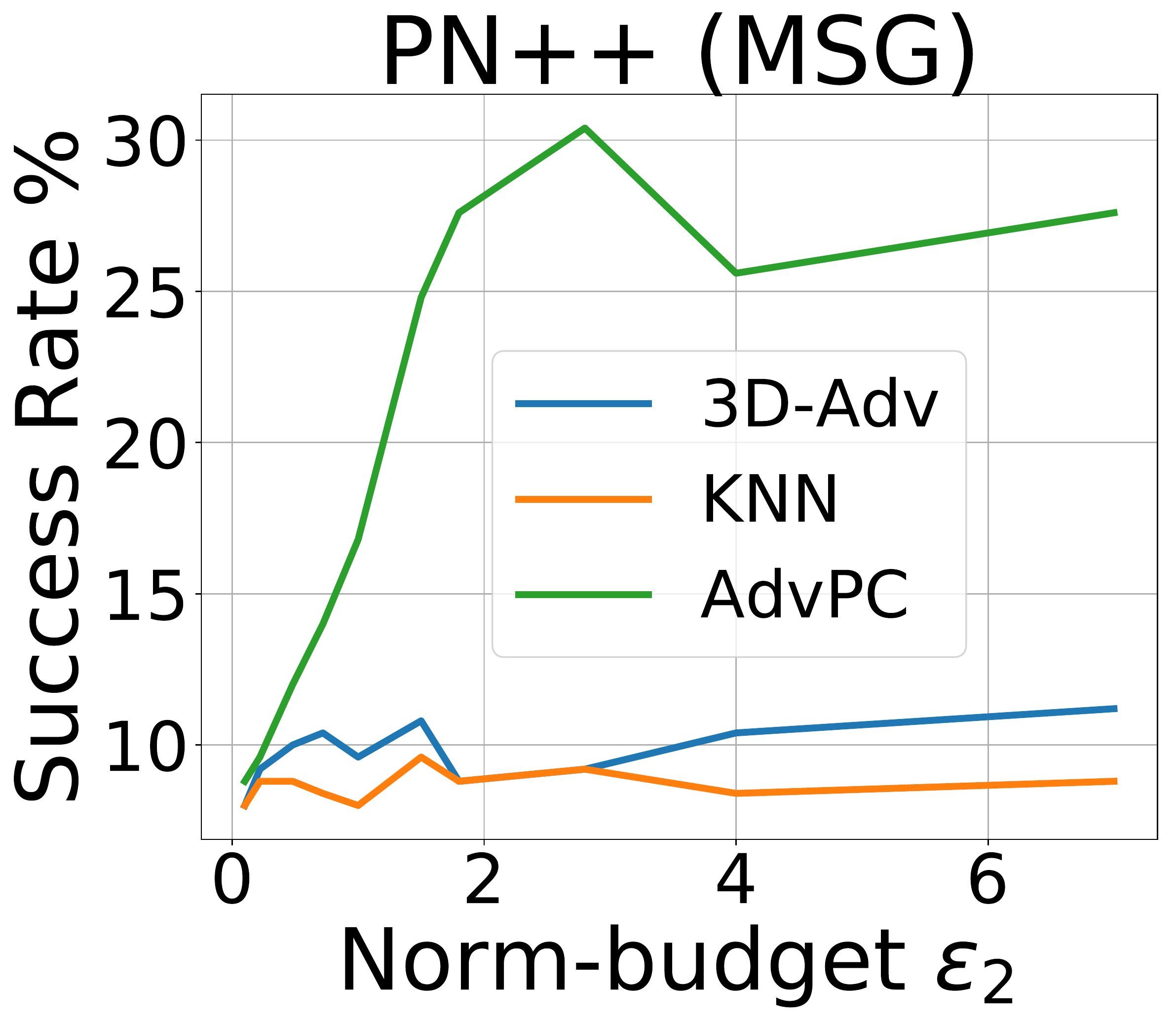} &
\includegraphics[width=0.25\linewidth]{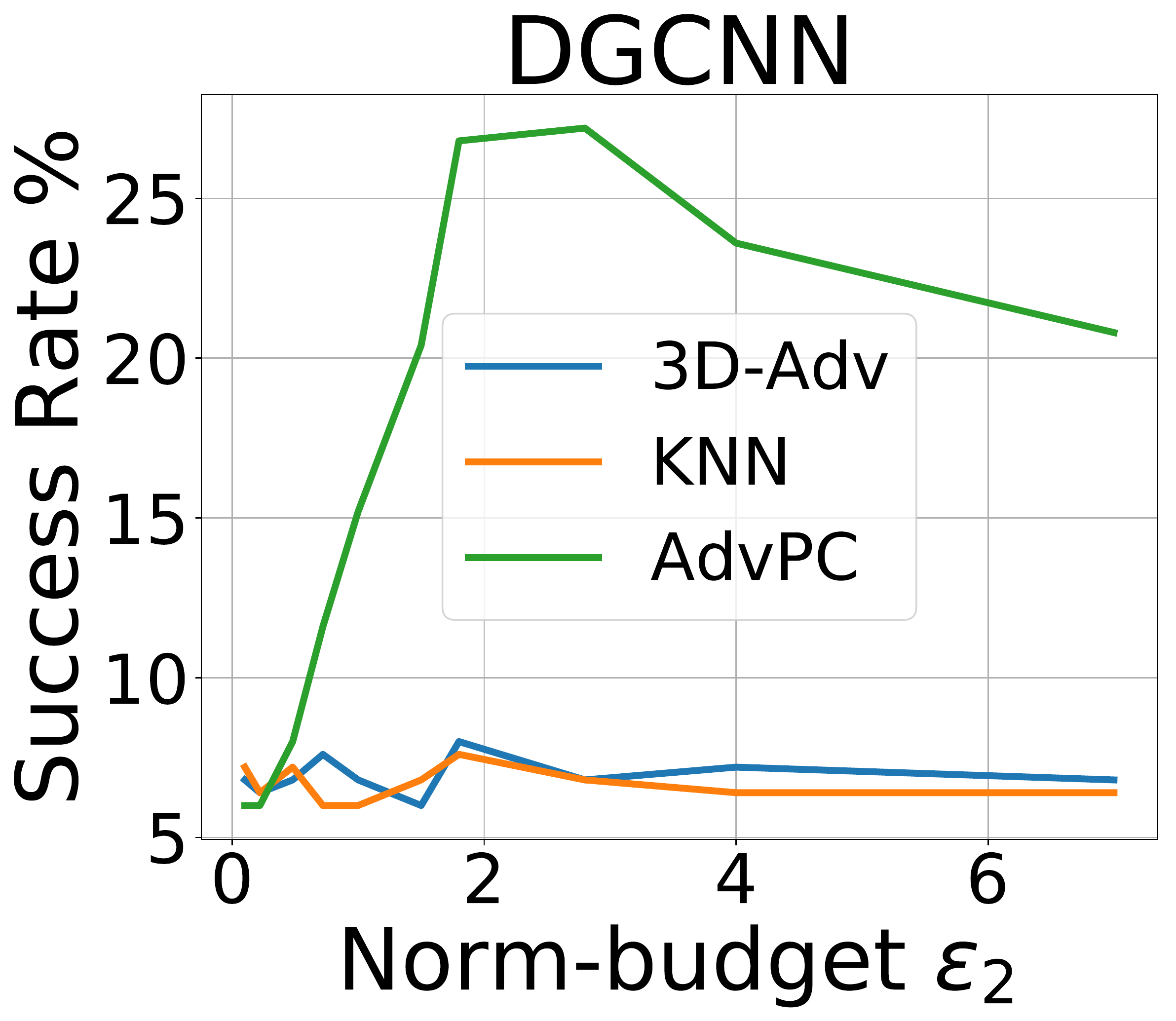} \\
\hline 
\includegraphics[width=0.24\linewidth]{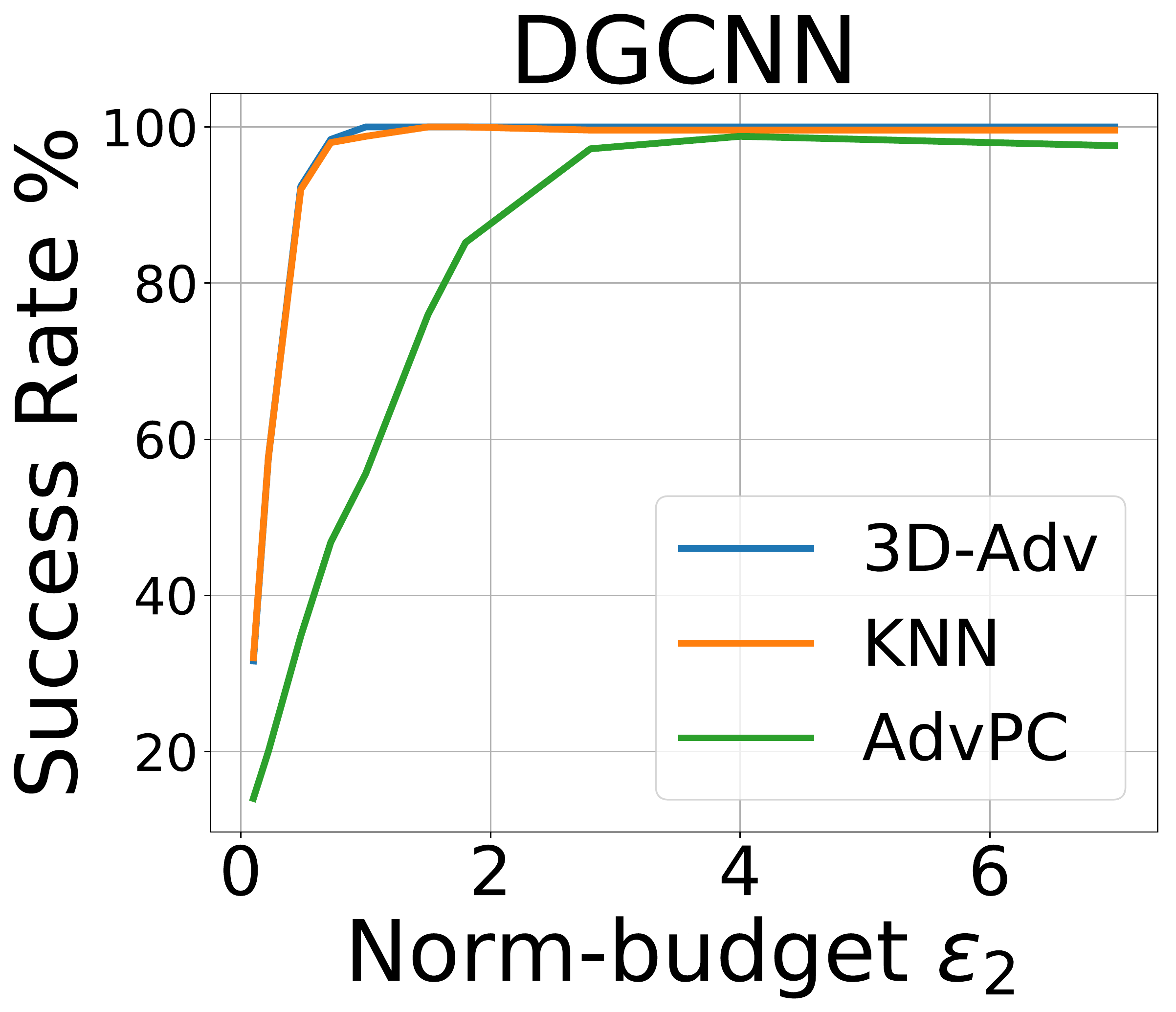} &
 \includegraphics[width=0.25\linewidth]{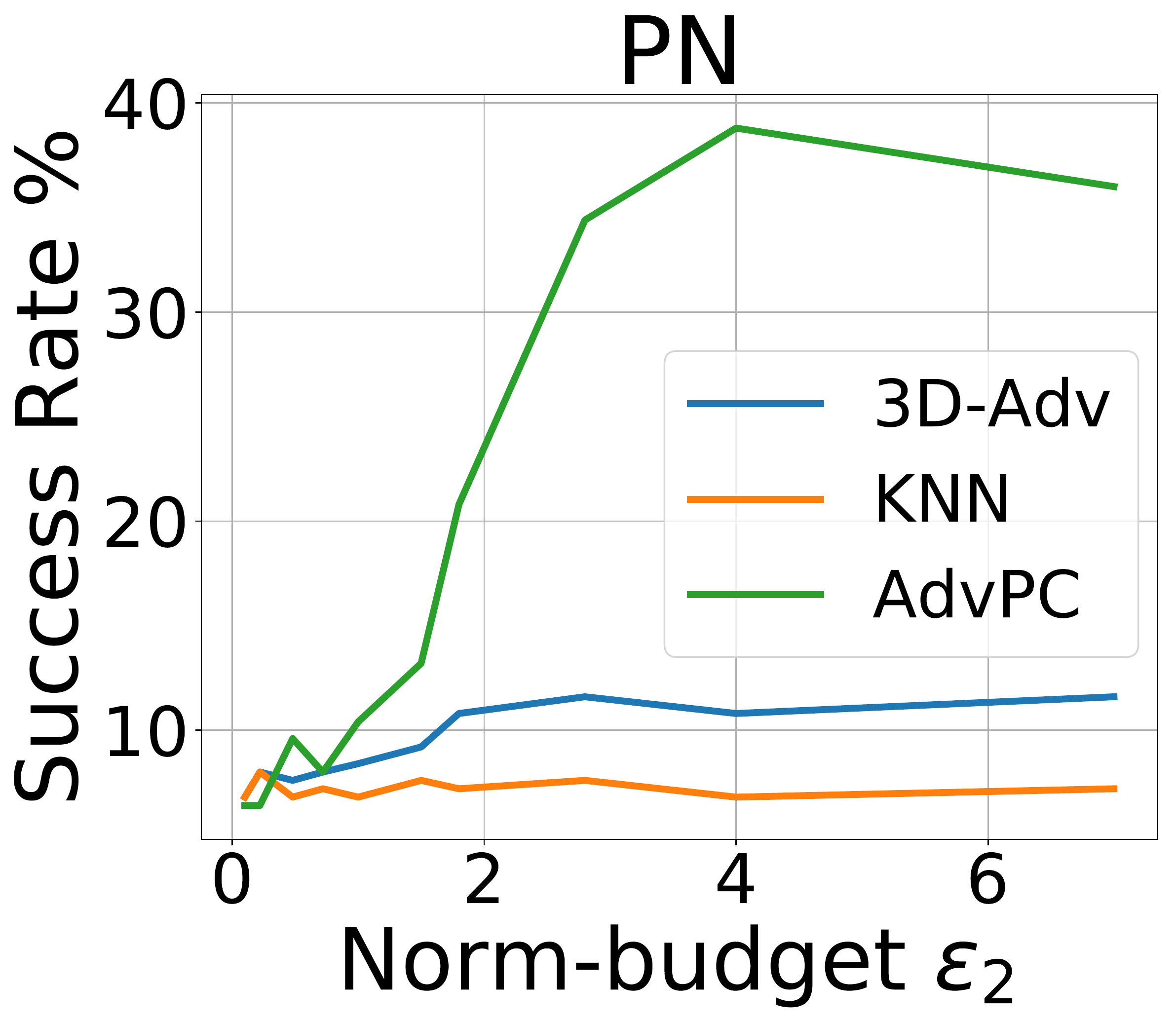} &
\includegraphics[width=0.25\linewidth]{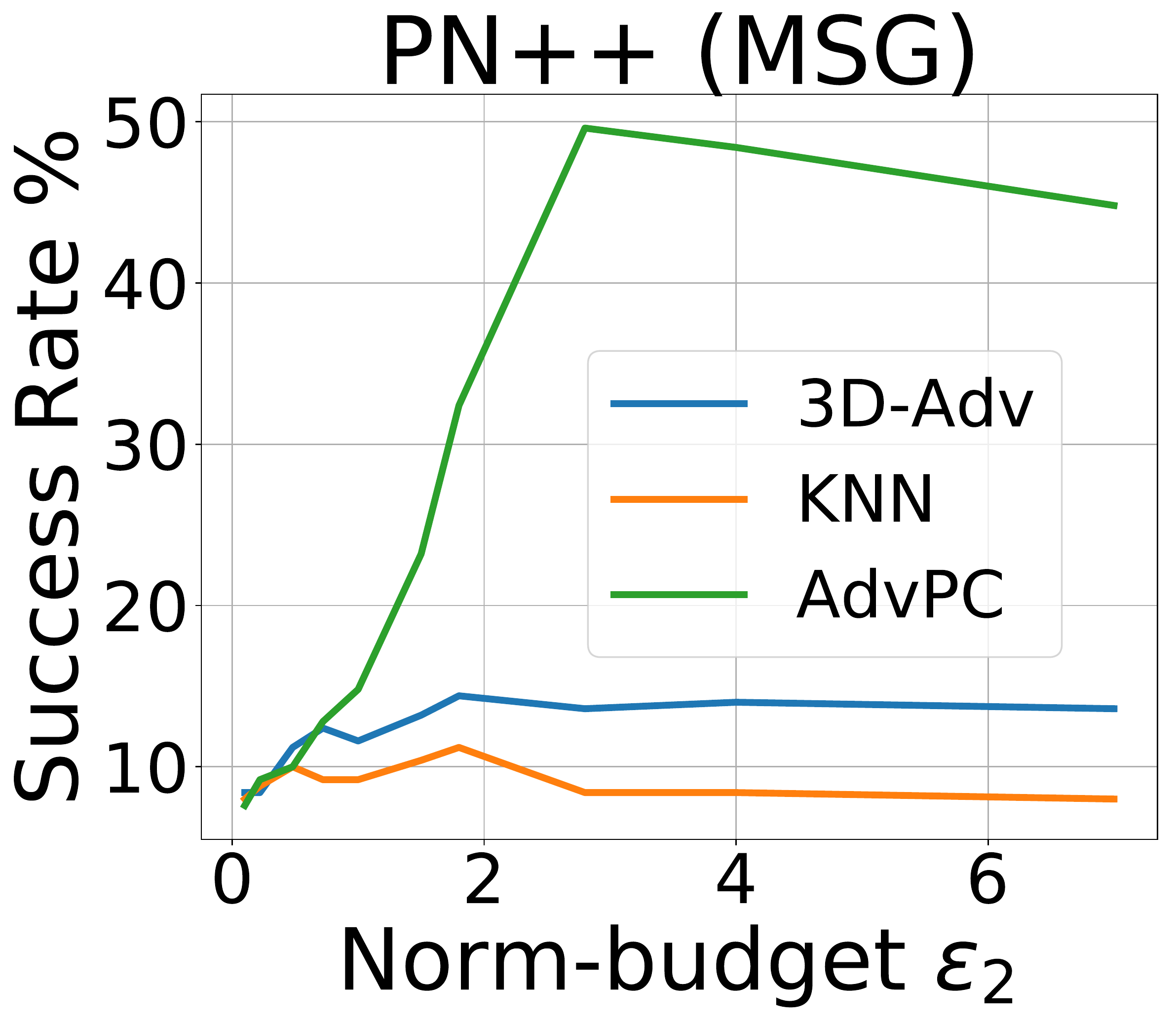} &
\includegraphics[width=0.25\linewidth]{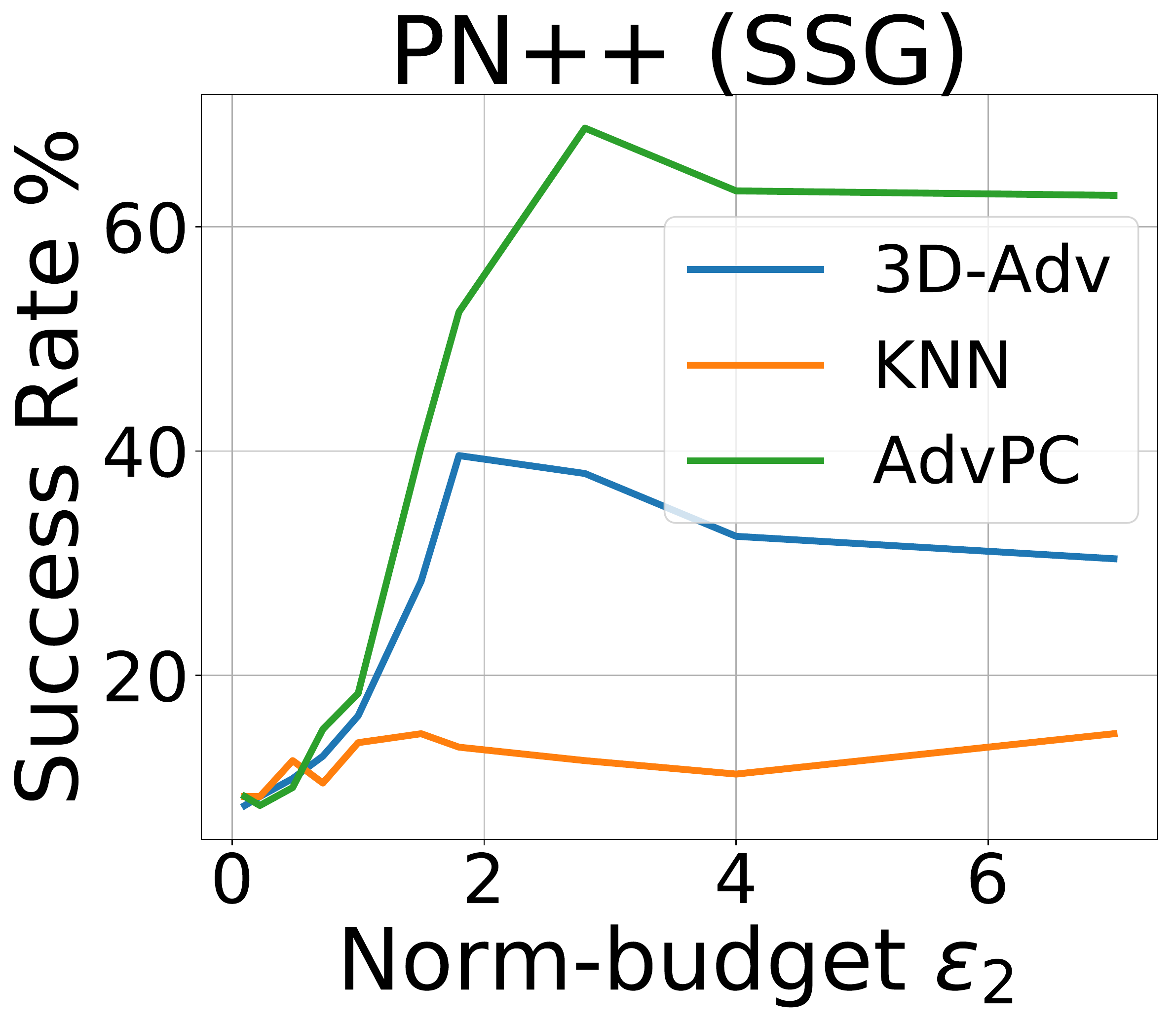} \\
\bottomrule

\end{tabular}

\caption{\small \textbf{Transferability Across Different $\epsilon_2$ Norm-Budgets}: Here, the attacks are optimized using different \textbf{$\epsilon_2$} norm-budgets. We report the attack success on all victim networks and the success of these attacks on each transfer network. We note that our AdvPC transfers better to the other networks across different  $\epsilon_2$ as compared to the baselines 3D-adv\cite{pcattack} and KNN attack \cite{robustshapeattack}.}
\label{fig:sup-transferbility-I}
\vspace{-10pt}
\end{figure}

\clearpage

\subsection{Transferability Matrices} \label{sec:sup-transfer-mat}

\begin{figure}[]
\tabcolsep=0.03cm
\begin{tabular}{ccc}

\includegraphics[width=0.33\linewidth]{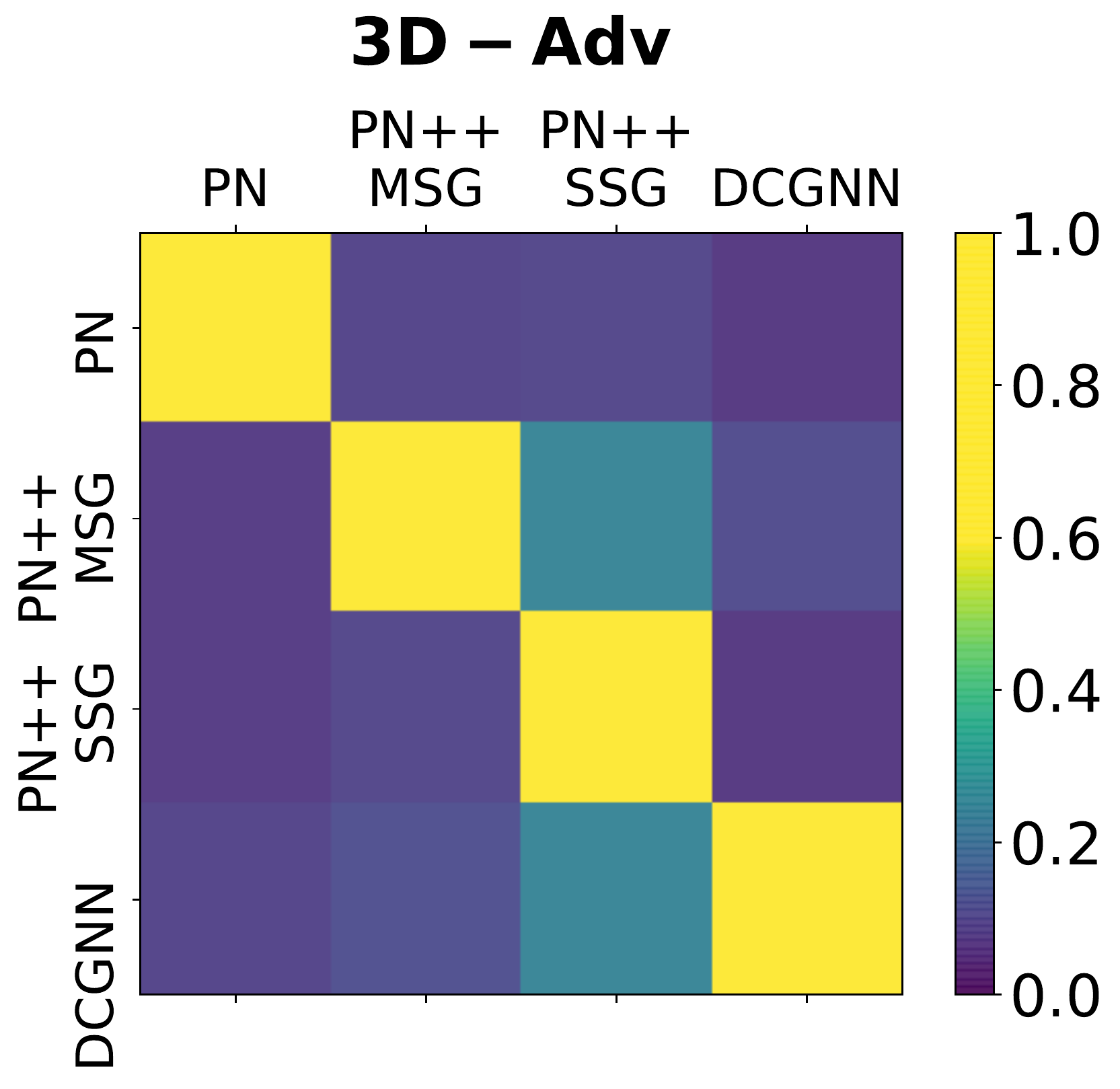} &
\includegraphics[width=0.33\linewidth]{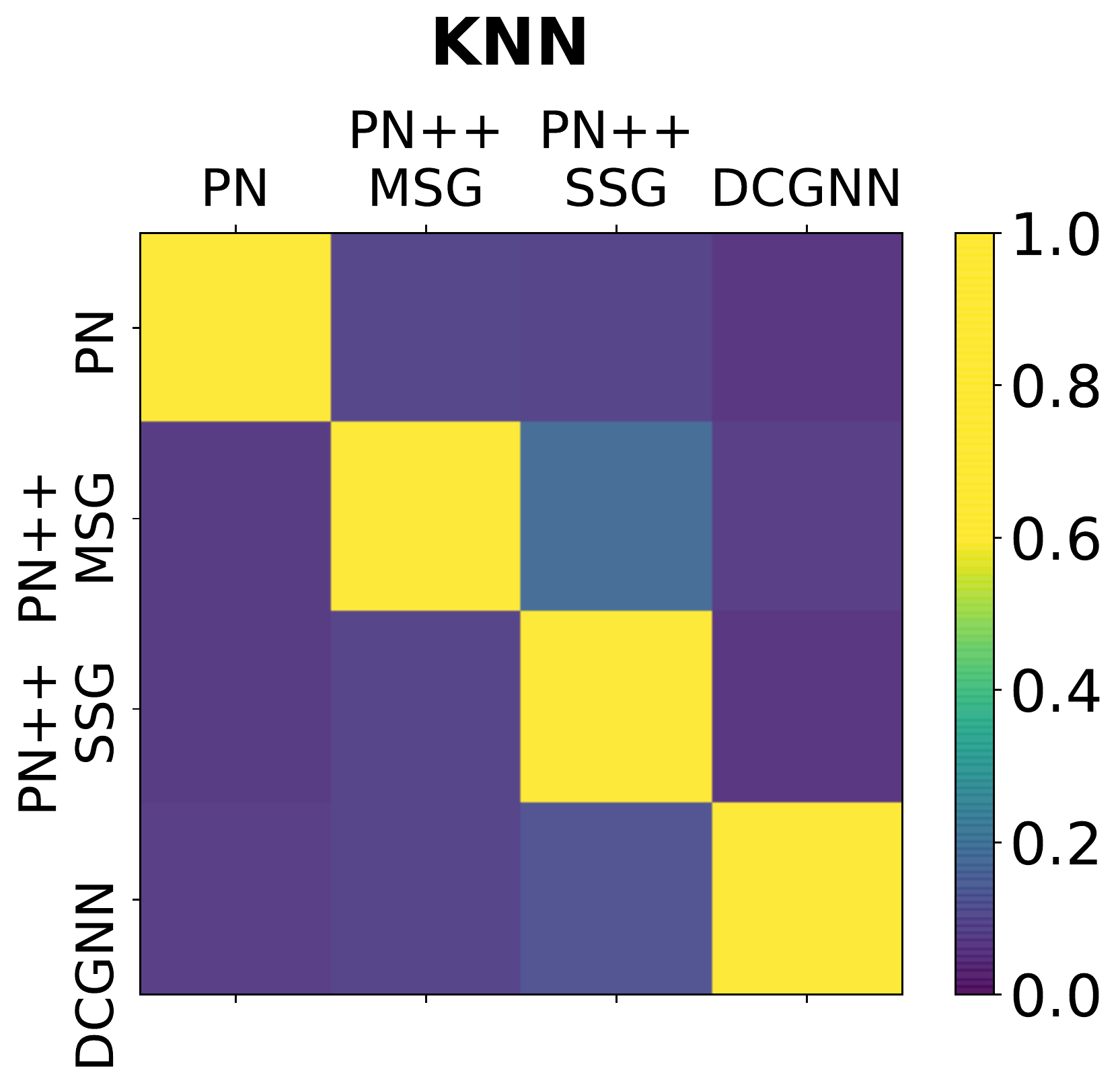} &
\includegraphics[width=0.33\linewidth]{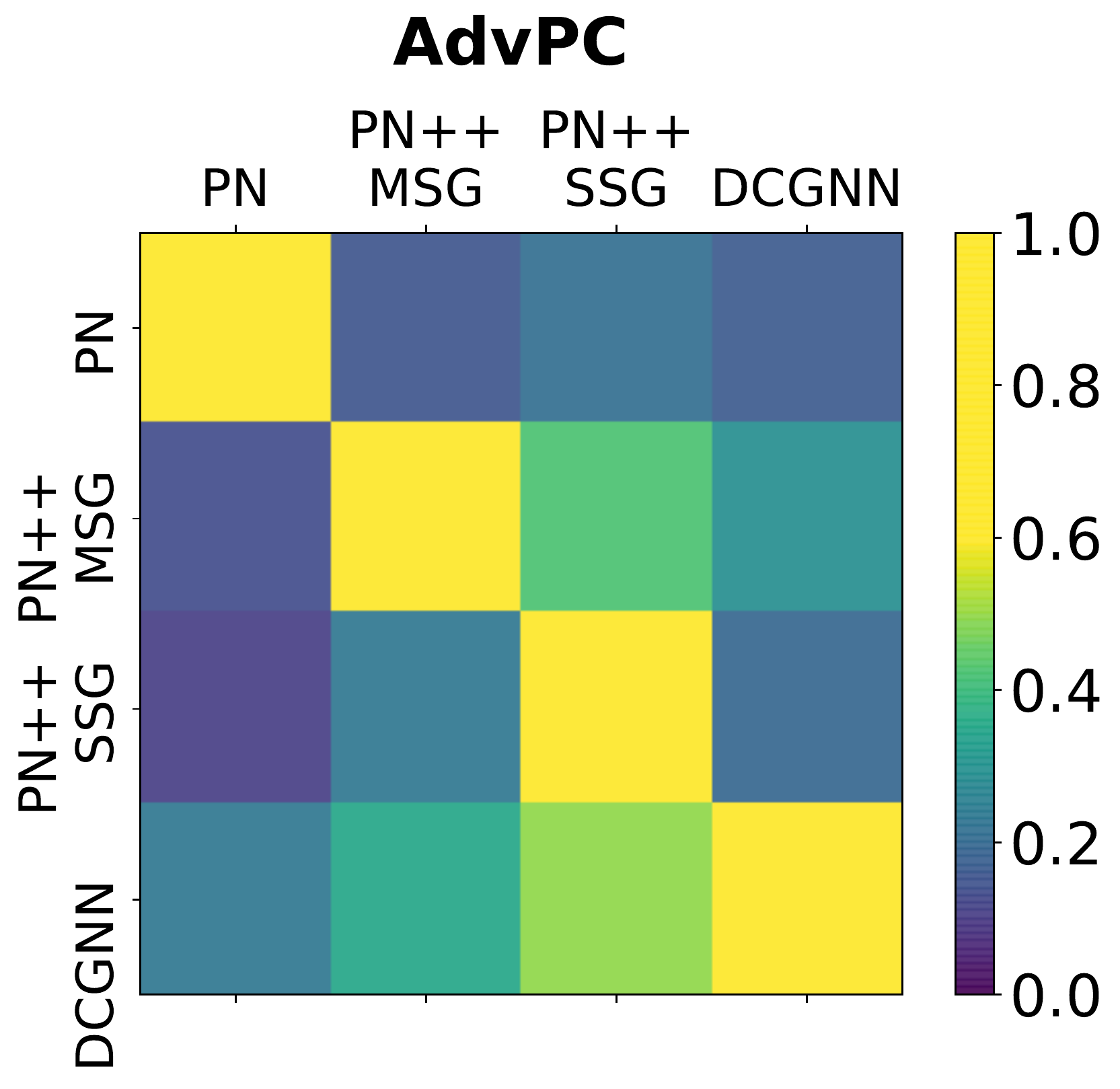}  \\
transferability: 11.5 \% &
transferability: 8.92 \% &
transferability: \textbf{24.9} \% \\
\end{tabular}

\caption{\small \textbf{Transferability Matrix for $\ell_\infty$}: Visualizing the overall transferability for 3D-adv \cite{pcattack} (\textit{left}), KNN attack \cite{robustshapeattack}(\textit{middle}), and our AdvPC (\textit{right}). Elements in the same row correspond to the same victim network used in the attack, while those in the same column correspond to the network that the attack is transferred to. Each matrix element measures the average success rate over the range of $\epsilon_\infty$ for the transfer network. We expect the diagonal elements of each transferability matrix (average success rate on the victim network) to have high values, since each attack is optimized on the same network it is transferred to. More importantly, brighter off-diagonal matrix elements indicate better transferability. 
We observe that our proposed AdvPC attack is more transferable than the other attacks and that DGCNN is a more transferable victim network than the other point cloud networks. 
The transferability score under each matrix is the average of the off-diagonal matrix values, which scores overall transferability for an attack.
}
\label{fig:sup-transmatrix_I}
\end{figure}

\begin{figure}[]
\tabcolsep=0.03cm
\begin{tabular}{ccc}

\includegraphics[width=0.33\linewidth]{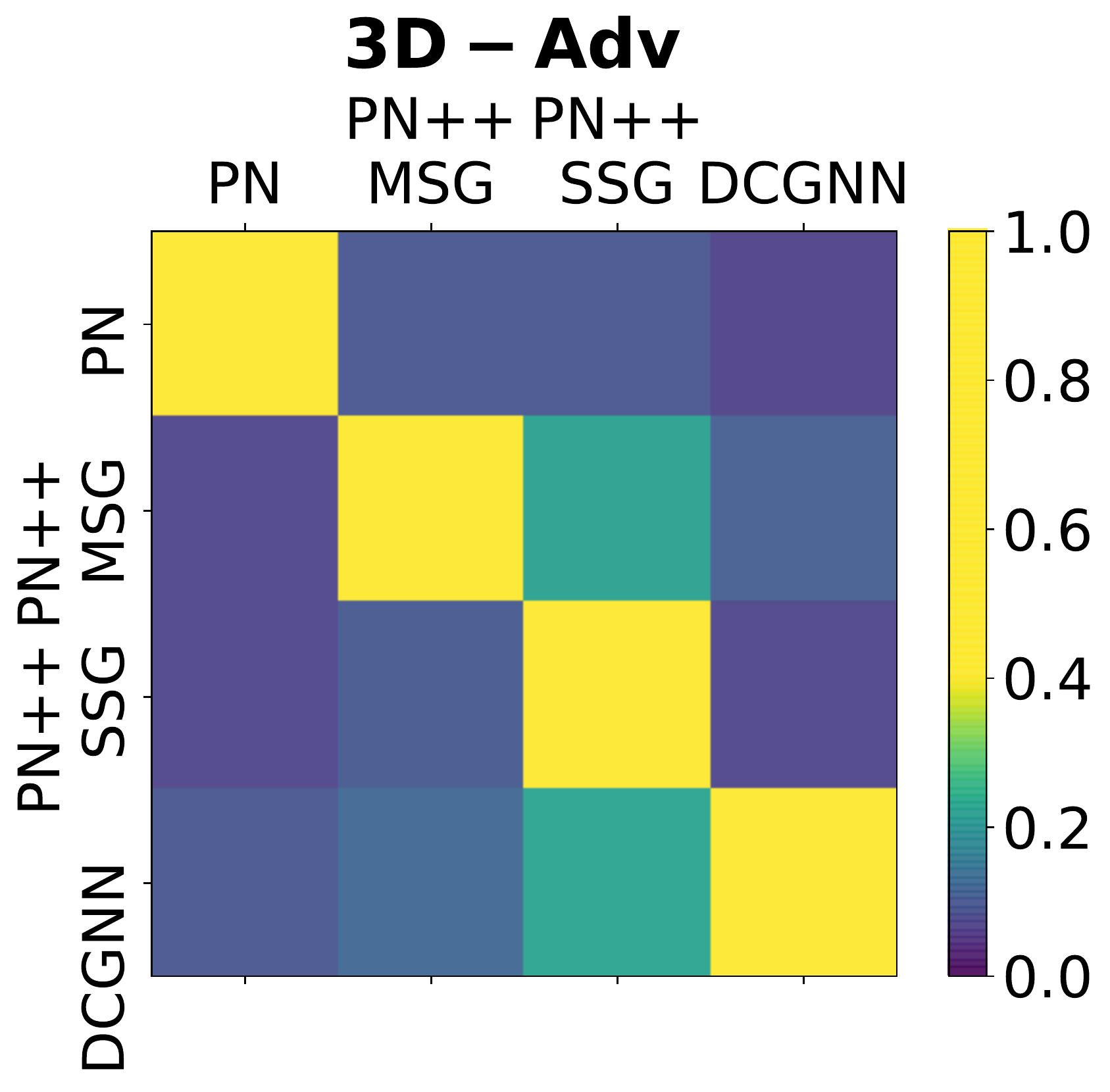} &
\includegraphics[width=0.33\linewidth]{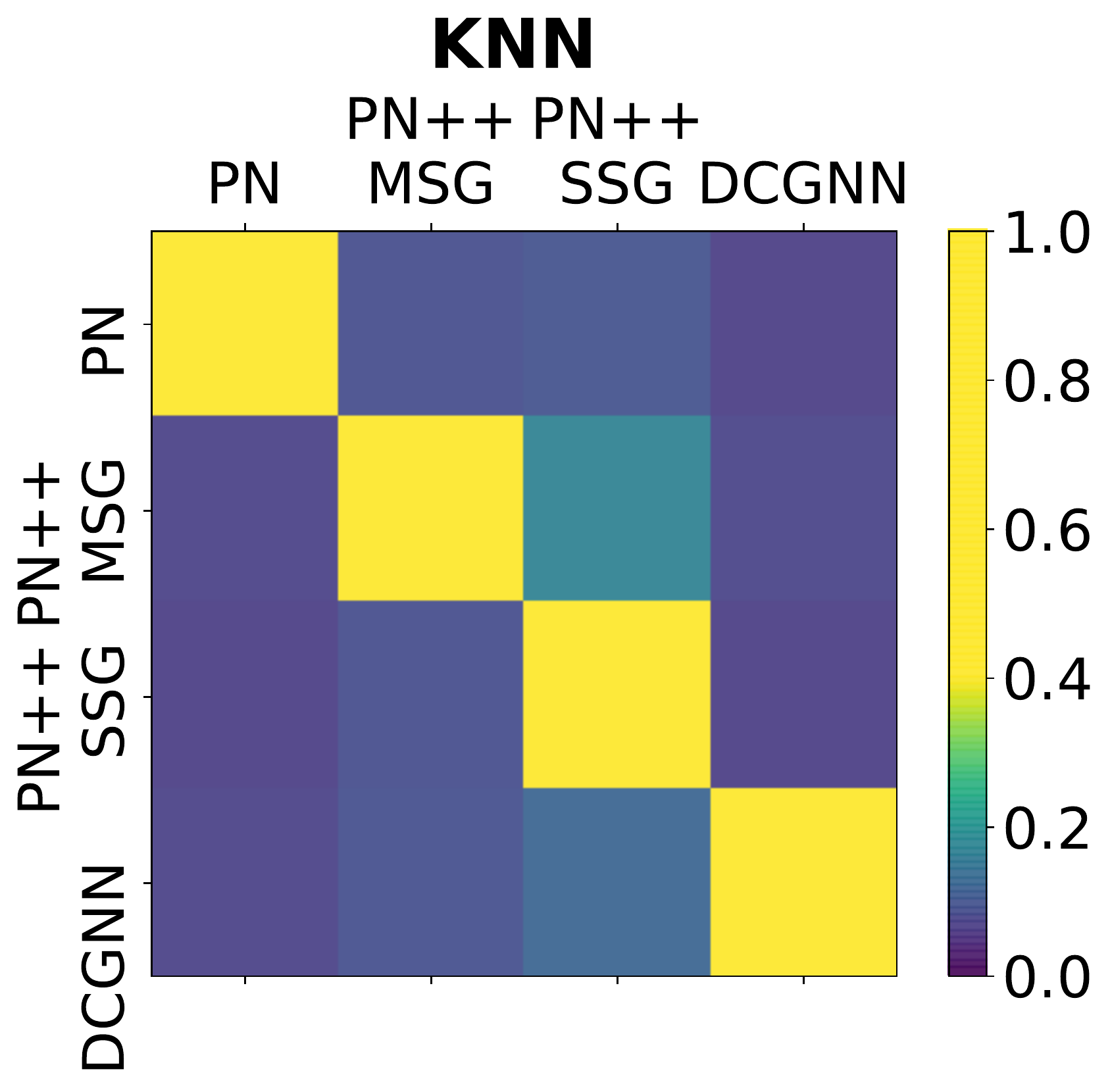} &
\includegraphics[width=0.33\linewidth]{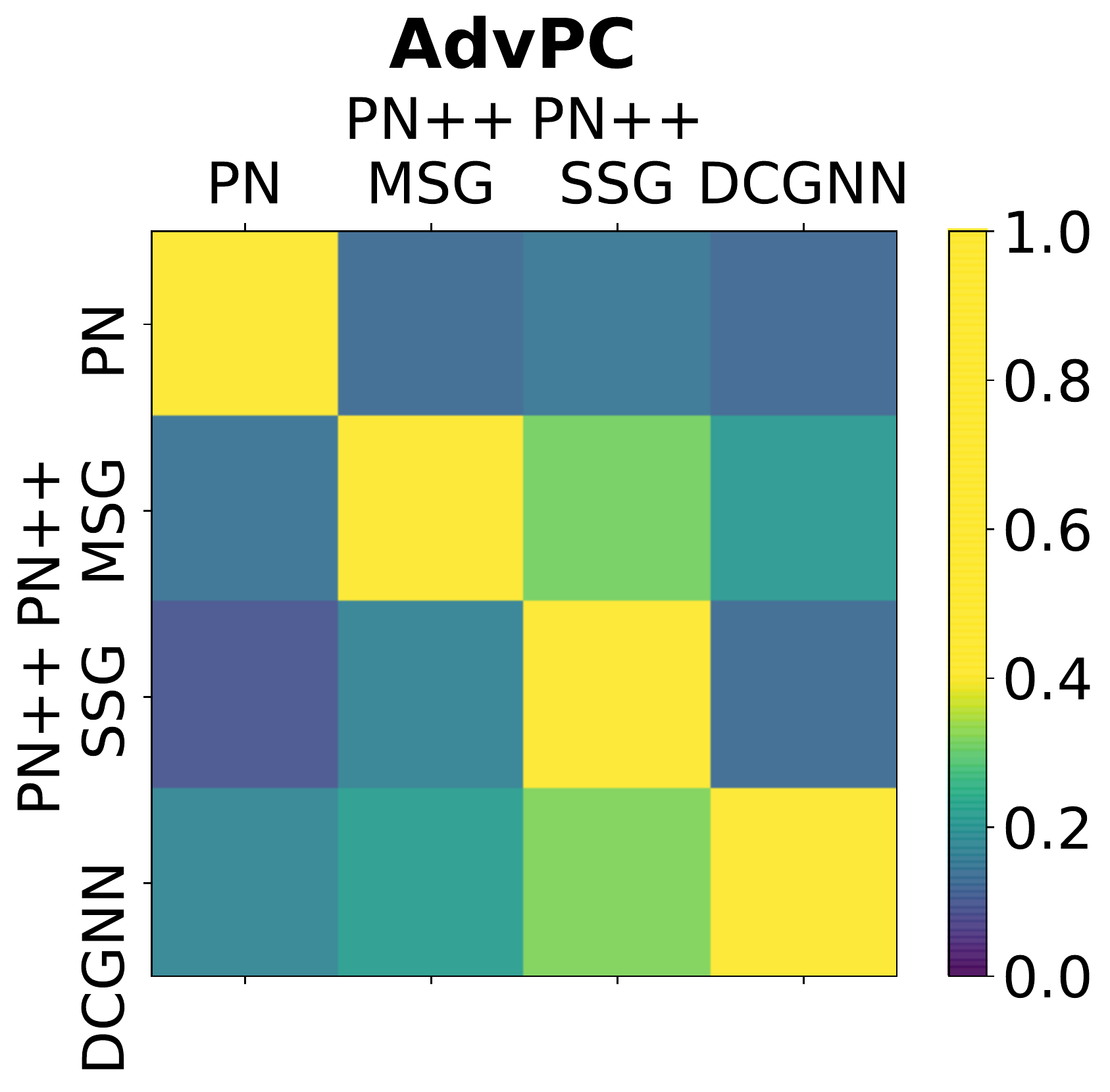}  \\
transferability: 11.0 \% &
transferability: 8.89 \% &
transferability: \textbf{17.9} \% \\
\end{tabular}

\caption{\small \textbf{Transferability Matrix for $\ell_2$}: Visualizing the overall transferability for 3D-adv \cite{pcattack} (\textit{left}), KNN attack \cite{robustshapeattack}(\textit{middle}), and our AdvPC (\textit{right}). Elements in the same row correspond to the same victim network used in the attack, while those in the same column correspond to the network that the attack is transferred to. Each matrix element measures the average success rate over the range of $\epsilon_2$ for the transfer network. We expect the diagonal elements of each transferability matrix (average success rate on the victim network) to have high values, since each attack is optimized on the same network it is transferred to. More importantly, brighter off-diagonal matrix elements indicate better transferability. 
We observe that our proposed AdvPC attack is more transferable than the other attacks and that DGCNN is a more transferable victim network than the other point cloud networks. 
The transferability score under each matrix is the average of the off-diagonal matrix values, which scores overall transferability for an attack.
}
\label{fig:sup-transmatrix_2}
\end{figure}

\clearpage
\section{Defenses Results (Untargeted Attacks)} \label{sec:sup-def-unt}

\subsection{$\ell_\infty$ Defense Results}

\begin{table}[]
\footnotesize
\centering
\vspace{-8pt}
\setlength{\tabcolsep}{4pt} %
\renewcommand{\arraystretch}{1} %
\begin{tabular}{c|ccc|ccc} 
\toprule
 & \multicolumn{3}{c}{$\epsilon_\infty = 0.18$} & \multicolumn{3}{c}{$\epsilon_\infty = 0.45$} \\
\textbf{Defenses} & \textbf{\specialcell{3D-Adv\\ \cite{pcattack} }} & \textbf{\specialcell{KNN \\ \cite{robustshapeattack}}} & \textbf{\specialcell{AdvPC \\(ours)}}  & \textbf{\specialcell{3D-Adv \\\cite{pcattack}}} & \textbf{\specialcell{KNN \\ \cite{robustshapeattack}}} &  \textbf{\specialcell{AdvPC \\(ours)}}    \\
\midrule
No defense & \textbf{100} & 99.6 & 94.8  & \textbf{100} & 99.6 &  97.2\\ 
AE (newly trained) & 9.2 & 10.0 & \textbf{17.2} & 12.0 & 10.0 & \textbf{21.2} \\
Adv Training \cite{pcattack} & 7.2 &  7.6 & \textbf{39.6} &  8.8 &  7.2 & \textbf{42.4} \\
SOR \cite{Deflecting} &18.8 & 17.2 & \textbf{36.8} & 19.2 & 19.2 & \textbf{32.0} \\
DUP Net \cite{Deflecting}  & 28 & 28.8 &\textbf{43.6} & 28 & 31.2 & \textbf{37.2}\\ 
SRS \cite{Deflecting} & 43.2 & 29.2 & \textbf{80.0} & 47.6 & 31.2 & \textbf{85.6} \\
 \bottomrule
\end{tabular}
\caption{\small \textbf{Attacking Point Cloud Defenses ($\ell_\infty$ Untargeted DGCNN):} We evaluate untargeted attacks using norm-budgets of $\epsilon_\infty = 0.18$ and $\epsilon_\infty = 0.45$ with DGCNN \cite{dgcn} as the victim network under different defenses for 3D point clouds. Similar to before, we report attack success rates (\textbf{higher} indicates better attack). AdvPC consistently outperforms the other attacks \cite{pcattack,robustshapeattack} for all defenses. %
Note that both the attacks \textit{and} evaluations are performed on DGCNN, which has an accuracy of 93.7\% without input perturbations (for reference).
}
\label{tbl:breaking-dgcn-I}
\vspace{-6pt}
\end{table}

\begin{table}[]
\footnotesize
\centering
\vspace{-8pt}
\setlength{\tabcolsep}{4pt} %
\renewcommand{\arraystretch}{1} %
\begin{tabular}{c|ccc|ccc} 
\toprule
 & \multicolumn{3}{c}{$\epsilon_\infty = 0.18$} & \multicolumn{3}{c}{$\epsilon_\infty = 0.45$} \\
\textbf{Defenses} & \textbf{\specialcell{3D-Adv\\ \cite{pcattack} }} & \textbf{\specialcell{KNN \\ \cite{robustshapeattack}}} & \textbf{\specialcell{AdvPC \\(ours)}}  & \textbf{\specialcell{3D-Adv \\\cite{pcattack}}} & \textbf{\specialcell{KNN \\ \cite{robustshapeattack}}} &  \textbf{\specialcell{AdvPC \\(ours)}}    \\
\midrule
No defense & \textbf{100} & \textbf{100} & 99.2  & \textbf{100} & \textbf{100} &  99.2\\ 
AE (newly trained) & 14.8 & 13.6 & \textbf{17.6} & 12.0 & 13.2 & \textbf{19.6} \\
Adv Training \cite{pcattack} & 12.0 &  7.6 & \textbf{76.4} & 11.2 & 10.8 & \textbf{76.4} \\
SOR \cite{Deflecting} &20.4 & 18.4 & \textbf{51.2} & 18.8 & 16.0 & \textbf{51.2} \\
DUP Net \cite{Deflecting}  & 18.0 & 16.4 & \textbf{33.6} & 16.8 & 18.4 & \textbf{38.8}\\ 
SRS \cite{Deflecting} & 53.2 & 40.8 & \textbf{90.4} & 49.2 & 42.4 & \textbf{89.6} \\
 \bottomrule
\end{tabular}
\caption{\small \textbf{Attacking Point Cloud Defenses ($\ell_\infty$ Untargeted PointNet++ SSG):} We evaluate untargeted attacks using norm-budgets of $\epsilon_\infty = 0.18$ and $\epsilon_\infty = 0.45$ with PointNet++ SSG \cite{pointnet++} as the victim network under different defenses for 3D point clouds. Similar to before, we report attack success rates (\textbf{higher} indicates better attack). AdvPC consistently outperforms the other attacks \cite{pcattack,robustshapeattack} for all defenses. %
Note that both the attacks \textit{and} evaluations are performed on PointNet++ SSG, which has an accuracy of 91.5\% without input perturbations (for reference).
}
\label{tbl:breaking-pn2-I}
\vspace{-6pt}
\end{table}

\begin{table}[]
\footnotesize
\centering
\vspace{-8pt}
\setlength{\tabcolsep}{4pt} %
\renewcommand{\arraystretch}{1} %
\begin{tabular}{c|ccc|ccc} 
\toprule
 & \multicolumn{3}{c}{$\epsilon_\infty = 0.18$} & \multicolumn{3}{c}{$\epsilon_\infty = 0.45$} \\
\textbf{Defenses} & \textbf{\specialcell{3D-Adv\\ \cite{pcattack} }} & \textbf{\specialcell{KNN \\ \cite{robustshapeattack}}} & \textbf{\specialcell{AdvPC \\(ours)}}  & \textbf{\specialcell{3D-Adv \\\cite{pcattack}}} & \textbf{\specialcell{KNN \\ \cite{robustshapeattack}}} &  \textbf{\specialcell{AdvPC \\(ours)}}    \\
\midrule
No defense & \textbf{100} & \textbf{100} & 97.2  & \textbf{100} & \textbf{100} &  98.0\\ 
AE (newly trained) & 13.2 & 10.0 & \textbf{20.0} & 12.4 & 12.0 & \textbf{18.4} \\
Adv Training \cite{pcattack} & 6.8 & 26.0 & \textbf{36.4} &  8.0 & 31.2 & \textbf{32.8} \\
SOR \cite{Deflecting} &21.6 & 26.0 & \textbf{53.2} & 24.4 & 34.0 & \textbf{42.4} \\
DUP Net \cite{Deflecting}  & 29.6 & 27.6 & \textbf{43.2} & 24.8 & 30.8 & \textbf{42.0}\\ 
SRS \cite{Deflecting} & 43.6 & 45.6 & \textbf{80.4} & 41.2 & 50.0 & \textbf{78.8} \\
 \bottomrule
\end{tabular}
\caption{\small \textbf{Attacking Point Cloud Defenses ($\ell_\infty$ Untargeted PointNet++ MSG):} We evaluate untargeted attacks using norm-budgets of $\epsilon_\infty = 0.18$ and $\epsilon_\infty = 0.45$ with PointNet++ MSG \cite{pointnet++} as the victim network under different defenses for 3D point clouds. Similar to before, we report attack success rates (\textbf{higher} indicates better attack). AdvPC consistently outperforms the other attacks \cite{pcattack,robustshapeattack} for all defenses. %
Note that both the attacks \textit{and} evaluations are performed on PointNet++ MSG, which has an accuracy of 91.5\% without input perturbations (for reference).
}
\label{tbl:breaking-pn1-I}
\vspace{-6pt}
\end{table}

\begin{table}[]
\footnotesize
\centering
\vspace{-8pt}
\setlength{\tabcolsep}{4pt} %
\renewcommand{\arraystretch}{1} %
\begin{tabular}{c|ccc|ccc} 
\toprule
 & \multicolumn{3}{c}{$\epsilon_\infty = 0.18$} & \multicolumn{3}{c}{$\epsilon_\infty = 0.45$} \\
\textbf{Defenses} & \textbf{\specialcell{3D-Adv\\ \cite{pcattack} }} & \textbf{\specialcell{KNN \\ \cite{robustshapeattack}}} & \textbf{\specialcell{AdvPC \\(ours)}}  & \textbf{\specialcell{3D-Adv \\\cite{pcattack}}} & \textbf{\specialcell{KNN \\ \cite{robustshapeattack}}} &  \textbf{\specialcell{AdvPC \\(ours)}}    \\
\midrule
No defense & \textbf{100} & \textbf{100} & 98.8  & \textbf{100} & \textbf{100} &  98.8\\ 
AE (newly trained) & 8.0 &  7.6 & \textbf{11.6} &  8.0 &  7.6 & \textbf{12.4} \\
Adv Training \cite{pcattack} &  8.0 &  8.4 & \textbf{41.6} &  8.4 &  9.2 & \textbf{44.8} \\
SOR \cite{Deflecting} &16.0 & 15.6 & \textbf{29.2} & 16.8 & 15.2 & \textbf{28.4} \\
DUP Net \cite{Deflecting}  & 10.0 & 10.4 & \textbf{12.4} & 11.2 & 8.4 & \textbf{11.2}\\ 
SRS \cite{Deflecting} & 80.8 & 81.6 & \textbf{97.6} & 85.6 & 77.6 & \textbf{97.2} \\
 \bottomrule
\end{tabular}
\caption{\small \textbf{Attacking Point Cloud Defenses ($\ell_\infty$ Untargeted PointNet):} We evaluate untargeted attacks using norm-budgets of $\epsilon_\infty = 0.18$ and $\epsilon_\infty = 0.45$ with PointNet \cite{pointnet} as the victim network under different defenses for 3D point clouds. Similar to before, we report attack success rates (\textbf{higher} indicates better attack). AdvPC consistently outperforms the other attacks \cite{pcattack,robustshapeattack} for all defenses. %
Note that both the attacks \textit{and} evaluations are performed on PointNet, which has an accuracy of 92.8\% without input perturbations (for reference).
}
\label{tbl:breaking-pn-I}
\vspace{-6pt}
\end{table}

\clearpage
\subsection{$\ell_2$ Defense Results}

\begin{table}[]
\footnotesize
\centering
\vspace{-8pt}
\setlength{\tabcolsep}{4pt} %
\renewcommand{\arraystretch}{1} %
\begin{tabular}{c|ccc|ccc} 
\toprule
 & \multicolumn{3}{c}{$\epsilon_2 = 1.8$} & \multicolumn{3}{c}{$\epsilon_2 = 4.0$} \\
\textbf{Defenses} & \textbf{\specialcell{3D-Adv\\ \cite{pcattack} }} & \textbf{\specialcell{KNN \\ \cite{robustshapeattack}}} & \textbf{\specialcell{AdvPC \\(ours)}}  & \textbf{\specialcell{3D-Adv \\\cite{pcattack}}} & \textbf{\specialcell{KNN \\ \cite{robustshapeattack}}} &  \textbf{\specialcell{AdvPC \\(ours)}}    \\
\midrule
No defense & \textbf{100} & \textbf{100} & 85.2  & \textbf{100} & 99.6 &  98.8\\ 
AE (newly trained) &  9.6 &  9.6 & \textbf{11.6} & 10.8 & 10.0 & \textbf{21.6} \\
Adv Training \cite{pcattack} & 16.8 & 37.6 & \textbf{48.0} &  8.0 & 13.2 & \textbf{40.8} \\
SOR \cite{Deflecting} &22.0 & 29.2 & \textbf{36.8} & 18.0 & 20.4 & \textbf{27.2} \\
DUP Net \cite{Deflecting}  & 34.8 & 36.0 & \textbf{36.8} & 28.8 & 28.4 & \textbf{31.2}\\ 
SRS \cite{Deflecting} & 63.6 & 61.6 & \textbf{76.0} & 50.8 & 34.0 & \textbf{88.4} \\
 \bottomrule
\end{tabular}
\caption{\small \textbf{Attacking Point Cloud Defenses ($\ell_2$ Untargeted DGCNN):} We evaluate untargeted attacks using norm-budgets of $\epsilon_2 = 1.8$ and $\epsilon_2 = 4.0$ with DGCNN \cite{dgcn} as the victim network under different defenses for 3D point clouds. Similar to before, we report attack success rates (\textbf{higher} indicates better attack). AdvPC consistently outperforms the other attacks \cite{pcattack,robustshapeattack} for all defenses. %
Note that both the attacks \textit{and} evaluations are performed on DGCNN, which has an accuracy of 93.7\% without input perturbations (for reference).
}
\label{tbl:breaking-dgcn-2}
\vspace{-6pt}
\end{table}

\begin{table}[]
\footnotesize
\centering
\vspace{-8pt}
\setlength{\tabcolsep}{4pt} %
\renewcommand{\arraystretch}{1} %
\begin{tabular}{c|ccc|ccc} 
\toprule
 & \multicolumn{3}{c}{$\epsilon_2 = 1.8$} & \multicolumn{3}{c}{$\epsilon_2 = 4.0$} \\
\textbf{Defenses} & \textbf{\specialcell{3D-Adv\\ \cite{pcattack} }} & \textbf{\specialcell{KNN \\ \cite{robustshapeattack}}} & \textbf{\specialcell{AdvPC \\(ours)}}  & \textbf{\specialcell{3D-Adv \\\cite{pcattack}}} & \textbf{\specialcell{KNN \\ \cite{robustshapeattack}}} &  \textbf{\specialcell{AdvPC \\(ours)}}    \\
\midrule
No defense & \textbf{100} & \textbf{100} & 96.4  & \textbf{100} & \textbf{100} &  98.8\\ 
AE (newly trained) & 13.2 & 14.0 &\textbf{ 18.0} & 13.6 & 14.0 & \textbf{17.6} \\
Adv Training \cite{pcattack} & 20.8 & 19.2 & \textbf{74.4} & 10.8 & 11.6 & \textbf{71.2} \\
SOR \cite{Deflecting} &24.8 & 17.2 & \textbf{49.6} & 17.6 & 14.4 & \textbf{48.4} \\
DUP Net \cite{Deflecting}  &18.4 & 15.2 &\textbf{33.6} & 18.0 & 16.0 & \textbf{32.8}\\ 
SRS \cite{Deflecting} & 60.4 & 55.2 & \textbf{86.4} & 50.8 & 42.4 & \textbf{89.2} \\
 \bottomrule
\end{tabular}
\caption{\small \textbf{Attacking Point Cloud Defenses ($\ell_2$ Untargeted PointNet++ SSG):} We evaluate untargeted attacks using norm-budgets of $\epsilon_2 = 1.8$ and $\epsilon_2 = 4.0$ with PointNet++ SSG \cite{pointnet++} as the victim network under different defenses for 3D point clouds. Similar to before, we report attack success rates (\textbf{higher} indicates better attack). AdvPC consistently outperforms the other attacks \cite{pcattack,robustshapeattack} for all defenses. %
Note that both the attacks \textit{and} evaluations are performed on PointNet++ SSG, which has an accuracy of 91.5\% without input perturbations (for reference).
}
\label{tbl:breaking-pn2-2}
\vspace{-6pt}
\end{table}

\begin{table}[]
\footnotesize
\centering
\vspace{-8pt}
\setlength{\tabcolsep}{4pt} %
\renewcommand{\arraystretch}{1} %
\begin{tabular}{c|ccc|ccc} 
\toprule
 & \multicolumn{3}{c}{$\epsilon_2 = 1.8$} & \multicolumn{3}{c}{$\epsilon_2 = 4.0$} \\
\textbf{Defenses} & \textbf{\specialcell{3D-Adv\\ \cite{pcattack} }} & \textbf{\specialcell{KNN \\ \cite{robustshapeattack}}} & \textbf{\specialcell{AdvPC \\(ours)}}  & \textbf{\specialcell{3D-Adv \\\cite{pcattack}}} & \textbf{\specialcell{KNN \\ \cite{robustshapeattack}}} &  \textbf{\specialcell{AdvPC \\(ours)}}    \\
\midrule
No defense & \textbf{100} & \textbf{100} & 94.8  & \textbf{100} & \textbf{100} &  98.4\\ 
AE (newly trained) & 13.2 & 11.2 & \textbf{18.4} & 14.8 &  9.6 & \textbf{20.0} \\
Adv Training \cite{pcattack} & 18.8 & 46.0 & \textbf{48.4} &  8.0 & 34.4 & \textbf{36.8} \\
SOR \cite{Deflecting} &32.8 & 37.2 & \textbf{49.2} & 19.2 & 37.2 & \textbf{47.2} \\
DUP Net \cite{Deflecting}  & 31.6 & 33.6 & \textbf{42.8} & 26.8 & 32.8 & \textbf{40.4}\\ 
SRS \cite{Deflecting} & 63.6 & 64.8 & \textbf{83.6} & 44.8 & 49.6 & \textbf{80.0} \\
 \bottomrule
\end{tabular}
\caption{\small \textbf{Attacking Point Cloud Defenses ($\ell_2$ Untargeted PointNet++ MSG):} We evaluate untargeted attacks using norm-budgets of $\epsilon_2 = 1.8$ and $\epsilon_2 = 4.0$ with PointNet++ MSG \cite{pointnet++} as the victim network under different defenses for 3D point clouds. Similar to before, we report attack success rates (\textbf{higher} indicates better attack). AdvPC consistently outperforms the other attacks \cite{pcattack,robustshapeattack} for all defenses. %
Note that both the attacks \textit{and} evaluations are performed on PointNet++ MSG, which has an accuracy of 91.5\% without input perturbations (for reference).
}
\label{tbl:breaking-pn1-2}
\vspace{-6pt}
\end{table}

\begin{table}[]
\footnotesize
\centering
\vspace{-8pt}
\setlength{\tabcolsep}{4pt} %
\renewcommand{\arraystretch}{1} %
\begin{tabular}{c|ccc|ccc} 
\toprule
 & \multicolumn{3}{c}{$\epsilon_2 = 1.8$} & \multicolumn{3}{c}{$\epsilon_2 = 4.0$} \\
\textbf{Defenses} & \textbf{\specialcell{3D-Adv\\ \cite{pcattack} }} & \textbf{\specialcell{KNN \\ \cite{robustshapeattack}}} & \textbf{\specialcell{AdvPC \\(ours)}}  & \textbf{\specialcell{3D-Adv \\\cite{pcattack}}} & \textbf{\specialcell{KNN \\ \cite{robustshapeattack}}} &  \textbf{\specialcell{AdvPC \\(ours)}}    \\
\midrule
No defense & \textbf{100} & \textbf{100} & 98.0  & \textbf{100} & \textbf{100} &  98.8\\ 
AE (newly trained) & 7.6 &  7.6 & \textbf{13.2} &  8.0 &  7.6 & \textbf{12.8} \\
Adv Training \cite{pcattack} &  9.2 & 10.0 & \textbf{43.6} &  8.8 &  8.4 & \textbf{44.0} \\
SOR \cite{Deflecting} &20.0 & 14.4 & \textbf{27.6} & 16.4 & 15.2 & \textbf{25.6} \\
DUP Net \cite{Deflecting}  & 12.0 & 9.2 & \textbf{15.6} & 10.4 & 9.2 & \textbf{11.6}\\ 
SRS \cite{Deflecting} & 88.8 & 84.0 & \textbf{96.4} & 86.8 & 84.4 & \textbf{98.4} \\
 \bottomrule
\end{tabular}
\caption{\small \textbf{Attacking Point Cloud Defenses ($\ell_2$ Untargeted PointNet):} We evaluate untargeted attacks using norm-budgets of $\epsilon_2 = 1.8$ and $\epsilon_2 = 4.0$ with PointNet \cite{pointnet} as the victim network under different defenses for 3D point clouds. Similar to before, we report attack success rates (\textbf{higher} indicates better attack). AdvPC consistently outperforms the other attacks \cite{pcattack,robustshapeattack} for all defenses. %
Note that both the attacks \textit{and} evaluations are performed on PointNet, which has an accuracy of 92.8\% without input perturbations (for reference).
}
\label{tbl:breaking-pn-2}
\vspace{-6pt}
\end{table}

\clearpage

\section{Analysis of the results }
We perform several analytical experiments to explore further the results obtained in so far. We perform several analytical experiments to further explore the results obtained in Sections \ref{sec:sup-transfer-spec},\ref{sec:sup-transfer-all},\ref{sec:sup-transfer-mat} and \ref{sec:sup-def-unt}. We first study the effect of different factors that play a role in the 
transferability of our attacks. We also show some interesting insights related to the sensitivity of point cloud networks and the effect of the AE on the attacks.

\subsection{Ablation Study (hyperparameter $\gamma$)} \label{sec:sup-ablation}
\begin{figure}[]
\tabcolsep=0.03cm
\begin{tabular}{cc}
\includegraphics[width=0.5\columnwidth]{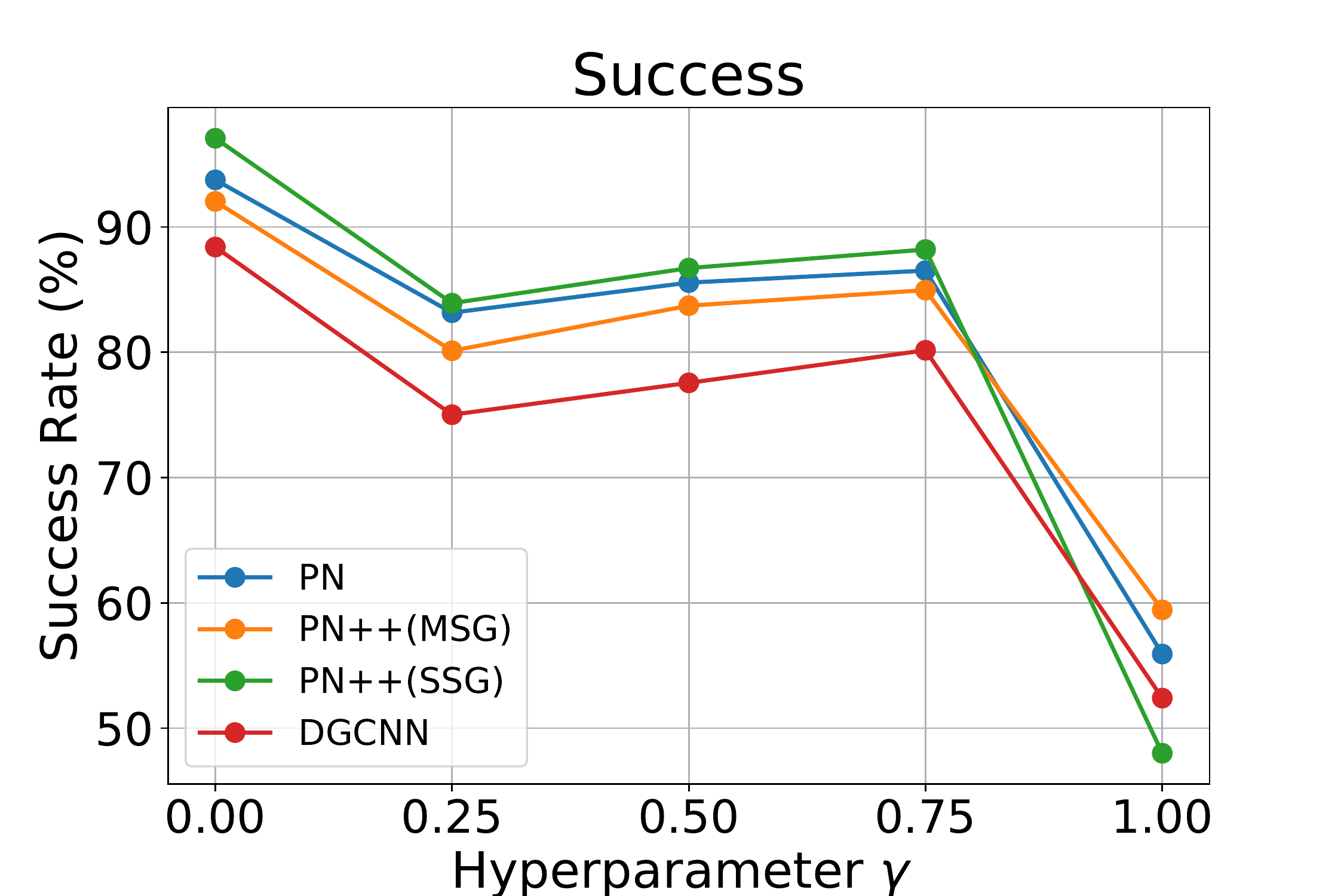} &
\includegraphics[width=0.5\columnwidth]{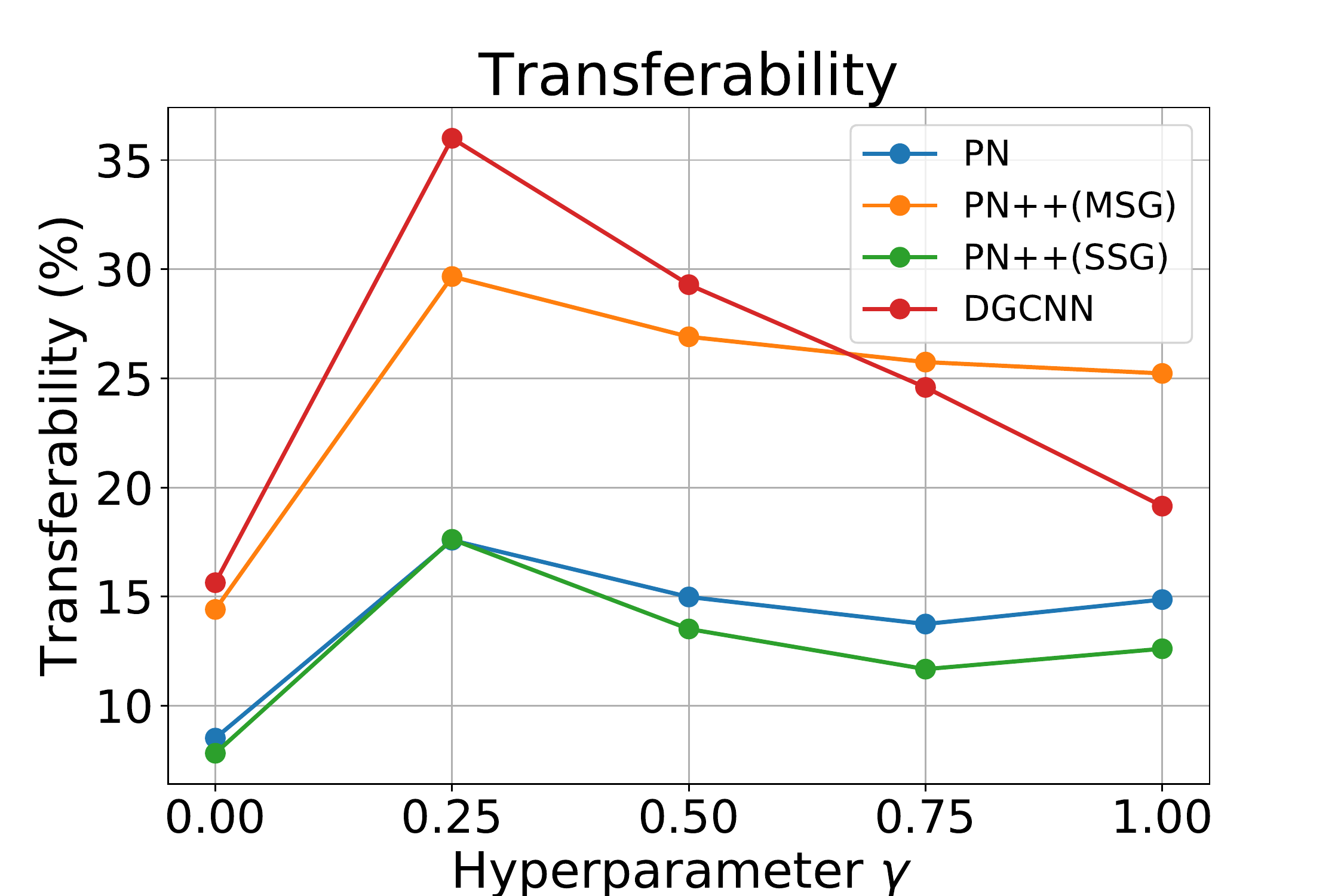} \\
\end{tabular}
\caption{\small \textbf{Ablation Study in $\ell_\infty$}: Studying the effect of changing AdvPC hyperparameter ($\gamma$) on the success rate of the attack (\textit{left}) and on its transferability (\textit{right}). The transferability score reported for each victim network is the average success rate on the transfer networks averaged across all different norm-budgets $\epsilon_\infty$. We note that as $\gamma$ increases, the success rate of the attack on the victim network drops, and the transferability varies with $\gamma$. We pick $\gamma=0.25$ in all of our experiments.}
\label{fig:sup-gamma-I}
  \vspace{-8pt}
\end{figure}

\begin{figure}[]
\tabcolsep=0.03cm
\begin{tabular}{cc}
\includegraphics[width=0.5\columnwidth]{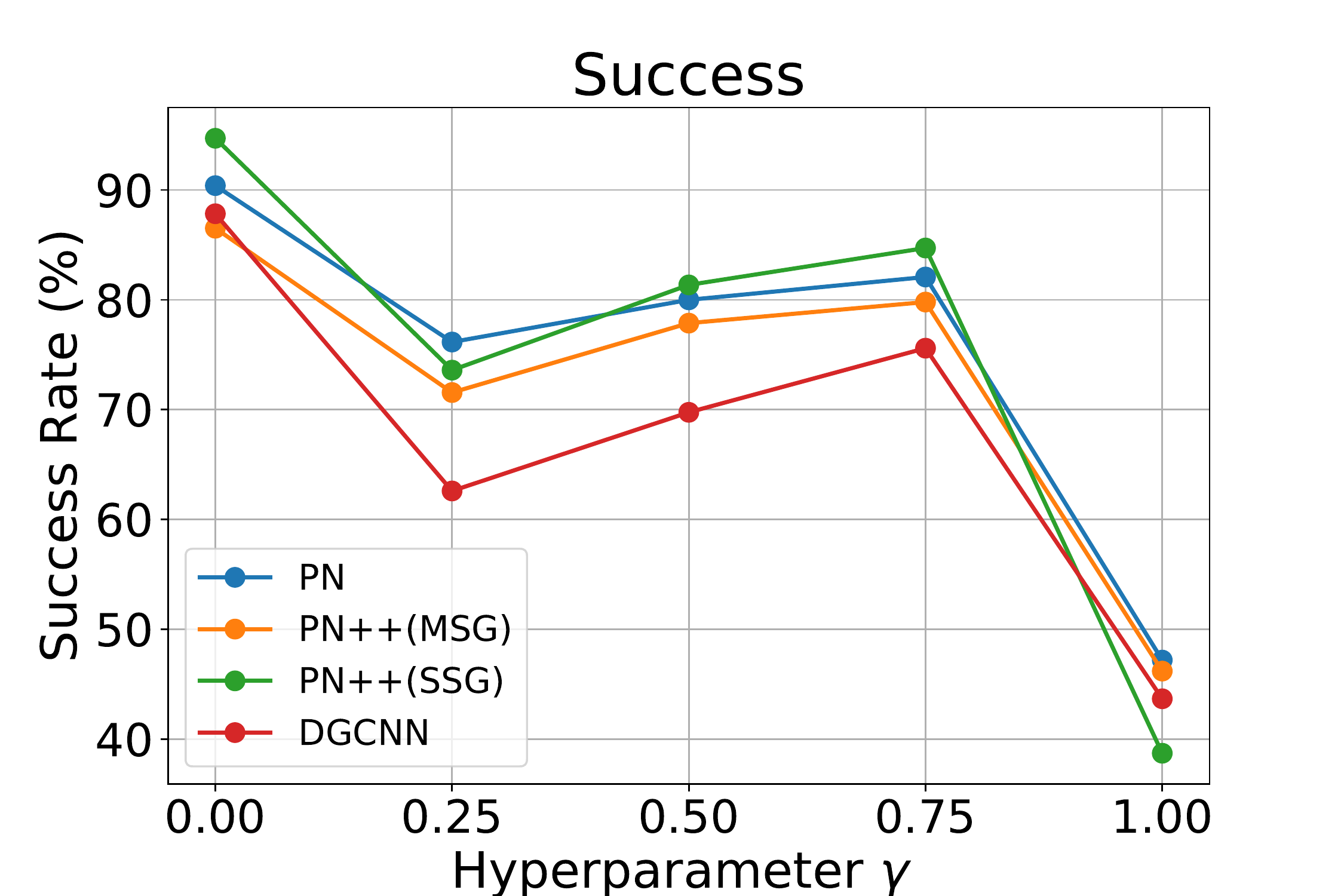} &
\includegraphics[width=0.5\columnwidth]{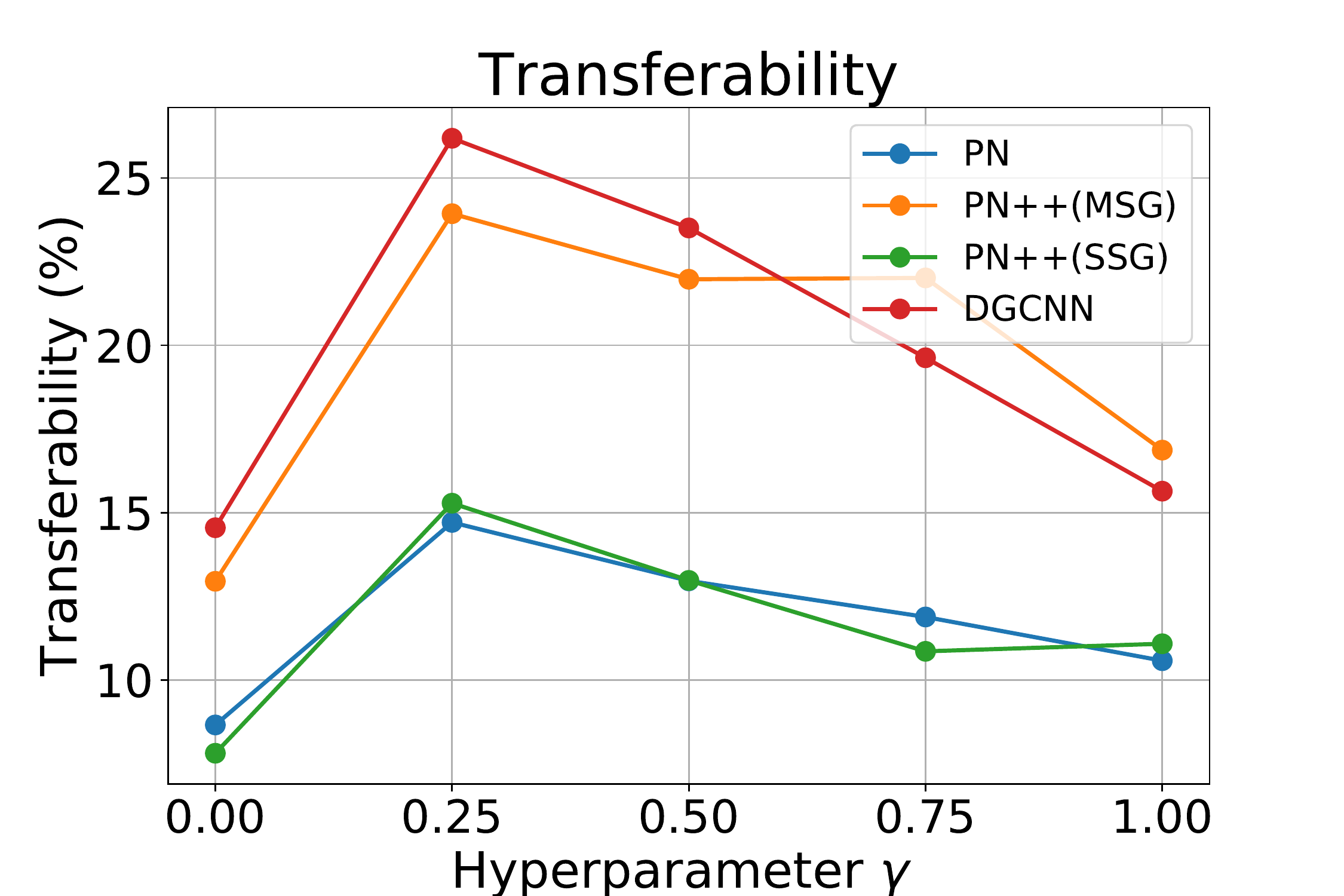} \\
\end{tabular}
\caption{\small \textbf{Ablation Study in $\ell_2$}: Studying the effect of changing AdvPC hyperparameter ($\gamma$) on the success rate of the attack (\textit{left}) and on its transferability (\textit{right}). The transferability score reported for each victim network is the average success rate on the transfer networks averaged across all different norm-budgets $\epsilon_2$. We note that as $\gamma$ increases, the success rate of the attack on the victim network drops, and the transferability varies with $\gamma$. We pick $\gamma=0.25$ in all of our experiments.}
\label{fig:sup-gamma-2}
  \vspace{-8pt}
\end{figure}

Here, we study the effect of $\gamma$ used in \eqLabel{\ref{eq:sup-final-objective}} on the performance of our attacks. 
While varying $\gamma$ between 0 and 1, we record the attack success rate on the victim network and report the transferability to all of the other three transfer networks (average success rate on the transfer networks). We present our results (averaged over all $\epsilon_\infty$ norm-budgets) in \figLabel{\ref{fig:sup-gamma-I}} and in \figLabel{\ref{fig:sup-gamma-2}} (averaged over all $\epsilon_2$ norm-budgets) for the four victim networks. %
One observation is that, while adding the AE loss with $\gamma>0$ indeed improves transferability, it tends to deteriorate the success rate. We pick $\gamma=0.25$ in our experiments to balance success and transferability.   

\subsection{Network Sensitivity to Point Cloud Attacks} \label{sec:sup-sens}
\begin{figure}[]
\tabcolsep=0.03cm
\begin{tabular}{ccc}
\includegraphics[width=0.33\columnwidth]{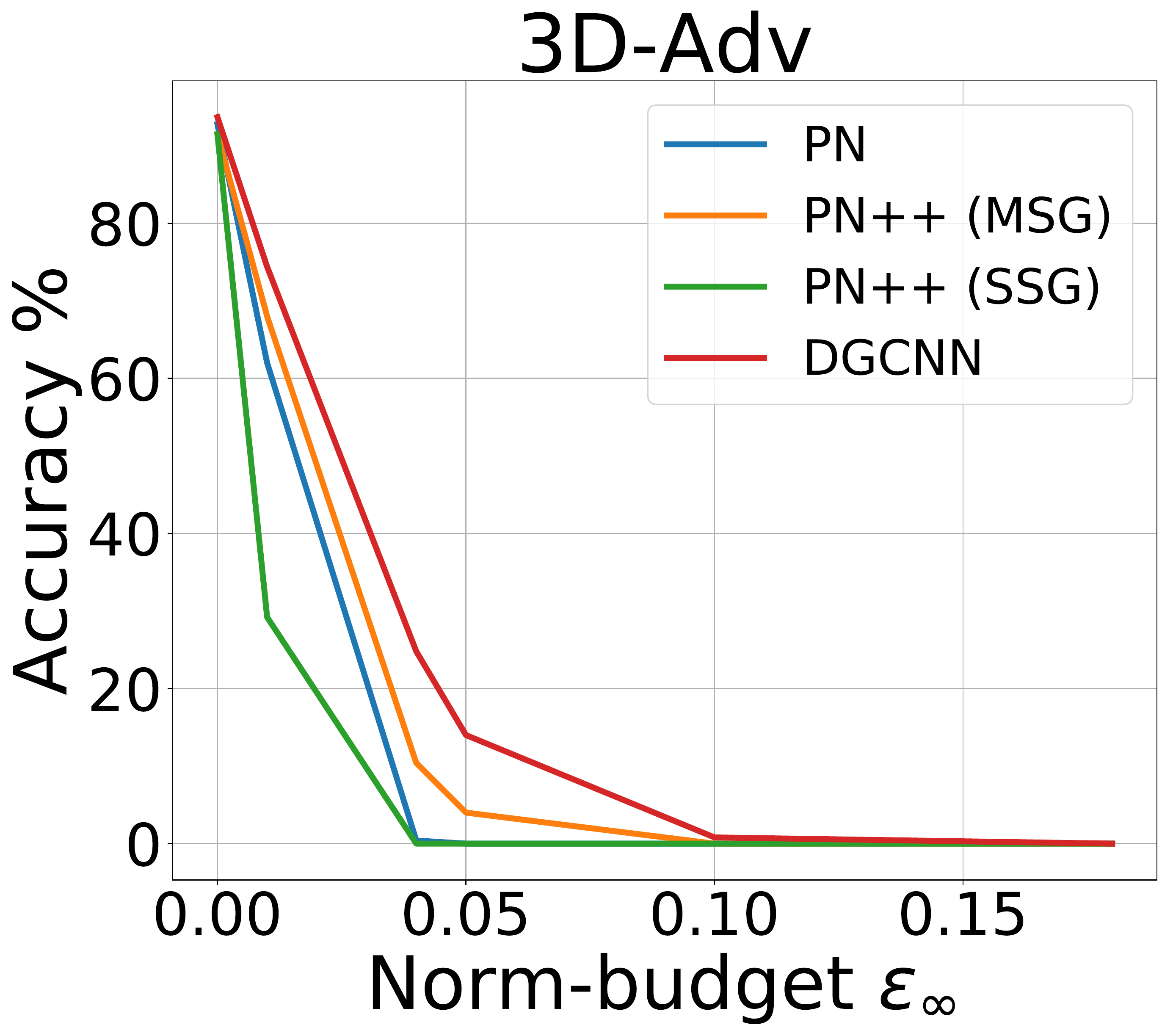} &
\includegraphics[width=0.33\columnwidth]{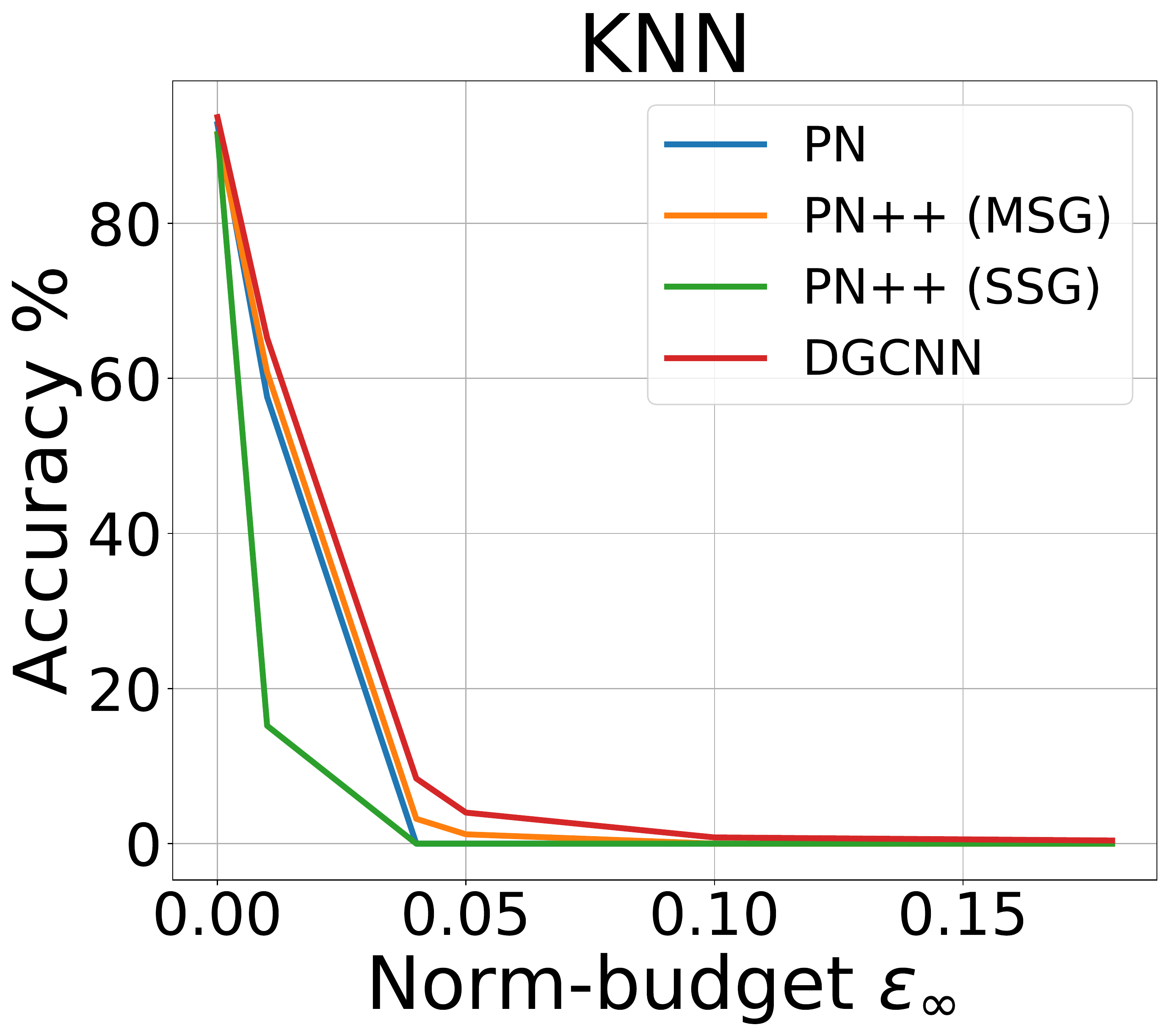} &
\includegraphics[width=0.33\columnwidth]{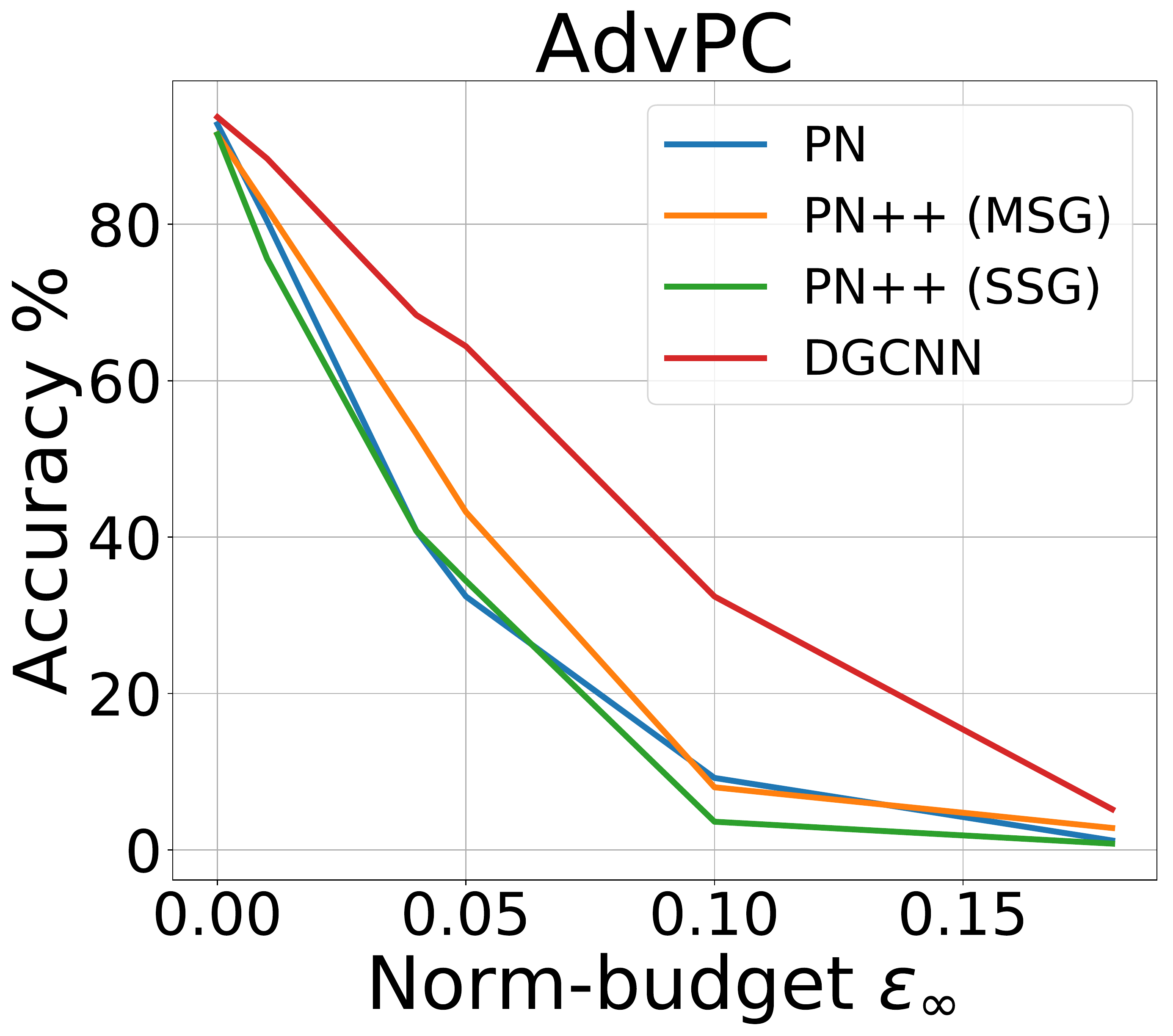} \\
\end{tabular}

\caption{\small \textbf{Sensitivity of Architectures in $\ell_\infty$}: We evaluate the sensitivity of each of the four networks for increasing norm-budget. For each network, we plot the classification accuracy under 3D-Adv perturbation \cite{pcattack} (\textit{left}), KNN attack \cite{robustshapeattack} (\textit{middle}), and our AdvPC attack (\textit{right}). Overall, DGCNN \cite{dgcn} is affected the least by adversarial perturbation.}
\label{fig:sup-:sensitivity-I}
  \vspace{-8pt}
\end{figure}

\begin{figure}[]
\tabcolsep=0.03cm
\begin{tabular}{ccc}
\includegraphics[width=0.33\columnwidth]{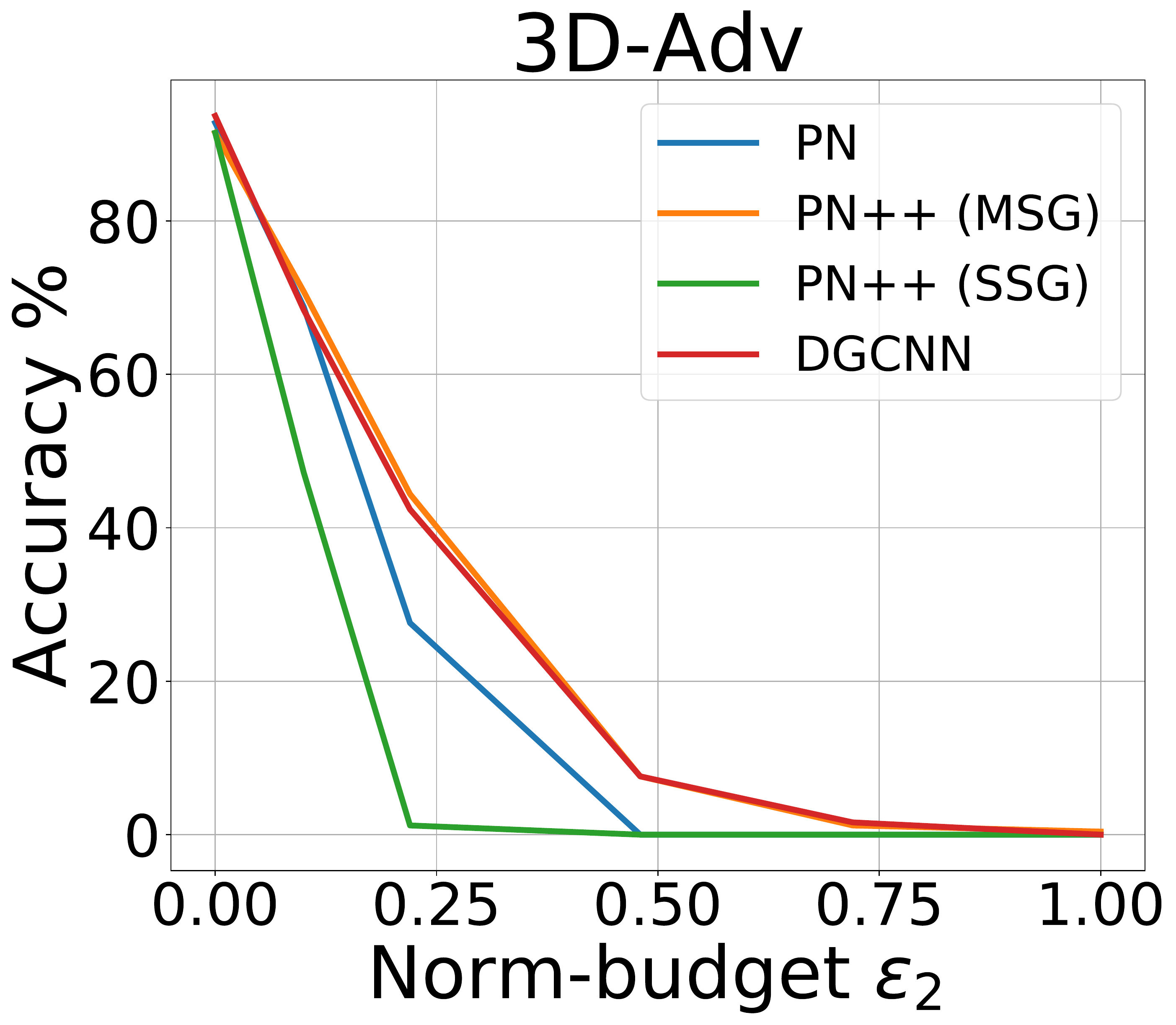} &
\includegraphics[width=0.33\columnwidth]{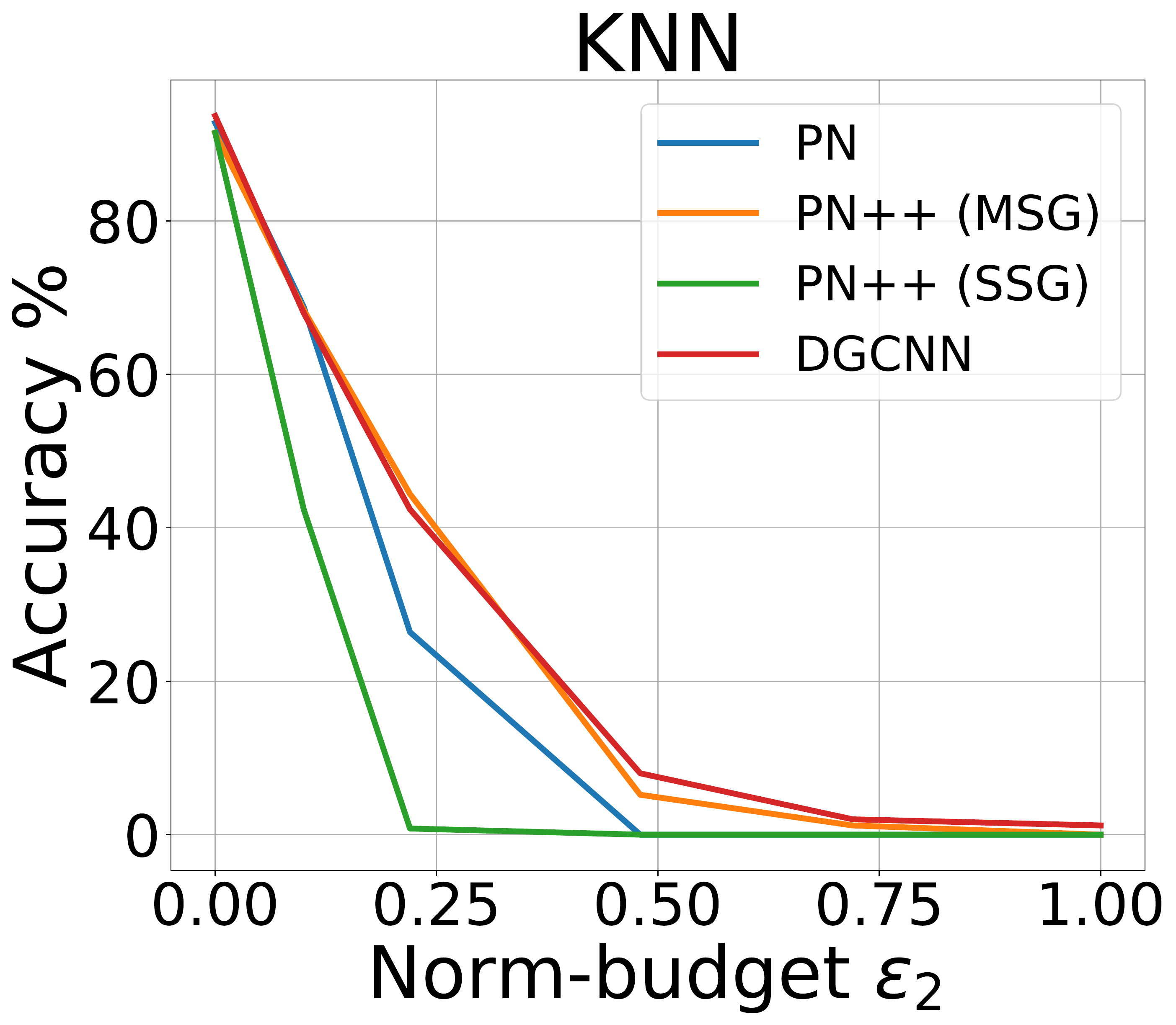} &
\includegraphics[width=0.33\columnwidth]{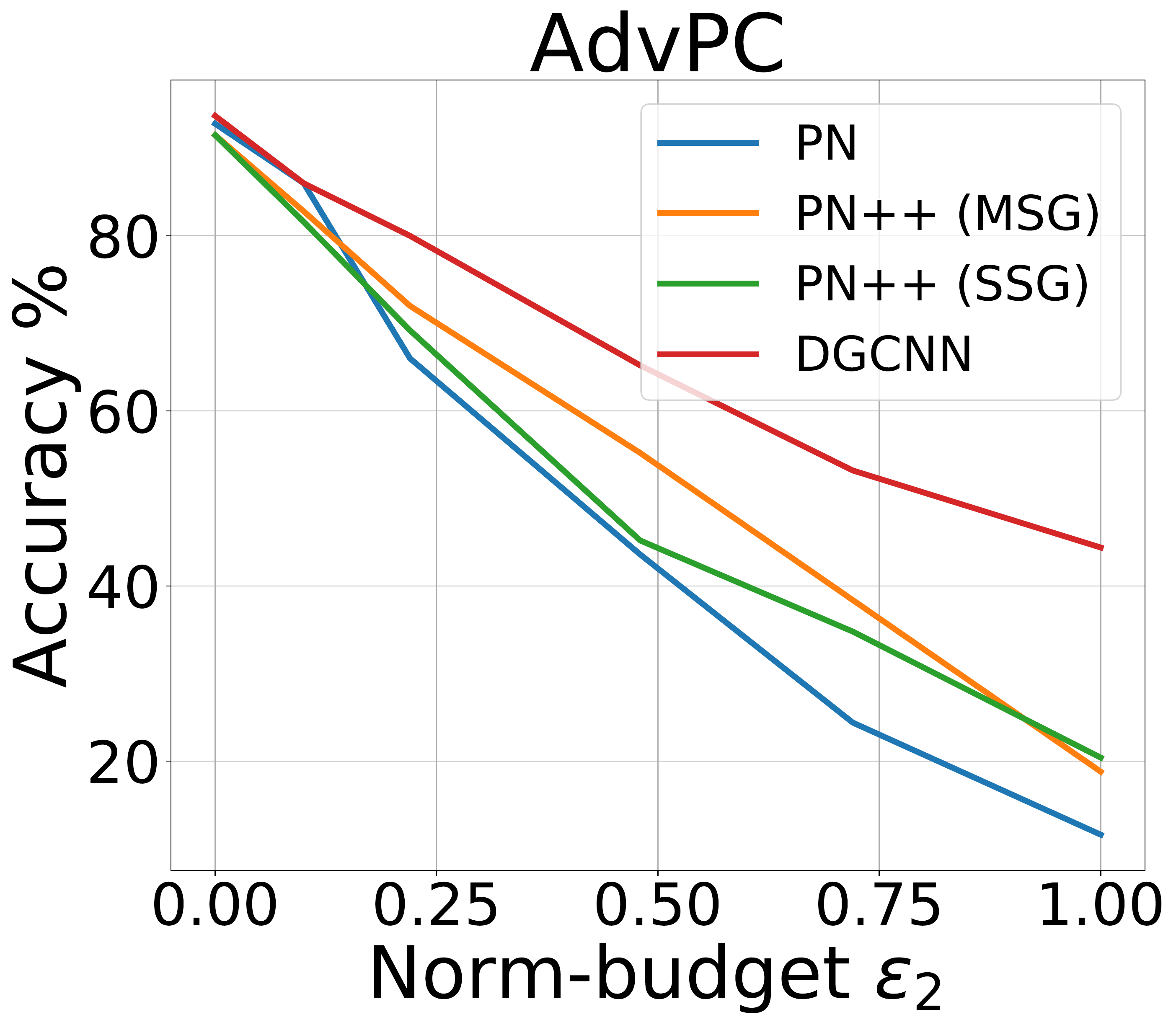} \\
\end{tabular}

\caption{\small \textbf{Sensitivity of Architectures in $\ell_2$}: We evaluate the sensitivity of each of the four networks for increasing norm-budget. For each network, we plot the classification accuracy under 3D-Adv perturbation \cite{pcattack} (\textit{left}), KNN attack \cite{robustshapeattack} (\textit{middle}), and our AdvPC attack (\textit{right}). Overall, DGCNN \cite{dgcn} is affected the least by adversarial perturbation.}
\label{fig:sup-:sensitivity-2}
  \vspace{-8pt}
\end{figure}

\vspace{-6pt}
\figLabel{\ref{fig:sup-:sensitivity-I}} and \figLabel{\ref{fig:sup-:sensitivity-2}} plot the sensitivity of the various networks when they are subject to input perturbations of varying norm-budgets $\epsilon_\infty$ and $\epsilon_2$ respectively. We measure the classification accuracy of each  network under our AdvPC attack ($\gamma=0.25$), 3D-Adv \cite{pcattack}, and KNN attack \cite{robustshapeattack}. We observe that 
DGCNN \cite{dgcn} tends to be the most robust to adversarial perturbations in general. This might be explained by the fact that the convolution neighborhoods in DGCNN are dynamically updated across layers and iterations. This dynamic behavior in network structure may hinder the effect of the attack because gradient directions can change significantly from one iteration to another. This leads to failing attacks and higher robustness for DGCNN \cite{dgcn}.

\subsection{Effect of the Auto-Encoder (AE)} \label{sec:sup-intrep}
In \figLabel{\ref{fig:sup-comparisontoilet}}, we show an example of how AE reconstruction preserves the details of the unperturbed point cloud and does not change the classifier prediction. When a perturbed point cloud passes through the AE, it recovers a natural-looking shape. The AE's ability to reconstruct natural-looking 3D point clouds from various perturbed inputs might explain why it is a strong defense against attacks in \secLabel{\ref{sec:sup-def-unt}}. Another observation from \figLabel{\ref{fig:sup-comparisontoilet}} is that when we fix the target $t^\prime$ and do not enforce a specific incorrect target $t^{\prime\prime}$ (\ie untargeted attack setting) for the data adversarial loss on the reconstructed point cloud  in the AdvPC attack (\eqLabel{\ref{eq:sup-final-objective}}), the optimization mechanism tends to pick $t^{\prime\prime}$ to be a \textit{similar} class to the correct one. For example, a \emph{Toilet} point cloud perturbed by AdvPC can be transformed into a \emph{Chair} (similar in appearance to a toilet), if reconstructed by the AE. %
This effect is not observed for the other attacks \cite{pcattack,robustshapeattack}, which do not consider the data distribution and optimize solely for the network.  %

\begin{figure}[]
\tabcolsep=0.03cm
\resizebox{\textwidth}{!}{%
\begin{tabular}{cc|cc|cc|cc}
\toprule
\multicolumn{2}{c}{\specialcell{unperturbed \\ point cloud}}  & \multicolumn{2}{c}{3D-adv \cite{pcattack}} & \multicolumn{2}{c}{ KNN \cite{robustshapeattack}} & \multicolumn{2}{c}{AdvPC (ours)} \\ \hline
before AE & after AE & before AE & after AE  & before AE & after AE   & before AE & after AE \\ 
\includegraphics[trim={1cm 0 1.1cm 0},clip,width = 0.124\columnwidth]{images/analysis/interpret/orig_ptc.png} &
\includegraphics[trim={1cm 0 1.1cm 0},clip,width = 0.124\columnwidth]{images/analysis/interpret/orig__rec_ptc.png} &
\includegraphics[trim={1cm 0 1.1cm 0},clip,width = 0.124\columnwidth]{images/analysis/interpret/adv_ptc.png} &
\includegraphics[trim={1cm 0 1.1cm 0},clip,width = 0.124\columnwidth]{images/analysis/interpret/adv_rec_ptc.png} &
\includegraphics[trim={1cm 0 1.1cm 0},clip,width = 0.124\columnwidth]{images/analysis/interpret/knn_ptc.png} &
\includegraphics[trim={1cm 0 1.1cm 0},clip,width = 0.120\columnwidth]{images/analysis/interpret/knn_rec_ptc.png} &
\includegraphics[trim={1cm 0 1.1cm 0},clip,width = 0.124\columnwidth]{images/analysis/interpret/nat_ptc.png} &
\includegraphics[trim={1cm 0 1.1cm 0},clip,width = 0.124\columnwidth]{images/analysis/interpret/nat_rec_ptc.png} \\
PN: & PN: &PN:  & PN: &PN:  & PN:  &PN:  & PN: \\
\textbf{Toilet} \color{ForestGreen}\cmark & \textbf{Toilet} \color{ForestGreen}\cmark 
 & \textbf{Bed} \color{red}\xmark  & \textbf{Toilet} \color{ForestGreen}\cmark & \textbf{Bed} \color{red}\xmark  & \textbf{Toilet} \color{ForestGreen}\cmark & \textbf{Bed}  \color{red}\xmark  & \textbf{Chair}  \color{red}\xmark  \\
 \bottomrule
\end{tabular}
}
\caption{\small \textbf{Effect of the Auto-Encoder (AE):} The AE does not affect the unperturbed point cloud (classified correctly by PN before and after AE). The AE cleans the perturbed point cloud by 3D-Adv and KNN \cite{pcattack,robustshapeattack}, which allows PN to predict the correct class label. However, our AdvPC attack can fool PN before and after AE reconstruction. Perturbed samples by AdvPC, if passed through the AE, transform into similar looking objects but from different classes (Chair looks similar to Toilet).
}
\label{fig:sup-comparisontoilet}
  \vspace{-8pt}
\end{figure}

\clearpage
\subsection{Ablation Study on the Losses} \label{sec:sup-ablation-losses}
We ablate each component of our pipeline and show their effect in our attacks. We evaluate this components by looking Attack Success Rate (ASR), transferability, and the final norm obtained under the attack. In theses experiments, we allow unconstrained attacks as well as constrained attacks. We show the effect of optimizing using EMD, CD, $\ell_2$, and $\ell_\infty$. We show the results on Tables \ref{tbl:soft-pn},\ref{tbl:soft-pn1},\ref{tbl:soft-pn2},\ref{tbl:soft-gcn} for all the four networks. We observe transferability is better when using hard constraints. Constraining the attack norm allows the optimization to learn more from the AE data distribution. The EMD doesn't work well while the Chamfer loss is comparable to the $\ell_2$ loss.

\begin{table*}[h!]
\footnotesize
\setlength{\tabcolsep}{4pt} %
\renewcommand{\arraystretch}{1.1} %
\centering
\resizebox{\hsize}{!}{
\begin{tabular}{cccccc|cccccc} 
\toprule
\multicolumn{6}{c|}{\textbf{Attack Setup}} & \multicolumn{6}{c}{\textbf{Results}}\\
\textbf{soft CD} & \textbf{soft EMD} & \textbf{soft $\ell_2$} & \textbf{hard $\ell_\infty$} & \textbf{hard $\ell_2$} & \textbf{AE} & \textbf{CD} & \textbf{EMD} & \textbf{$\ell_\infty$} & \textbf{$\ell_2$} & \textbf{ASR}& \textbf{TR}     \\
\midrule
\checkmark & - & - & - & - & - &    \textbf{0.15} &  4.25 & 0.12 & \textbf{0.31} & \textbf{100} &  9.02 \\ \hline
\checkmark & - & - & - & - & \checkmark &   0.19 &  5.01 & 0.13 & 0.36 &  99.69 &  9.51 \\ \hline
- & \checkmark & - & - & - & - &   0.17 &  2.83 & 0.23 & 0.39 &  68.36 &  9.01 \\ \hline
- & \checkmark & - & - & - & \checkmark &  0.16 &  \textbf{2.53} & 0.25 & 0.37 &  18.04 &  7.35 \\ \hline
- & - & \checkmark & - & - & - &    0.16 &  4.38 & 0.11 & \textbf{0.31} & \textbf{100} &  8.92 \\ \hline
- & - & \checkmark & - & - & \checkmark &    0.21 &  5.22 & 0.13 & 0.36 & \textbf{100} &  9.35 \\ \hline
- & - & - & \checkmark & - & - &  0.49 & 12.37 & \textbf{0.04} & 0.55 & \textbf{100} &  9.16 \\ \hline
- & - & - & \checkmark & - & \checkmark &   0.73 & 13.66 & 0.07 & 0.72 &  96.93 & \textbf{13.14} \\ \hline
- & - & - & - & \checkmark & - &   0.26 &  7.41 & 0.09 & 0.38 & \textbf{100} &  8.87 \\ \hline
- & - & - & - & \checkmark & \checkmark &   0.37 &  7.35 & 0.16 & 0.48 &  99.87 & 11.08 \\ 
 \bottomrule
\end{tabular}
}
\caption{\small \textbf{Soft vs Hard on PointNet}: study the effect of every bit of the loss on the norms , Attack Success Rate (ASR) and Transferability (TR) under unconstrained setup vs constrained setup in PointNet \cite{pointnet}.($\epsilon_{\infty} =0.1$ , $\epsilon_{2} =1.8$),$\lambda = 1$,$\gamma=0.5$. Please refer to \secLabel{\ref{sec:sup-full-formulation}} for details. \textbf{Bold} numbers are the best.}
\vspace{-6pt}
\label{tbl:soft-pn}
\end{table*}

\begin{table*}[h!]
\footnotesize
\setlength{\tabcolsep}{4pt} %
\renewcommand{\arraystretch}{1.1} %
\centering
\resizebox{\hsize}{!}{
\begin{tabular}{cccccc|cccccc} 
\toprule
\multicolumn{6}{c|}{\textbf{Attack Setup}} & \multicolumn{6}{c}{\textbf{Results}}\\
\textbf{soft CD} & \textbf{soft EMD} & \textbf{soft $\ell_2$} & \textbf{hard $\ell_\infty$} & \textbf{hard $\ell_2$} & \textbf{AE} & \textbf{CD} & \textbf{EMD} & \textbf{$\ell_\infty$} & \textbf{$\ell_2$} & \textbf{ASR}& \textbf{TR}     \\
\midrule
\checkmark & - & - & - & - & - &    1.01 & 25.80 & 0.19 & 1.00 &  99.78 & 11.57 \\ \hline
\checkmark & - & - & - & - & \checkmark &   0.88 & 26.25 & 0.21 & 1.15 &  95.69 & 12.19 \\ \hline
- & \checkmark & - & - & - & - &   0.21 &  5.04 & 0.23 & 0.56 &  14.58 &  6.83 \\ \hline
- & \checkmark & - & - & - & \checkmark &  \textbf{0.07} &  \textbf{2.23} & 0.12 & \textbf{0.20} &   2.31 &  6.61 \\ \hline
- & - & \checkmark & - & - & - &    1.35 & 26.58 & 0.20 & 0.96 &  99.96 & 13.38 \\ \hline
- & - & \checkmark & - & - & \checkmark &    1.43 & 26.42 & 0.22 & 0.98 & \textbf{100} & 16.61 \\ \hline
- & - & - & \checkmark & - & - &  3.71 & 53.83 & \textbf{0.06} & 1.84 &  94.71 & 18.46 \\ \hline
- & - & - & \checkmark & - & \checkmark &   2.53 & 38.78 & 0.10 & 1.45 &  97.64 & \textbf{25.82} \\ \hline
- & - & - & - & \checkmark & - &   0.59 & 15.95 & 0.10 & 0.58 & \textbf{100} &  8.84 \\ \hline
- & - & - & - & \checkmark & \checkmark &   0.93 & 20.08 & 0.15 & 0.75 &  99.20 & 11.91 \\ 
 \bottomrule
\end{tabular}
}
\caption{\small \textbf{Soft vs Hard on PointNet++ MSG}: study the effect of every bit of the loss on the norms , Attack Success Rate (ASR) and Transferability (TR) under unconstrained setup vs constrained setup in PointNet++ MSG \cite{pointnet++}. ($\epsilon_{\infty} =0.18$ , $\epsilon_{2} =1.8$),$\lambda = 1$,$\gamma=0.5$. Please refer to \secLabel{\ref{sec:sup-full-formulation}} for details. \textbf{Bold} numbers are the best.}
\vspace{-6pt}
\label{tbl:soft-pn1}
\end{table*}

\begin{table*}[h!]
\footnotesize
\setlength{\tabcolsep}{4pt} %
\renewcommand{\arraystretch}{1.1} %
\centering
\resizebox{\hsize}{!}{
\begin{tabular}{cccccc|cccccc} 
\toprule
\multicolumn{6}{c|}{\textbf{Attack Setup}} & \multicolumn{6}{c}{\textbf{Results}}\\
\textbf{soft CD} & \textbf{soft EMD} & \textbf{soft $\ell_2$} & \textbf{hard $\ell_\infty$} & \textbf{hard $\ell_2$} & \textbf{AE} & \textbf{CD} & \textbf{EMD} & \textbf{$\ell_\infty$} & \textbf{$\ell_2$} & \textbf{ASR}& \textbf{TR}     \\
\midrule
\checkmark & - & - & - & - & - &  0.25 &  9.39 & 0.07 & 0.37 & \textbf{100} & 7.01 \\\hline
\checkmark & - & - & - & - & \checkmark &  0.25 &  9.38 & 0.08 & 0.39 &  99.51 & 7.17 \\\hline
- & \checkmark & - & - & - & - & 0.08 &  3.71 & 0.09 & 0.28 &  37.20 & 6.74 \\\hline
- & \checkmark & - & - & - & \checkmark & \textbf{0.07} &  \textbf{2.95} & 0.10 & \textbf{0.24} &   4.84 & 6.52 \\\hline
- & - & \checkmark & - & - & - &  0.30 &  9.90 & 0.07 & 0.39 & \textbf{100} & 6.92 \\\hline
- & - & \checkmark & - & - & \checkmark &  0.28 &  9.63 & 0.07 & 0.38 & \textbf{100} & 7.56 \\\hline
- & - & - & \checkmark & - & - & 1.20 & 24.52 & 0.02 & 0.83 &  96.80 & \textbf{7.84} \\\hline
- & - & - & \checkmark & - & \checkmark & 0.80 & 17.24 & 0.05 & 0.70 & \textbf{100} & 7.72 \\\hline
- & - & - & - & \checkmark & - &  0.19 &  8.08 & \textbf{0.04} & 0.30 & \textbf{100} & 6.99 \\\hline
- & - & - & - & \checkmark & \checkmark & 0.46 & 12.42 & 0.09 & 0.50 & \textbf{100} & 7.44 \\
 \bottomrule
\end{tabular}
}
\caption{\small \textbf{Soft vs Hard on PointNet++ SSG}: study the effect of every bit of the loss on the norms , Attack Success Rate (ASR) and Transferability (TR) under unconstrained setup vs constrained setup in PointNet++ SSG \cite{pointnet++}.($\epsilon_{\infty} =0.1$ , $\epsilon_{2} =1.8$),$\lambda = 1$,$\gamma=0.5$. Please refer to \secLabel{\ref{sec:sup-full-formulation}} for details. \textbf{Bold} numbers are the best.}
\vspace{-6pt}
\label{tbl:soft-pn2}
\end{table*}
\clearpage
\begin{table*}[h!]
\footnotesize
\setlength{\tabcolsep}{4pt} %
\renewcommand{\arraystretch}{1.1} %
\centering
\resizebox{\hsize}{!}{
\begin{tabular}{cccccc|cccccc} 
\toprule
\multicolumn{6}{c|}{\textbf{Attack Setup}} & \multicolumn{6}{c}{\textbf{Results}}\\
\textbf{soft CD} & \textbf{soft EMD} & \textbf{soft $\ell_2$} & \textbf{hard $\ell_\infty$} & \textbf{hard $\ell_2$} & \textbf{AE} & \textbf{CD} & \textbf{EMD} & \textbf{$\ell_\infty$} & \textbf{$\ell_2$} & \textbf{ASR}& \textbf{TR}     \\
\midrule
\checkmark & - & - & - & - & - &   1.06 & 32.22 & 0.20 & 1.55 & 67.91 & 10.46 \\ \hline
\checkmark & - & - & - & - & \checkmark &  0.71 & 25.03 & 0.17 & 1.18 & 41.07 &  9.21 \\ \hline
- & \checkmark & - & - & - & - &  0.03 &  2.47 & 0.07 & 0.14 &  2.07 &  7.18 \\ \hline
- & \checkmark & - & - & - & \checkmark & \textbf{0.01} &  \textbf{1.62 }& \textbf{0.01} &\textbf{ 0.05} &  0.76 &  7.21 \\ \hline
- & - & \checkmark & - & - & - &   2.81 & 39.95 & 0.28 & 1.53 & \textbf{99.20} & 23.23 \\ \hline
- & - & \checkmark & - & - & \checkmark &   2.89 & 40.55 & 0.31 & 1.58 & 96.89 & 29.91 \\ \hline
- & - & - & \checkmark & - & - & 4.39 & 53.49 & 0.12 & 2.12 & 86.67 & 26.22 \\ \hline
- & - & - & \checkmark & - & \checkmark &  5.10 & 58.24 & 0.16 & 2.40 & 83.56 & \textbf{35.59} \\ \hline
- & - & - & - & \checkmark & - &  2.46 & 39.85 & 0.23 & 1.45 & 99.82 & 23.45 \\ \hline
- & - & - & - & \checkmark & \checkmark &  2.82 & 43.19 & 0.30 & 1.63 & 98.80 & 33.26 \\ 
 \bottomrule
\end{tabular}
}
\caption{\small \textbf{Soft vs Hard on DGCNN}: study the effect of every bit of the loss on the norms , Attack Success Rate (ASR) and Transferability (TR) under unconstrained setup vs constrained setup in DGCNN \cite{dgcn}. ($\epsilon_{\infty} =0.18$ , $\epsilon_{2} =2.8$),$\lambda = 1$,$\gamma=0.5$. Please refer to \secLabel{\ref{sec:sup-full-formulation}} for details. \textbf{Bold} numbers are the best.}
\vspace{-6pt}
\label{tbl:soft-gcn}
\end{table*}

\clearpage
\section{Defenses Results (Targeted Attacks)} \label{sec:sup-def-tar}
We note from targeted attack results in Tables \ref{tbl:breaking-gcn-I-tar},\ref{tbl:breaking-pn2-I-tar},\ref{tbl:breaking-pn1-I-tar},\ref{tbl:breaking-pn-I-tar} that our AdvPC still outperforms the other baselines in most defenses but fail in some defenses. This can be explained because the targeted attacks with specific target label $t^\prime$ in \eqLabel{\ref{eq:sup-pre-final-objective}} is too strict given that the reconstruction of the AE needs to fool the classifier to unspecified label $t^{\prime\prime}$ that might be different from $t^\prime$. This restriction makes the optimization in \eqLabel{\ref{eq:sup-final-objective}} very difficult to optimize and hence leads to less successful attacks.

\begin{table}[]
\footnotesize
\centering
\vspace{-8pt}
\setlength{\tabcolsep}{4pt} %
\renewcommand{\arraystretch}{1} %
\begin{tabular}{c|ccc|ccc} 
\toprule
 & \multicolumn{3}{c}{$\epsilon_\infty = 0.18$} & \multicolumn{3}{c}{$\epsilon_\infty = 0.45$} \\
\textbf{Defenses} & \textbf{\specialcell{3D-Adv\\ \cite{pcattack} }} & \textbf{\specialcell{KNN \\ \cite{robustshapeattack}}} & \textbf{\specialcell{AdvPC \\(ours)}}  & \textbf{\specialcell{3D-Adv \\\cite{pcattack}}} & \textbf{\specialcell{KNN \\ \cite{robustshapeattack}}} &  \textbf{\specialcell{AdvPC \\(ours)}}    \\
\midrule
No defense & 77.1 & \textbf{77.6} & 66.4 & \textbf{85.4} & 80.1 & 77.7 \\
AE (newly trained) &   13.1 &  9.8 & \textbf{16.5} & 13.9 &  9.8 & \textbf{18.0} \\
Adv Training \cite{pcattack} &   5.1 &  2.5 &  \textbf{6.2} &  5.2 &  2.2 &  \textbf{5.3} \\
SOR \cite{Deflecting} & 24.1 & \textbf{25.6} & 21.9 & 21.2 &\textbf{ 26.8} & 19.2 \\
DUP Net \cite{Deflecting}  & \textbf{32.0} & 30.3 & 27.2 & 30.3 & \textbf{36.5} & 26.7\\ 
SRS \cite{Deflecting} &  34.8 & \textbf{36.2} & \textbf{36.2} & 31.8 & \textbf{38.7} & 30.4 \\
 \bottomrule
\end{tabular}
\caption{\small \textbf{Attacking Point Cloud Defenses ($\ell_\infty$ Targeted DGCNN):}  We evaluate targeted attacks using norm-budgets of $\epsilon_\infty = 0.18$ and $\epsilon_\infty = 0.45$ with DGCNN \cite{dgcn} as the victim network under different defenses for 3D point clouds. Similar to before, we report 1 - accuracy (\textbf{higher} indicates better attack). AdvPC consistently outperforms the other attacks \cite{pcattack,robustshapeattack} for all defenses. %
Note that both the attacks \textit{and} evaluations are performed on DGCNN, which has an accuracy of 93.7\% without input perturbations (for reference).
}
\label{tbl:breaking-gcn-I-tar}
\vspace{-6pt}
\end{table}

\begin{table}[]
\footnotesize
\centering
\vspace{-8pt}
\setlength{\tabcolsep}{4pt} %
\renewcommand{\arraystretch}{1} %
\begin{tabular}{c|ccc|ccc} 
\toprule
 & \multicolumn{3}{c}{$\epsilon_\infty = 0.18$} & \multicolumn{3}{c}{$\epsilon_\infty = 0.45$} \\
\textbf{Defenses} & \textbf{\specialcell{3D-Adv\\ \cite{pcattack} }} & \textbf{\specialcell{KNN \\ \cite{robustshapeattack}}} & \textbf{\specialcell{AdvPC \\(ours)}}  & \textbf{\specialcell{3D-Adv \\\cite{pcattack}}} & \textbf{\specialcell{KNN \\ \cite{robustshapeattack}}} &  \textbf{\specialcell{AdvPC \\(ours)}}    \\
\midrule
No defense &                      99.5 &\textbf{100}& 98.3 & 99.8 &\textbf{100}& 98.5 \\
AE (newly trained) &              13.0 &  12.9 & \textbf{15.5} & 12.6 &  12.8 & \textbf{14.7} \\
Adv Training \cite{pcattack} &    9.3 &  10.3 & \textbf{25.2} &  5.3 &  10.7 & \textbf{14.6} \\
SOR \cite{Deflecting} &           16.9 &  16.6 & \textbf{18.0} & 13.5 &  \textbf{20.4} & 15.2 \\
DUP Net \cite{Deflecting}  & 17.3 & 17.4 & \textbf{18.5} & 15.8 & \textbf{18.7} & 16.9\\ 
SRS \cite{Deflecting} &           17.3 &  17.4 & \textbf{60.8} & 35.4 &  51.4 & \textbf{53.0} \\
 \bottomrule
\end{tabular}
\caption{\small \textbf{Attacking Point Cloud Defenses ($\ell_\infty$ Targeted PointNet++ SSG):}  We evaluate targeted attacks using norm-budgets of $\epsilon_\infty = 0.18$ and $\epsilon_\infty = 0.45$ with PointNet++ SSG \cite{pointnet++} as the victim network under different defenses for 3D point clouds. Similar to before, we report 1 - accuracy (\textbf{higher} indicates better attack). AdvPC consistently outperforms the other attacks \cite{pcattack,robustshapeattack} for all defenses. %
Note that both the attacks \textit{and} evaluations are performed on PointNet++ SSG, which has an accuracy of 91.5\% without input perturbations (for reference).
}
\label{tbl:breaking-pn2-I-tar}
\vspace{-6pt}
\end{table}

\begin{table}[]
\footnotesize
\centering
\vspace{-8pt}
\setlength{\tabcolsep}{4pt} %
\renewcommand{\arraystretch}{1} %
\begin{tabular}{c|ccc|ccc} 
\toprule
 & \multicolumn{3}{c}{$\epsilon_\infty = 0.18$} & \multicolumn{3}{c}{$\epsilon_\infty = 0.45$} \\
\textbf{Defenses} & \textbf{\specialcell{3D-Adv\\ \cite{pcattack} }} & \textbf{\specialcell{KNN \\ \cite{robustshapeattack}}} & \textbf{\specialcell{AdvPC \\(ours)}}  & \textbf{\specialcell{3D-Adv \\\cite{pcattack}}} & \textbf{\specialcell{KNN \\ \cite{robustshapeattack}}} &  \textbf{\specialcell{AdvPC \\(ours)}}    \\
\midrule
No defense &                      96.6 & \textbf{99.9} & 94.7 & 99.2 & \textbf{99.9} & 97.6 \\
AE (newly trained) &              13.6 & 12.8 & \textbf{16.7} & 16.1 & 12.0 & \textbf{23.2} \\
Adv Training \cite{pcattack} &     7.2 &  3.6 & \textbf{11.8} &  6.6 &  3.6 & \textbf{12.7} \\
SOR \cite{Deflecting} &           \textbf{32.1} & 27.1 & 31.6 & 23.8 & \textbf{31.3} & 25.0 \\
DUP Net \cite{Deflecting}  & \textbf{36.6} & 9.6 & 36.2 & 27.6 & \textbf{31.3} & 30.6\\ 
SRS \cite{Deflecting} &           46.1 & 44.9 & \textbf{57.0} & 50.3 & 41.4 & \textbf{60.3} \\
 \bottomrule
\end{tabular}
\caption{\small \textbf{Attacking Point Cloud Defenses ($\ell_\infty$ Targeted PointNet++ MSG):}  We evaluate targeted attacks using norm-budgets of $\epsilon_\infty = 0.18$ and $\epsilon_\infty = 0.45$ with PointNet++ MSG \cite{pointnet++} as the victim network under different defenses for 3D point clouds. Similar to before, we report 1 - accuracy (\textbf{higher} indicates better attack). AdvPC consistently outperforms the other attacks \cite{pcattack,robustshapeattack} for all defenses. %
Note that both the attacks \textit{and} evaluations are performed on PointNet++ MSG, which has an accuracy of 91.5\% without input perturbations (for reference).
}
\label{tbl:breaking-pn1-I-tar}
\vspace{-6pt}
\end{table}

\begin{table}[]
\footnotesize
\centering
\vspace{-8pt}
\setlength{\tabcolsep}{4pt} %
\renewcommand{\arraystretch}{1} %
\begin{tabular}{c|ccc|ccc} 
\toprule
 & \multicolumn{3}{c}{$\epsilon_\infty = 0.18$} & \multicolumn{3}{c}{$\epsilon_\infty = 0.45$} \\
\textbf{Defenses} & \textbf{\specialcell{3D-Adv\\ \cite{pcattack} }} & \textbf{\specialcell{KNN \\ \cite{robustshapeattack}}} & \textbf{\specialcell{AdvPC \\(ours)}}  & \textbf{\specialcell{3D-Adv \\\cite{pcattack}}} & \textbf{\specialcell{KNN \\ \cite{robustshapeattack}}} &  \textbf{\specialcell{AdvPC \\(ours)}}    \\
\midrule
No defense &                     \textbf{100}&\textbf{100}& 97.4 &\textbf{100}&\textbf{100}& 98.4 \\
AE (newly trained) &                9.9 &   9.4 & \textbf{12.6} &   8.5 &   0.0 &  \textbf{9.7} \\
Adv Training \cite{pcattack} &     12.2 &  14.6 & \textbf{22.0} &  11.3 &  \textbf{28.0} & 13.2 \\
SOR \cite{Deflecting} &            \textbf{11.2} &  10.9 & 10.7 &   \textbf{9.6} &   4.0 &  8.6 \\
DUP Net \cite{Deflecting}  & 8.5 & \textbf{9.7} & 7.8 & 8.0 & \textbf{9.6} & 7.8\\ 
SRS \cite{Deflecting} &            70.7 &  63.8 & \textbf{81.4} &  52.0 &  52.0 & \textbf{63.7} \\
 \bottomrule
\end{tabular}
\caption{\small \textbf{Attacking Point Cloud Defenses ($\ell_\infty$ Targeted PointNet):}  We evaluate targeted attacks using norm-budgets of $\epsilon_\infty = 0.18$ and $\epsilon_\infty = 0.45$ with PointNet \cite{pointnet} as the victim network under different defenses for 3D point clouds. Similar to before, we report 1 - accuracy (\textbf{higher} indicates better attack). AdvPC consistently outperforms the other attacks \cite{pcattack,robustshapeattack} for all defenses. %
Note that both the attacks \textit{and} evaluations are performed on PointNet, which has an accuracy of 92.8\% without input perturbations (for reference).
}
\label{tbl:breaking-pn-I-tar}
\vspace{-6pt}
\end{table}

\clearpage

\section{Other Tried Approaches (Less Successful)}
\subsection{Point Cloud GAN:}
\begin{figure*}[h!]
\centering
\tabcolsep=0.03cm
\begin{tabular}{cc}

 \includegraphics[trim={0cm 1.6cm 0cm 0cm},clip, width=0.5\columnwidth]{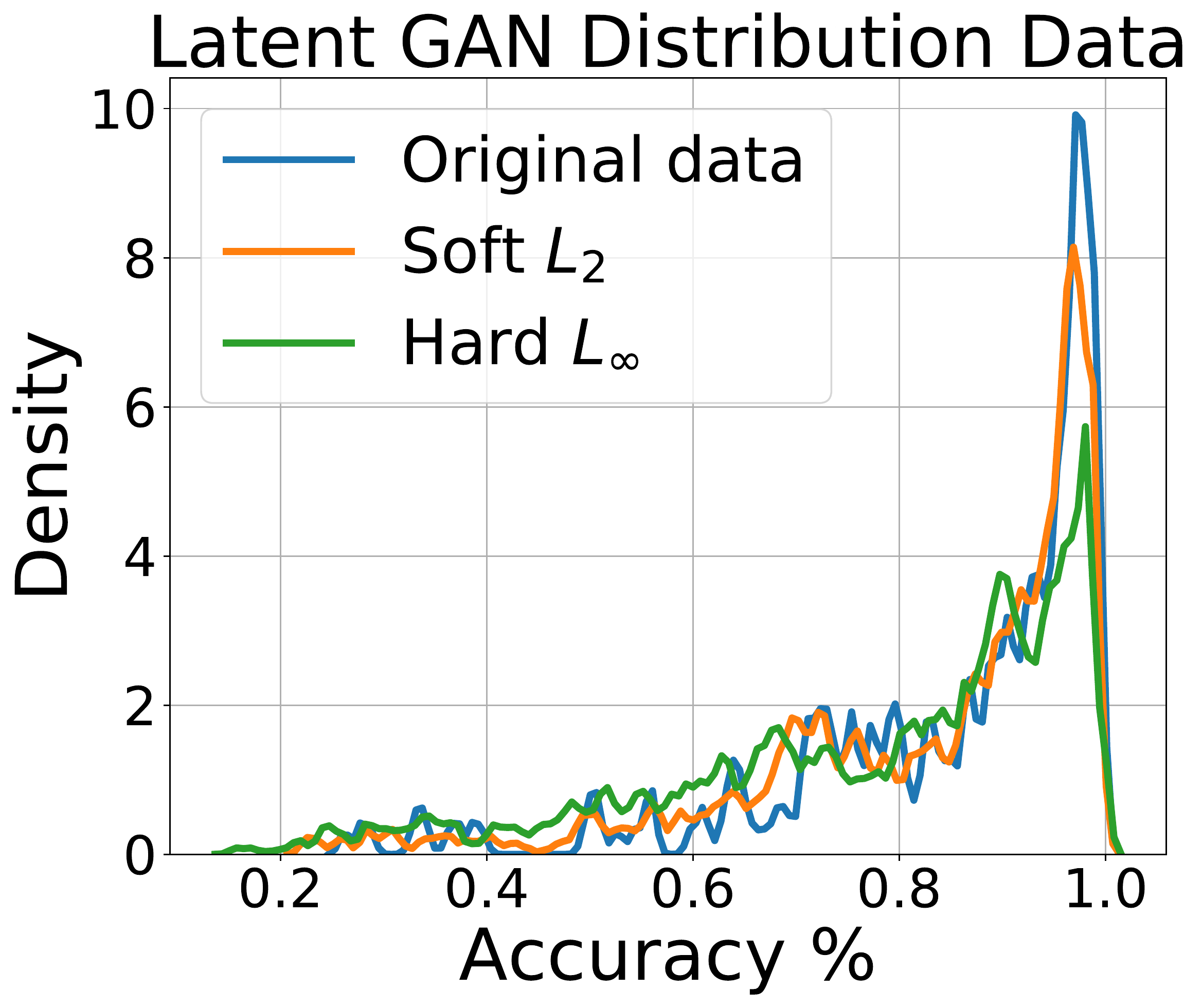} &
 \includegraphics[trim={0cm 1.6cm 0cm 0cm},clip,width=0.5\columnwidth]{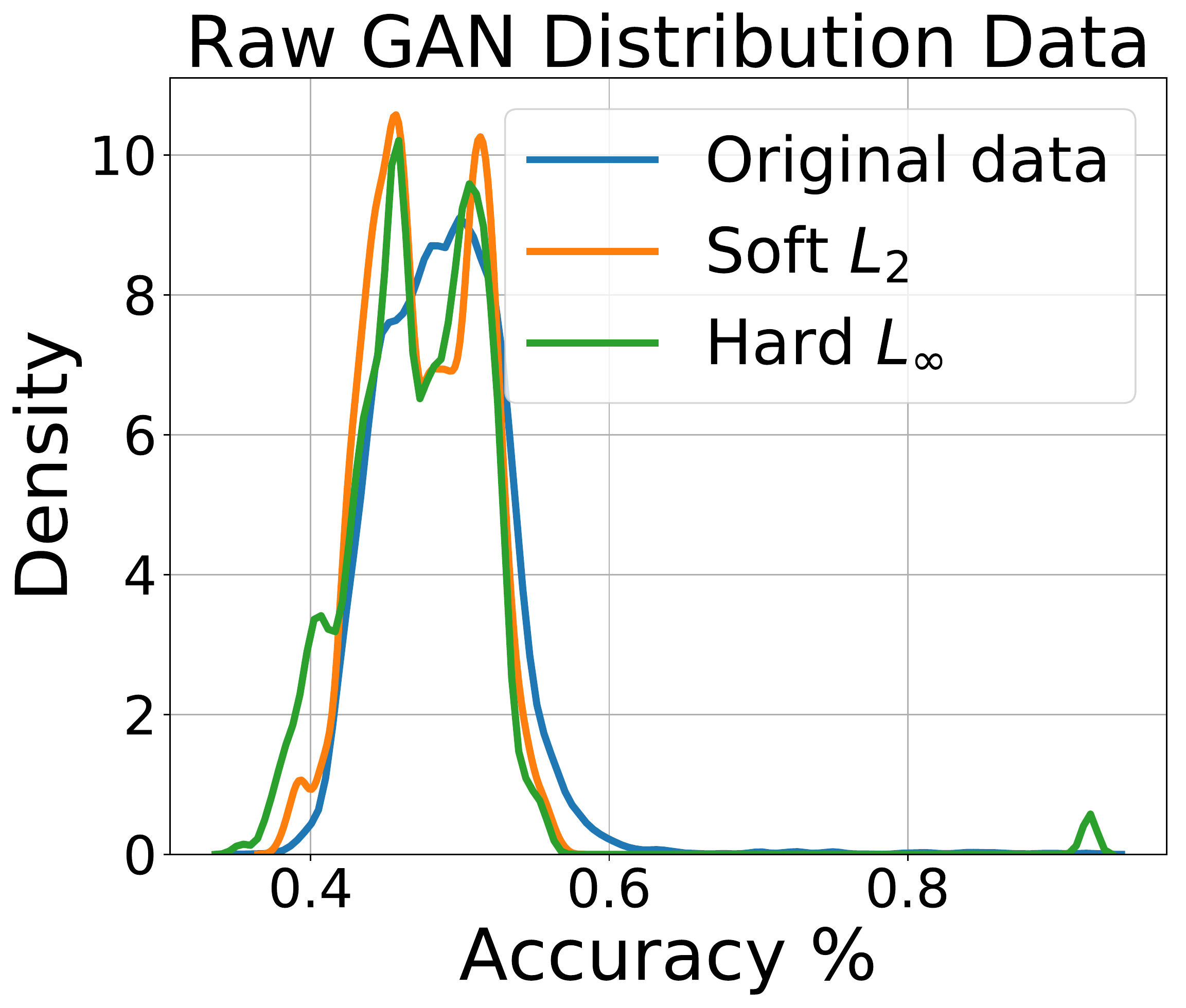} \\
\hspace{8pt} \large Discriminator $\mathbf{D}$ score & \hspace{8pt} \large Discriminator $\mathbf{D}$ score \\

\end{tabular}
\caption{\small \textbf{GAN instead of AE:} We tried to use l-GAN and r-GAN from \cite{pc-ae} as natural priors for AdvPC attacks instead of the AE. The discriminators of Both l-GAN and r-GAN could not discriminate between the original data and the attacks data by soft $\ell_2$ loss or Hard $\ell_\infty$. We show the histogram distribution of discriminator scores of the original data and attacked data using l-GAN discriminator (\textit{left}) and r-GAN discriminator (\textit{right}). }
\label{fig:sup-l-GAN}
\end{figure*}

We try to use the l-GAN and r-GAN from \cite{pc-ae} to create more natural attacks to the input point clouds. We try to leverage the discriminator signal of both r-GAN and l-GAN to differentiate between the perturbed point clouds and the original samples. The idea is that if the trained discriminator can distinguish between the unperturbed and attacked samples, then we add the discriminator loss as an additional loss to the attack objective in \eqLabel{\ref{eq:sup-attack-hard}} to craft a perturbation that passes the trained discriminator test of natural input. We train l-GAN and r-GAN with the same procedure advised by \cite{pc-ae} and on the same data as our AE $\mathbb{G}$. However, as \figLabel{\ref{fig:sup-l-GAN}} illustrates, neither l-GAN nor-GAN were able to distinguish between the unperturbed samples and the perturbed samples using the attack from \cite{pcattack}. This disappointing result leads us to abandon the approach in favor of the AE optimization (which works).

\subsection{Learning Approach}
Inspired by the success of \cite{learnattack1,learnattack2} in learning to attack, we tried to learn the AE $\mathbf{G}$ that produces the desired perturbed point cloud $\mathcal{X}^{\prime}$ by optimizing the output of the AE by the soft adversarial loss. To achieve this, the AE should output points clouds that are close as possible to the input point cloud $\mathcal{X}^{\prime}$ (by the Chamfer soft loss as in \eqLabel{\ref{eq:sup-attack-soft}}) and also the output of the AE should fool the classifier $\mathbf{F}$. We note that because of the nature of point cloud, we could not project the output of the AE back to the original sample with some norm, as performed by \cite{learnattack1}, and hence we used the soft Chamfer loss instead. We train the AE to perform untargeted and targeted attacks on the training set of ModelNet40 \cite{modelnet} and evaluate the adversary $\mathbf{G}$ on the test set of ModelNet. We report the results of targeted and untargeted attacks in Table \ref{tbl:learn-approach}. We note that for the untargeted attacks that indeed succeed, the final Chamfer Distance is way bigger than the ones obtained by optimization (see Tables \ref{tbl:soft-pn}). This might be attributed to the difficulty of learning an attack that works under varying distance penalties, unlike the \cite{learnattack1} where the adversarial objective is hardly conditioned on a constant distance between the attacked image and the original image. 
\begin{table*}[h!]
\footnotesize
\setlength{\tabcolsep}{4pt} %
\renewcommand{\arraystretch}{1.1} %
\centering
\resizebox{\hsize}{!}{

\begin{tabular}{ccccc|cc}
\toprule
\textbf{Loss}         & \textbf{\specialcell{ $\boldsymbol{\lambda_{\text{CD}}}$ }}& \textbf{Learning rate}      & \textbf{\specialcell{Untargeted/ \\ Targeted}} & \textbf{\specialcell{Training\\ Epochs}} & \textbf{Accuracy}      & \textbf{\specialcell{Chamfer \\Distance}}  \\
\midrule
Relativistic & 0       & 0.0001  & Untargeted   & 15    & 6.375  & 2.6161 \\ \hline
Relativistic & 1       & 0.0001  & Untargeted   & 15    & \textbf{5.0833} & 2.4832 \\ \hline
Relativistic & 3       & 0.0001  & Untargeted   & 13    & 9.9167 & 0.024229 \\ \hline
Relativistic & 10      & 0.0001  & Untargeted   & 13    & 8.9583 & 0.022948 \\ \hline
Relativistic & 30      & 0.0001  & Untargeted   & 13    & 10.125  & 0.021558 \\ \hline
Relativistic & 100     & 0.0001  & Untargeted   & 13    & 13.625  & 0.018275 \\ \hline
Relativistic & 300     & 0.0001  & Untargeted   & 16    & 17.875  & 0.014084 \\ \hline
Relativistic & 1000    & 0.0001  & Untargeted   & 13    & 38.0417 & 0.09286 \\ \hline
Relativistic & 0       & 0.00005 & Untargeted   & 13    & 9.7083 & 0.023431 \\ \hline
Relativistic & 0       & 0.0001  & Targeted     & 0     & 81.5    & \textbf{0.005556} \\ \hline
Relativistic & 0       & 0.001   & Targeted     & 19    & 31.875  & 0.015868 \\ 
 \bottomrule
\end{tabular}
}
\caption{\small \textbf{The Learning Approach on PointNet:} We tried to learn a network to attack PointNet \cite{pointnet} (approach similar to \cite{learnattack1} but on point clouds). While the approach mildly succeeds on untargeted attacks, the final average Chamfer distance on the succeeding attacks are much bigger than those obtained by optimization like in \figLabel{\ref{tbl:soft-pn}}. This implies that the optimization is actually better on point lcouds.}
\vspace{-10pt}
\label{tbl:learn-approach}
\end{table*}

\begin{table*}[]

\end{table*}
\end{document}